\documentclass{article} % For LaTeX2e
\usepackage{iclr2023_conference,times}

\usepackage{hyperref}
\usepackage{url}

\usepackage{nicefrac}       % compact symbols for 1/2, etc.
\usepackage{common}

\title{\ourtitle}

\newcommand{\xoffset}{\hspace*{-5pt}}
\author{%
	\xoffset\harpauthors\\
	\xoffset\kastel, \kitDE\\
}

\iclrfinalcopy % Uncomment for camera-ready version, but NOT for 
\begin{document}
	
\maketitle

%\vspace*{-1mm}

\begin{abstract}%\vspace*{-1mm}

Neural networks can be drastically shrunk in size by removing
redundant parameters. While crucial for the deployment on
resource-constraint hardware, oftentimes, compression comes with a
severe drop in accuracy and lack of adversarial robustness.
% lack of robustness against adversarial input manipulations.
% .
Despite recent advances, counteracting both aspects has only succeeded
for moderate compression rates so far. We propose a novel method,
\ourmethod, that copes with aggressive pruning significantly better than
prior work.
% .
For this, we consider the network holistically. \revision{We learn a
global compression strategy that optimizes how many parameters
(compression rate) and which parameters (scoring connections) to prune
\emph{specific to each \mbox{layer individually}}.}
% .
Our method fine-tunes an existing model with dynamic regularization, that
follows a step-wise incremental function balancing the different
objectives. It starts by favoring robustness before shifting focus on
reaching the target compression rate and only then handles the
objectives equally.
% .
The learned compression strategies allow us to maintain the pre-trained
model's natural accuracy and its adversarial robustness for a
reduction by \perc{99} of the network's original size. Moreover, we
observe a crucial influence of non-uniform compression across layers. 
The implementation of \ourmethod is publicly available at \smaller[0]{\projecturl}.

\end{abstract}%\vspace*{-3mm}

\section{Introduction}%\vspace*{-1mm}

Deep neural networks~(DNNs) yield remarkable performances in
classification tasks in various domains~\citep{He2016Resnet,
Cakir2018Malware, Schroff2017Facenet,Li2022TextClass} but are
vulnerable to input manipulation attacks such as adversarial
examples~\citep{Szegedy2014Intriguing}. Small perturbations to benign
inputs can cause worst-case errors in prediction. %, deceiving DNNs.
% .
To date, the most promising defensive approach against this sort of
attacks is adversarial training as introduced
by~\citet{Madry2018Towards} and further refined ever
since~\citep{Shafahi2019Adversarial, Zhang2019Theoretically,
Wang2020MART}. It introduces adversarial examples into the training
process, diminishing the generalization gap between natural performance
and adversarial robustness. 
% .
% While robust models are essential for the deployment of learning
% systems in safety-critical applications such as autonomous
% driving~\citep{Feng2018TowardsSA, Le2018UncertaintyAD}, also a small
% memory footprint and economic computations are crucial when run on
% resource-constraint platforms~\citep{Han2015Learning,
% Wu2016QuantizedCNN, Kim2015CompressCNN}.
% .
However, there is evidence indicating that higher robustness requires
over-parameterized networks that have wider layers and higher structural
complexity~\citep{Madry2018Towards, Zhang2019Theoretically,
Wu2021WiderNN}, rendering the task of combining both
objectives---compression and robustness---inherently difficult.

Neural network pruning~\citep{Han2016Deep, Yang2017EnergyPrune,
He2017Channel}, for instance, has been proven to be an extraordinary
valuable tool for compressing neural networks. The model can be reduced to
a fraction of its size by removing redundancy at different structural
granularity~\citep{Li2017Pruning, Mao2017Explore, Kollek2021MultiGrain}.
% .
% Fine-grained pruning or \emph{weight pruning} removes network
% parameters individually, while filter pruning directly removes output
% dimensions. The latter is commonly also referred to as
% \emph{channel~pruning} when the pruning objective focuses on the
% network's input dimensions~\citep{He2017Channel}.
% .
% XXX: CHANNEL FOCUS Pruning channels is particularly beneficial for the
% deployment on hardware with constraint resources as it reduces the
% dimensionality of the necessary computations, speeding up
% inference~\citep{Han2016Deep, Li2017Pruning, He2017Channel,
% Iandola2016SqueezeNet}.
% .
However, pruning inflicts a certain recession in model
accuracy~\citep{Han2016Deep} and adversarial
robustness~\citep{Timpl2021Understand} that is unavoidable the stronger
the model is compressed.
% .
The aim of adversarial robust pruning hence is to maintain the accuracy
and robustness of an already adversarially pre-trained model as good as
possible.
% .
Despite great efforts%in the recent past
~\citep{Ye2019Adversarial,
Sehwag2020HYDRA, Madaan2020Adversarial, Ozdenizci2021Bayesian,
Lee2022MAD}, %Han2015Learning, Han2016Deep
% .
this dual objective has only been achieved for moderate compression 
%rates 
so far.

\begin{figure*}[t]%\vspace{-3mm}
\begin{tikzpicture}
  \node(WP) [] at (0, 0.1) {
    \includegraphics[width=30mm, trim=5mm 0 7mm 0, 
    clip]{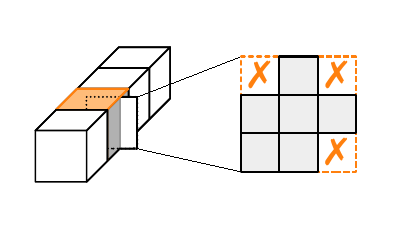}};
  \node(text) [below=8mm of WP, anchor=south] {
    \scriptsize Weight pruning 
  };

  \node(plot1) [] at (4.5, 0.1) {  % was:[right=0mm of WP] {
  	\begin{tikzpicture}
\begin{semilogxaxis}[
	log ticks with fixed point,
	height=0.2\textwidth,
	width=0.3\textwidth,
	legend pos=outer north east,
	enlarge x limits=0.02,
	enlarge y limits=0.05,
	grid=both,
	grid style={dashed,gray!30},
	smooth,
	tension=0.0,
	xlabel={Sparsity},
	ylabel={Accuracy [\%]},
	xmin=0.001,
	xmax=1,
	ymin=10,
	ymax=90,
	x dir=reverse,
	xtick={0.001, 0.01, 0.1, 1.0},
	ytick={10, 30, 50, 70, 90}, 
	xticklabels={99.9\%, 99\%, 90\%, 0\%},
	yticklabels={10, 30, 50, 70, 90},
	yticklabel style = {font=\fontsize{5}{5}\selectfont},
	xticklabel style = {font=\fontsize{5}{5}\selectfont},
	ylabel style = {font=\fontsize{6}{6}\selectfont, yshift=-4ex},
	xlabel style = {font=\fontsize{6}{6}\selectfont, yshift=1ex},
	%x tick label as interval
	scale only axis,
	legend image post style={scale=0.6},
	legend style={at={(0.0, 0.05)}, font=\legendsize, anchor=south 
	west, 
		legend columns=2, fill=white, draw=none, nodes={scale=0.6, 
		transform shape}, column sep=2pt, line width=0.8pt},
	legend cell align={left}
	]
	
	\addplot [mark=square*, color=secondarycolor, mark 
	options={scale=0.5}, 
	line 
	width=0.6pt]
	coordinates {
		(1, 79.68)
		(0.7, 80.10)
		(0.5, 80.23)
		(0.3, 80.14)
		(0.1,	80.12)
		(0.01,	78.50)
		(0.005,	74.42)
		(0.001,	59.13)
	}; \addlegendentry{\ourmethod}
	
	\addplot [mark=*, color=tertiarycolor, mark options={scale=0.5}, 
	line width=0.6pt]
	coordinates {
		(1, 79.68) % 47.60
		(0.7, 80.22)
		(0.5, 80.54)
		(0.3, 80.19)
		(0.1,	77.31)
		(0.01,	65.09)
		(0.005,	50.91)
		(0.001,	24.48)
	}; \addlegendentry{\hydra}

	\addplot [mark=triangle*, color=primarycolor, mark 
	options={scale=0.5}, 
	line 
	width=0.6pt]
	coordinates {
		(1, 79.68)
		(0.7, 80.43)
		(0.5, 80.38)
		(0.3, 79.90)
		(0.1,	75.41)
		(0.01,	61.84)
		(0.005,	52.95)
		(0.001,	22.59)
	}; \addlegendentry{\radmms}
	
	\addplot [mark=diamond*, color=black, mark options={scale=0.5}, 
	line 
	width=0.6pt]
	coordinates {
		(0.7, 79.16)
		(0.5, 79.32)
		(0.3, 79.65)
		(0.1,	78.32)
		(0.01,	69.81)
		(0.005,	64.96)
		(0.001,	10.00)
	}; \addlegendentry{\bcsp}
	
	\addplot [mark=square*, color=secondarycolor, mark 
	options={scale=0.5}, 
	line 
	width=0.6pt]
	coordinates {
		(1, 79.68)
		(0.7, 80.10)
		(0.5, 80.23)
		(0.3, 80.14)
		(0.1,	80.12)
		(0.01,	78.50)
		(0.005,	74.42)
		(0.001,	59.13)
	};
	
	\addplot [mark=*, densely dashed, color=tertiarycolor, mark 
	options={scale=0.5}, 
	line 
	width=0.6pt]
	coordinates {
		(1, 42.12)
		(0.7, 41.40)
		(0.5, 41.10)
		(0.3, 41.72)
		(0.1,	41.38)
		(0.01,	36.15)
		(0.005,	27.98)
		(0.001,	16.14)
	}; %\addlegendentry{\emph{Adv} (HYDRA)}
	
	\addplot [mark=triangle*, densely dashed, color=primarycolor, 
	mark 
	options={scale=0.5}, line 
	width=0.6pt]
	coordinates {
		(1, 42.12)
		(0.7, 42.61)
		(0.5, 42.88)
		(0.3, 43.20)
		(0.1,	40.37)
		(0.01,	32.21)
		(0.005, 28.64)
		(0.001, 19.74)
	}; %\addlegendentry{\emph{Adv} (ADMM)}
	
	\addplot [mark=diamond*, densely dashed, color=black, mark 
	options={scale=0.5}, line 
	width=0.6pt]
	coordinates {
		(0.7, 42.57)
		(0.5, 42.65)
		(0.3, 43.37)
		(0.1,	43.21)
		(0.01,	36.82)
		(0.005, 33.26)
		(0.001, 10.00)
	}; %\addlegendentry{\emph{Adv} (BCSP)}

	\addplot [mark=square*, densely dashed, color=secondarycolor, 
	mark 
	options={scale=0.5}, line 
	width=0.6pt]
	coordinates {
		(1, 42.12)
		(0.7, 41.98)
		(0.5, 41.35)
		(0.3, 41.65)
		(0.1,	41.72)
		(0.01,	42.48)
		(0.005,	40.78)
		(0.001,	31.74)
	}; %\addlegendentry{\emph{Adv} (HARP)}
	
\end{semilogxaxis}
	
\end{tikzpicture}

  };
  \node(text1) [below=2mm of plot1, anchor=south] {
	 	\scriptsize \qquad\cifar[10]
  };

  \node(plot2) [] at (10, 0.1) { % was:[right=2mm of plot1] {
  	\begin{tikzpicture}
\begin{semilogxaxis}[
	log ticks with fixed point,
	height=0.2\textwidth,
	width=0.3\textwidth,
	legend pos=outer north east,
	enlarge x limits=0.02,
	enlarge y limits=0.05,
	grid=both,
	grid style={dashed,gray!30},
	smooth,
	tension=0.0,
	xlabel={Sparsity},
	ylabel={Accuracy [\%]},
	xmin=0.001,
	xmax=1,
	ymin=10,
	ymax=90,
	x dir=reverse,
	xtick={0.001, 0.01, 0.1, 1.0},
	ytick={10, 30, 50, 70, 90}, 
	xticklabels={99.9\%, 99\%, 90\%, 0\%},
	yticklabels={10, 30, 50, 70, 90},
	yticklabel style = {font=\fontsize{5}{5}\selectfont},
	xticklabel style = {font=\fontsize{5}{5}\selectfont},
	ylabel style = {font=\fontsize{6}{6}\selectfont, yshift=-4ex},
	xlabel style = {font=\fontsize{6}{6}\selectfont, yshift=1ex},
	%x tick label as interval
	scale only axis,
	legend image post style={scale=0.6},
	legend style={at={(0.0, 0.05)}, font=\legendsize, anchor=south 
		west, 
		legend columns=2, fill=white, draw=none, nodes={scale=0.6, 
			transform shape}, column sep=2pt, line width=0.8pt},
	legend cell align={left}
	]
	
	\addplot [mark=square*, color=secondarycolor, mark 
	options={scale=0.5}, 
	line 
	width=0.6pt]
	coordinates {
		(1, 90.40)
		(0.7, 90.60)
		(0.5, 90.59)
		(0.3, 90.24)
		(0.1,	89.36)
		(0.01,	88.92)
		(0.005,	88.58)
		(0.001,	83.10)
	}; % \addlegendentry{\ourmethod}
	
	\addplot [mark=*, color=tertiarycolor, mark options={scale=0.5}, 
	line width=0.6pt]
	coordinates {
		(1, 90.40)
		(0.7, 91.22)
		(0.5, 91.08)
		(0.3, 91.11)
		(0.1,	90.49)
		(0.01,	85.68)
		(0.005,	76.34)
		(0.001,	23.87)
	}; % \addlegendentry{\hydra}

	\addplot [mark=triangle*, color=primarycolor, mark 
	options={scale=0.5}, 
	line 
	width=0.6pt]
	coordinates {
		(1, 90.40)
		(0.7, 88.80)
		(0.5, 89.28)
		(0.3, 88.55)
		(0.1,	89.51)
		(0.01,	79.39)
		(0.005,	67.04)
		(0.001,	10.00)
	}; % \addlegendentry{\radmms}
	
	\addplot [mark=diamond*, color=black, mark options={scale=0.5}, 
	line 
	width=0.6pt]
	coordinates {
		(0.7, 90.28)
		(0.5, 90.27)
		(0.3, 90.52)
		(0.1,	89.41)
		(0.01,	86.97)
		(0.005,	79.96)
		(0.001,	10.00)
	}; %\addlegendentry{\bcsp}
	
	\addplot [mark=square*, color=secondarycolor, mark 
	options={scale=0.5}, 
	line 
	width=0.6pt]
	coordinates {
		(1, 90.40)
		(0.7, 90.60)
		(0.5, 90.59)
		(0.3, 90.24)
		(0.1,	89.36)
		(0.01,	88.92)
		(0.005,	88.58)
		(0.001,	83.10)
	};
	
	%%%%%%%%%%%%%%%%%%%%%%%%%%%%%%%%%%%%%%%%%%%%%%%%%%%%%%%%%%%
	
	\addplot [mark=*, densely dashed, color=tertiarycolor, mark 
	options={scale=0.5}, 
	line width=0.6pt]
	coordinates {
		(1, 45.30)
		(0.7, 42.92)
		(0.5, 42.63)
		(0.3, 41.38)
		(0.1,	40.75)
		(0.01,	36.24)
		(0.005,	31.74)
		(0.001,	11.99)
	}; %\addlegendentry{\emph{Adv} (HYDRA)}
	
	\addplot [mark=triangle*, densely dashed, color=primarycolor, 
	mark 
	options={scale=0.5}, line 
	width=0.6pt]
	coordinates {
		(1, 45.30)
		(0.7, 44.12)
		(0.5, 44.08)
		(0.3, 44.23)
		(0.1,	42.82)
		(0.01,	25.79)
		(0.005, 21.44)
		(0.001, 10.00)
	}; %\addlegendentry{\emph{Adv} (ADMM)}
	
	\addplot [mark=diamond*, densely dashed, color=black, mark 
	options={scale=0.5}, line 
	width=0.6pt]
	coordinates {
		(0.7, 46.02)
		(0.5, 46.58)
		(0.3, 47.33)
		(0.1,	46.02)
		(0.01,	40.34)
		(0.005, 34.86)
		(0.001, 10.00)
	}; %\addlegendentry{\emph{Adv} (BCSP)}
	
	\addplot [mark=square*, densely dashed, color=secondarycolor, 
	mark options={scale=0.5}, line 
	width=0.6pt]
	coordinates {
		(1, 45.30)
		(0.7, 47.11)
		(0.5, 46.84)
		(0.3, 46.97)
		(0.1,	47.41)
		(0.01,	42.36)
		(0.005,	40.42)
		(0.001,	32.47)
	}; %\addlegendentry{\emph{Adv} (HARP)}
	
\end{semilogxaxis}
	
\end{tikzpicture}
  };
  \node(text2) [below=2mm of plot2, anchor=south] {
	\scriptsize \qquad\svhn
  };
\end{tikzpicture}%\vspace*{-2mm}
\vspace*{-1mm}
\caption{Overview of pruning weights of a \vgg model for
\cifar[10] (left) and \svhn (right) \revision{with \pgd-10 adversarial 
training}. 
Solid lines show the natural
accuracy of \ourmethod, \hydra, \radmm and \revision{\bcsp} (\cf \cref{tab:abbr-summary}).
Dashed lines represent the robustness \revision{against 
\Autoattack}.}\vspace{-4mm}
%Dashed lines represent the robustness \revision{against 
%	\Autoattack}. \revision{Abbreviations of related work are 
%	explained in~\cref{tab:abbr-summary}}}\vspace{-4mm}
\label{fig:overview}
\end{figure*}
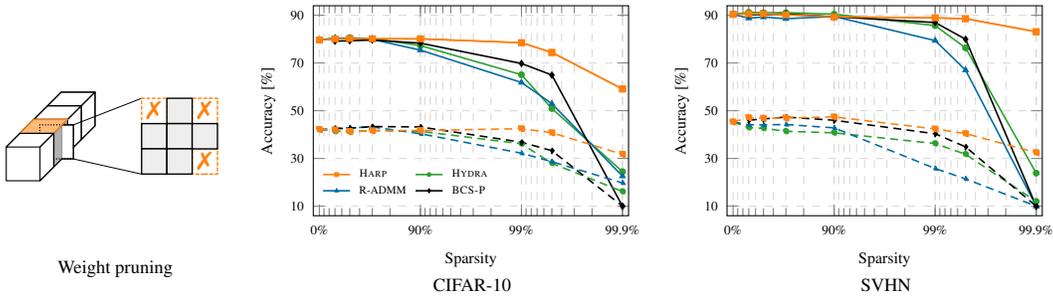

% \paragraph{Proposed method}
In this paper, we start from the hypothesis that effective adversarially
robust pruning requires a non-uniform compression strategy with
learnable pruning masks, and propose our method \ourmethod.
% .
We follow the three-stage pruning pipeline proposed by
\citet{Han2015Learning}
% and used by various works every since~\citep{Liu2019Rethinking,
% Han2015Learning, Molchanov2017Variational, Li2017PruneFilter}
to improve upon pre-trained models, where we jointly optimize
score-based pruning masks and layer-wise compression rates during
fine-tuning.
% .
As high robustness challenges the compactness
objective~\citep{Timpl2021Understand}, we employ a step-wise increasing
weighted-control of the number of weights to be pruned, such that we can
learn masks and rates simultaneously.
% .
%Our approach explores a global pruning strategy that allows for on-par
%natural accuracy in moderate compression and less robustness
%degradation when pruning highly aggressively.
Our approach explores a global pruning strategy that allows for on-par
natural accuracy with little robustness degradation only.
%
% Optimizing connection importance-scores and layer compression-rates
% are independent of the pruning granularity, such that \ourmethod is
% applicable to both coarse-grained channel pruning and fine-grained
% weight pruning.
%

\begin{revisionblock}
\noindent In summary, we make the following contributions:\\[-5mm]

\begin{itemize}[leftmargin=2em]
%\itemsep0em

\iffalse
\item\textbf{Novel pruning technique for pre-trained models.} We
optimize how many parameters and which parameters to prune \emph{for
each layers individually}. Additionally, we dynamically regularize the
pruning process to balance pruning objectives: natural accuracy and
adversarial robustness. In combination, these aspects allow for high
adversarialrobustness despite large compression (\cf
\cref{sec:ourmethod}).

\item\textbf{Importance of layer-wise compression.} In combination,
\ourmethod's objectives (compression and connection importance) yield a
global but non-uniform pruning strategy that allows for high adversarial
robustness despite large compression. We show that both aspects are
needed for \ourmethod to take full effect (\cf
\cref{sec:influence-factor}).
\fi

\item\textbf{Novel pruning technique for pre-trained models.} We
optimize how many parameters and which parameters to prune \emph{for
each layer individually}, resulting in a global but non-uniform pruning
strategy. That is, the overall network is reduced by a predetermined
rate governed by the target hardware’s limits, but layers are compressed
varyingly strongly. We show that both aspects are needed for \ourmethod to
take full effect (\cf~\cref{sec:influence-factor}).

\item\textbf{Significant improvement over related work.} An overview of
our method's performance  is presented in \cref{fig:overview} for a
\pgdat trained \vgg model learned on \cifar[10] (left) and \svhn
(right), providing a first glimpse of the yield advances. No less 
importantly, we
conduct experiments with \mbox{small (\cifar[10], \svhn) as well as
large-scale (\imagenet)} datasets across different robust training
methods and various attacks (\cf \cref{sec:compare-sota}).

\item\textbf{Importance of non-uniform pruning for adversarial
robustness.} We emphasize the superiority of non-uniform
strategies~\citep{Zhao2022Nonuniform} by extending existing
adversarially robust pruning-techniques.
% initially designed for uniform compression.
We demonstrate that \hydra~\citep{Sehwag2020HYDRA} and
\radmm \citep{Ye2019Adversarial} yield better results when used with
non-uniform strategies determined by \erk~\citep{Evci2020Rigging} and
\lamp~\citep{Lee2021Layer} (\cf \cref{sec:stg-compare}).
\end{itemize}
\end{revisionblock}

% into the empirical adversarial training optimization

\newcommand{\firstobs}{\oneb}
\newcommand{\secondobs}{\twob}
\newcommand{\thirdobs}{\threeb}

\section{Neural Network Pruning}\vspace*{-1mm}

Removing redundant parameters reduces a network's overall memory
footprint and necessary computations, allowing for a demand-oriented
adaptation of neural networks to resource-constrained
environments~\citep{Han2015Learning, Han2016Deep, Wen2016Learning,
Huang2018Learning, He2018AMC, He2017Channel, Li2017Pruning,
Mao2017Explore, Molchanov2017Variational}. Neural network pruning
attempts to find a binary mask~$\mask^{(\layer)}$ for each
layer~$\layer$ of a network with \Layer layers represented by its
parameters~\model and $\params^{(\layer)}$, respectively. The overall
pruning-mask thus is denoted as
\mbox{$\labelx{eq:masks}{\mask=(\mask^{(1)},\ldots,\mask^{(\layer)},\ldots,\mask^{(\Layer)})}$}.
These masks specify which parameters of the layer~$\params^{(\layer)}$
to remove (zero out) and which to keep, yielding a reduced parameter
set~$\prunedparamsl = \params^{(\layer)}\odot\mask^{(\layer)}$, where
$\odot$ is the Hadamard product.
% .
Based on this, we define the overall compression rate~\compression as
the ratio of parameters preserved after pruning, \npreservedparams, to the
number of parameters in the model, \nparams. Compression rates for
individual layers~$\compression^{(\layer)}$ are defined analogously.
% .
Note, that a network's \emph{sparsity} is defined inversely, meaning, a
\perc{99.9} sparsity refers to a compression of \num{0.001}.
% .
% .
In further follow, we consider $\params^{(\layer)} \in
\mathbb{R}^{c_i^{(\layer)} \times c_o^{(\layer)} \times k^{(\layer)}
\times k^{(\layer)}}$ where $c_i^{(\layer)}$ and $c_o^{(\layer)}$
represent input and output channels, respectively, and $k^{(\layer)}$ is
the kernel size.

\citet{Han2015Learning} propose a three stage pipeline for network
pruning, starting with (1)~training an over-parameterized network,
followed by (2)~removing redundancy per layer according to some pruning
criterion, before (3)~recovering network performance via fine-tuning.
% .
The actual pruning strategy, that decides on the pruning mask~\mask, is
obtained in the second step.
% and depends on the aimed for pruning granularity.
% \citet{Han2016Deep} further propose ``Lowest Weight Magnitude''~(LWM)
% to weight the importance of parameters and remove the least important
% once.
% % For weight pruning, they use the absolute value directly, and the %
% channel-wise $l_1$-norm for pruning channels.
% % .
% Other methods propose to use standard deviation~\citep{Sun2019Pruning}
% or develop a learnable criterion that are optimized by machine
% learning approaches~\citep{He2017Channel, Huang2018Learning}.
% .
While integrated approaches (pruning during model training)
exist~\citep[\eg][]{Vemparala2021Adversarial, Ozdenizci2021Bayesian},
the staged process remains most common~\citep{Liu2019Rethinking,
Sehwag2020HYDRA, Lee2022MAD} as it allows to benefit from recent
advantages in adversarial training~\citep{Shafahi2019Adversarial,
Zhang2019Theoretically, Wang2020MART} out-of-the-box.

\section{\ourtitle} 

Two aspects are crucial when pruning neural networks:
\emph{How many and which parameters to prune}.
\revision{Current state-of-the-art methods for adversarially robust
pruning focus on the latter, allocating a fixed compression budget
governed by the target hardware's limits and using it uniformly for each
layer~\citep[\eg][]{Ye2019Adversarial, Sehwag2020HYDRA}.
% and then either increase model sparsity during
% training~\citep{Ye2019Adversarial} or searching for better structural
% connections~\citep{Sehwag2020HYDRA} given this budget.
% .
We argue that learning the optimal pruning amount---the compression
rate per layer---is equally crucial.} \ourmethod thus integrates
measures for compression rates and scoring connections in an empirical
min-max optimization problem based on adversarial training to learn a
suitable pruning mask~\mask for a pre-trained model~\model:
% .
% Simultaneously, adversarial training ensures robustness against
% attacks at competitive natural accuracy:
% .
\begin{equation}
%\labelx{eq:total-prune}
{
  \underset{\rates, \scores}{\min}~~~
  \labelx{eq:total-prune-inner}
  {
    \underset{(\x, \y) \sim	\mathcal{D}}{\mathbb{E}}
    \left[
      \underset{\delta}{\max} \left\{
        \lossrob (\params \odot \mask, \x+\delta, y)
      \right\}
    \right] +
    \gamma \cdot \losshw(
      \params \odot \mask,\compression_{t}
    ),
  }
}
\end{equation}

The inner maximization generates adversarial examples from a benign
input~\x with label~\y from a distribution $\mathcal{D} = \{\x, \y\}$ by
adding noise~$\delta$, and incorporates these when minimizing the
training loss~$\lossce$~\citep{Madry2018Towards}. The exact
formulation of \lossrob depends on the pre-trained model, for which we
evaluate \pgdat~\citep{Madry2018Towards},
\tradesat~\citep{Zhang2019Theoretically}, and
\martat~\citep{Wang2020MART}.
% .
Simultaneously, the model is subject to a pruning mask~\mask that involves two
parameters: First, the compression quota~\rates, which is a learnable
representation of compression rates, and second, scores~\scores for
determining the importance of the network's connections.
% .
In the following, we discuss \ourmethod's global compression control and
describe how pruning strategies are learned, before we elaborate on how
to dynamically regularize the training process using~$\gamma$ by
following a step-wise incremental function balancing the different
objectives.
% .
% An overview of the entire process is given in \cref{alg:harp-alg}.

\paragraph{Global compression control} The compression control is
encoded as an additional loss term~\losshw, that considers compression
for the entire network to reach a specific target
compression~$\compression_t$. However, it explicitly allows for
layer-specific rates, yielding an overall non-uniform compression
strategy: We measure the number of currently preserved (\ie non-zero)
weights across all \Layer~layers of the network, $\npreservedparams =
\sum_{\layer=1}^{\Layer}
\|\left(\mathbbm{1}_{w\neq0}\right)_{w\in\prunedparams^{(\layer)}}\|_1$,
%\|\left(\mathbbm{1}_{w\neq0}\right)_{w\in\params^{(\layer)}\odot\mask^{(\layer)}}\|_1$,
relative to the targeted total number of parameters at a compression
rate~$\compression_t$:
%~$\nparams=|w\in\params^{(\layer)}\}|$:
% .
\begin{equation}
\label{eq:hw-loss}
  \losshw(\prunedparams,\compression_{t}) := \max \left\{
    \frac{
      \npreservedparams
    }{
      \compression_{t}\cdot\nparams
    }
    -1
    ~,~ 0
\right\}
\end{equation}
\iffalse
\begin{equation}
\label{eq:hw-loss}
  wrong \losshw = \max \left\{
    -1 +
    \cfrac{1}{\compression_{t}} \cdot
    \sum_{\layer=1}^{\Layer}\frac{
      \npreservedparams^{(\layer)}
    }{
      \nparams^{(\layer)}
    }
    ~,~ 0
\right\}
\end{equation}
\begin{equation}
\label{eq:hw-loss}
  wrong \losshw = \max \left\{
  	-1 +
    \frac{1}{\compression_{t}} \cdot
    \sum_{\layer=1}^{\Layer}
    \frac{\left\|\left(\mathbbm{1}_{w\neq0}
      \right)_{w\in\prunedparams^{(\layer)}}
      \right\|_1
    }{
      \mid\{ w\in\params^{(\layer)}\} \mid
      %c_i^{(\layer)} c_o^{(\layer)} k^{(\layer)} k^{(\layer)}
    }  
    ~,~ 0
    \right\}~.
\end{equation}
\begin{equation}
\label{eq:hw-loss}
  \losshw = \max \left\{
  	\frac{
      \sum_{\layer=1}^{\Layer}
      \left\|\left(\mathbbm{1}_{w\neq0} \right)_{w\in\prunedparams^{(\layer)}}\right\|_1
    }{
      \compression_{t}\cdot|\{w\in\params^{(\layer)}\}|
    }
    -1
    ~,~ 0
    \right\}~.
\end{equation}
\fi
Additionally, we clip rates that would exceed the targeted compression-rate.
Constraining the loss to positive values ensures to not encourage
even lower compression rates that would potentially harm accuracy and
robustness unnecessarily.
% (see more details in Appendix).

For learning the layer-specific compression rates, moreover, it is
crucial that we constrain the reduction by a minimal
compression~$\compression_{min} = 0.1 \times \compression_{t}$ as we 
need to prevent a layer from been
removed completely. We thus limit the layers' compression
$\compression^{(\layer)}$ to $[\compression_{min}, 1]$.
% .
In order to better control learning of the optimal pruning rate in this
range, we introduce a trainable parameter $\rate^{(\layer)} \in
\mathbb{R}$, which we refer to as \emph{compression quota}, and
constrain it by an activation function~\mbox{$g:
\mathbb{R}\rightarrow[\compression_{min}, 1]$}. For the sake of
the continuous derivability on $\rate$, we use the sigmoid 
function squeezed to the desired output range, yielding a layer-wise 
compression rate
% $g:r\mapsto(1-\compression_{min}) \cdot \sigmoid(\rate^{(l)}) +
% \compression_{min}$.
$\compression^{(l)} = g(r^{(l)}), \text{where }
g:r\mapsto(1-\compression_{min}) \cdot \sigmoid(\rate) +
\compression_{min}$.

\paragraph{Connection importance}
For learning the actual pruning mask, in turn, we introduce another
learnable parameter $\scores$, the connection importance-score matrix,
to rate each pruning connection.
% .
In weight pruning, $\scores^{(l)}$ originates
$\mathbb{R}^{c_i^{(\layer)} \times c_o^{(\layer)} \times k^{(l)} \times
k^{(l)}}$, and is assigned values from activation function~$g$.
%
%Additionally, we use $P_{\alpha}(\cdot)$
%to represent the function that
%finds the cutoff threshold for pruning parameters in accordance with
%compression rate~$\compression^{(l)}$, that is, it determines the 
%$\alpha^{(l)}$-th
%percentile, with $\alpha = 1-\compression^{(\layer)}$.
%The binary pruning mask $\mask^{(\layer)}$ is thus defined as
%\begin{equation}
%\labelx{eq:p-mask}
%{
%	\mask^{(l)} = \left(
%	\mathbbm{1}_{s > P_{\alpha}(S^{(l)})}
%	\right)_{\score \in \scores^{(l)}}%,
%}
%\end{equation}
%
%\begin{equation} \label{eq:p-mask} \mask(\score, r,
%\compression_{min}) = \left\{ \begin{array}{ c l } 1  & \quad
%\textrm{if } \score > \score_{th}\\
%0  & \quad \textrm{otherwise} \end{array} \right.
%\textrm{, where: } s_{th} = percentile(\scores, \compression)
%\end{equation}
%
Additionally, we use 
$P(\alpha^{(l)}, \scores^{(l)})$
to represent the function that
determines the $\alpha$-th percentile on layer $l$ as the cutoff 
threshold for pruning parameters in accordance with
compression rate~$\compression^{(l)}$, that is: 
$\alpha^{(l)} = 1-\compression^{(\layer)}$.
Due to the relation with $\scores^{(l)}$ (and therefore also with 
$\rate^{(l)}$ via compression rate $\compression^{(l)}$), the binary 
pruning mask $\mask^{(\layer)}$ is thus trainable and defined as:
\begin{equation}
\labelx{eq:p-mask}
{
	\mask^{(l)} := \left(
	\mathbbm{1}_{s > P(\alpha^{(l)}, \scores^{(l)})}
	\right)%,
}
\end{equation}	
Pruning the $\layer$-th layer can hence be expressed by the Hadamard
product of the mask and the model's parameters~\params yielding the
pruned model parameters $\prunedparamsl =
\model^{(\layer)}\odot\mask^{(\layer)}$.
%.
%While weight pruning is largely independent of the network architecture
%as it does not change the layer's size (we~merely zero out weights),
%pruning channels or filter needs special attention.
%.
%Note that, when facing \resnet[-like] architectures, for instance, we 
%do not train
%the compression rates of all shortcut layers directly but update them 
%by
%assigning the rate of the connected input layer in the residual block
%instead. This way, the pruned input layer aligns with the channel
%dimensionality of connected pruned shortcut-layers.

% (outputs of the activation function~$g$)
\paragraph{Learning pruning strategies}
Binarizing importance scores to obtain a pruning mask as described above
is a non-differentiable operation and prevents us from performing
gradient descent over the masks. \revision{We, thus, follow the
``straight through estimation''~(STE)
strategy~\citep{Hubara2016Binarized} as proposed
by~\citet{Kusupati2021Soft}} to assign the updated gradients to the
importance scores~$\scores^{(\layer)}$ directly and proceed similarly
for compression quotas~$\rate^{(\layer)}$:
\begin{alignat}{3}
\label{eq:ste-opt}
    &
%       \tikz[baseline, remember picture]{
%       \node[anchor=west] (a) {$
        \cfrac{\partial \mathcal{L}}{\partial \scores^{(\layer)}}
%       $};}
    && = 
     \cfrac{\partial \mathcal{L}}{\partial \prunedparamsl} \cdot
     \cfrac{\partial \prunedparamsl}{\partial \mask^{(\layer)}} \cdot
     \cfrac{\partial \mask^{(\layer)}}{\partial \scores^{(\layer)}}
    && \ste
     \cfrac{\partial \mathcal{L}}{\partial \prunedparamsl} \cdot
     \cfrac{\partial \prunedparamsl}{\partial \mask^{(\layer)}} \\[2mm]
    &
%       \tikz[baseline, remember picture]{
%       \node[anchor=west] (b) {$
        \cfrac{\partial \mathcal{L}}{\partial \rate^{(\layer)}}
%       $};}
    && =
      \cfrac{\partial \mathcal{L}}{\partial \prunedparamsl} \cdot 
      \cfrac{\partial \prunedparamsl}{\partial \mask^{(\layer)}} \cdot
      \cfrac{\partial \mask^{(\layer)}}{\partial g(\rate^{(\layer)})} 
      \cdot g'(\rate^{(\layer)})
    && \ste
    \langle \cfrac{\partial \mathcal{L}}{\partial \prunedparamsl} \cdot 
    \cfrac{\partial \prunedparamsl}{\partial \mask^{(\layer)}} \rangle \cdot g'(\rate^{(\layer)})
\end{alignat} % \\[-0.5em]
Finally, also the initialization of the connection importance-scores is
crucial. We follow the lead of \citet{Sehwag2020HYDRA} and initialize
the scores proportional to the weights of the pre-trained model and use
a scaling factor $\eta = \sqrt{\nicefrac{k}{\fanin^{(l)}}}$ with $k=6$
and \fanin being the product of the receptive field size and the number
of input channels. Our initialization for pruning weights, hence, 
follows the order of weight magnitudes \citep{Ye2019Adversarial}: 
%In addition to the initial formulation for weight pruning (left), we 
%extend the initialization concept to channel pruning (right): 
%%%\\[-3.5em]
%.
%\begin{multicols}{2}
\begin{equation}
\labelx{eq:scores-init-w}
{
  \scores_{W}^{(\layer)} =
  \left(
    \eta\cdot\cfrac{\params^{(\layer)}}{\max(|\params^{(\layer)}|)}
  \right)_{c_i^{(\layer)}\times c_o^{(\layer)}\times k^{(\layer)}\times 
  k^{(\layer)}}
}
\end{equation}
%\end{multicols}
%
\ourmethod does not use a fixed ratio for pruning parameters, but learns
the optimal compression rate per layer. However, for initializing the
compression quota~\rates, we first align with a uniform compression
strategy that sets the same rate~$\compression_{init}$ on each layer and
let the optimizer improve it (\cf \cref{app:init-analysis} for more
details). For the activation function~$g(\cdot)$ defined above, we thus
yield:
% .
\begin{equation}
\labelx{eq:rate-init}
{
	\rate^{(\layer)} = \log \left(
	\cfrac{\compression_{init}-\compression_{min}}{1-\compression_{init}}
	\right)
}
\end{equation}

%\begin{equation}
%\labelx{eq:rate-init}
%{
%  \rate^{(\layer)} = \log \left(
%    \cfrac{\compression-\compression_{min}}{1-\compression}
%  \right)
%}
%\end{equation}

%%%%%%%%%% Algorithm %%%%%%%%%%%%%
%\begin{algorithm}[t]
%\caption{\ourtitle}
%\label{alg:harp-alg}
%\input{algorithms/harp}
%\end{algorithm}

% \tosolve{Description about rates initialization should be linked with
% experimental tables}

% Our formulation integrate layer compression rates and connection scores
% as the robust training objectives while softly controls the global
% compression rate such that model after pruning results in the desired
% compression degree with higher performance preservation.

\paragraph{Balancing pruning objectives}

\ourmethod strives to strike a balance between natural accuracy,
adversarial robustness and compactness. For this, we employ a step-wise
incremental function to adapt the regularization parameter~$\gamma$
throughout the learning process. We start by favoring robustness before
shifting focus on reaching the target compression rate and only then
handle the objectives equally.
% .
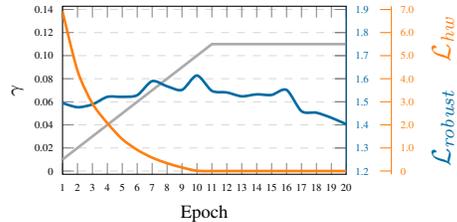
\begin{wrapfigure}{r}{0.45\textwidth}
  \vspace*{-2mm}
  \centering
  \begin{tikzpicture}
% f1
\begin{axis}[
%height=0.16\textwidth,
%width=0.27\textwidth,
%color=black,
%scale only axis,
%xmin=0.5,
%xmax=20.5,
%xtick={1,2,3,4,5,6,7,8,9,10,11,12,13,14,15,16,17,18,19,20},
%xticklabels={1,2,3,4,5,6,7,8,9,10,11,12,13,14,15,16,17,18,19,20},
%xlabel={Epoch},
%ylabel={$\gamma$},
%ymin=-0.01,
%ymax=0.14,
%ytick={0, 0.02, ..., 0.16},
%yticklabels={.0, .02, .04, .06, .08, .10, .12, .14, .16},
%%y axis line style={blue},
%%ytick style={blue},
%ylabel style={font=\fontsize{8}{8}\selectfont, yshift=-2ex},
%xlabel style = {font=\fontsize{8}{8}\selectfont, yshift=0ex},
%yticklabel style={font=\fontsize{6}{6}\selectfont},
%xticklabel style={font=\fontsize{5}{5}\selectfont},
%axis x line*=bottom,
%axis y line*=left,
%tick align = outside,]
height=0.16\textwidth,
width=0.27\textwidth,
legend pos=outer north east,
enlarge x limits=0.0,
enlarge y limits=0.02,
ymajorgrids, 
grid style={dashed,gray!30},
smooth,
tension=0.0,
xlabel={Epoch},
ylabel={$\gamma$},
xmin=1,
xmax=20,
ymin=0.0,
ymax=0.14,
xtick={1,2,...,20},
ytick={0.0, 0.02, ..., 0.16},
xticklabels={1,2,3,4,5,6,7,8,9,10,11,12,13,14,15,16,17,18,19,20},
yticklabels={0, 0.02, 0.04, 0.06, 0.08, 0.10, 0.12, 0.14, 0.16},
yticklabel style = {font=\fontsize{4}{4}\selectfont, 
yshift=0.0ex},
xticklabel style = {font=\fontsize{4}{4}\selectfont},
ylabel style = {font=\fontsize{7}{7}\selectfont, yshift=-4ex},
xlabel style = {font=\fontsize{7}{7}\selectfont, yshift=1.5ex},
%x tick label as interval
scale only axis,
legend style={at={(0.8,1)}, font=\fontsize{7pt}{7pt}\selectfont, 
anchor=north east, legend columns=2, fill=white, draw=gray, 
nodes={scale=0.7, transform shape}, column sep=2pt},
legend cell align={left}
]

\addplot [mark=none, color=cbthree, mark options={scale=0.5}, line 
width=1pt]
table[row sep=crcr]{
1	0.01\\
2	0.02\\
3	0.03\\
4	0.04\\
5	0.05\\
6	0.06\\
7	0.07\\
8	0.08\\
9	0.09\\
10	0.10\\
11	0.11\\
12	0.11\\
13	0.11\\
14	0.11\\
15	0.11\\
16	0.11\\
17	0.11\\
18	0.11\\
19	0.11\\
20	0.11\\
}; \label{gamma}
\end{axis}
% Loss_CE
\begin{axis}[
height=0.16\textwidth,
width=0.27\textwidth,
enlarge x limits=0.0,
enlarge y limits=0.02,
color=primarycolor,
scale only axis,
xmin=1,
xmax=20,
ymin=1.2,
ymax=1.9,
ytick={1.2, 1.3, ..., 2.0},
yticklabels={1.2, 1.3, 1.4, 1.5, 1.6, 1.7, 1.8, 1.9, 2.0},
yticklabel style = {font=\fontsize{4}{4}\selectfont,  
	yshift=0.0ex},
axis x line*=none,
axis y line*=right,
hide x axis,
smooth,
tension=0.5,
legend style={at={(1, 1)}, font=\fontsize{7pt}{7pt}\selectfont, 
anchor=north east, legend columns=3, fill=white, draw=none, 
nodes={scale=1.0, transform shape}, column sep=2pt},
legend cell align={left}
]

\addplot [mark=none, color=primarycolor, mark options={scale=0.5}, line 
width=1pt]
table[row sep=crcr]{
1	1.4949\\
2	1.4766\\
3	1.4888\\
4	1.5214\\
5	1.5215\\
6	1.5286\\
7	1.5896\\
8	1.5673\\
9	1.5518\\
10	1.6139\\
11	1.5476\\
12	1.5406\\
13	1.5239\\
14	1.5323\\
15	1.5298\\
16	1.5515\\
17	1.4596\\
18	1.4528\\
19	1.4310\\
20	1.4029\\
}; % \addlegendentry{$\lossce$}
\end{axis}

% Loss_HW
\begin{axis}[
height=0.16\textwidth,
width=0.27\textwidth,
enlarge x limits=0.0,
enlarge y limits=0.02,
ylabel={\textcolor{primarycolor}{$\lossrob$}~~~~\textcolor{secondarycolor}{$\losshw$}},
ylabel near ticks, yticklabel pos=right,
color=secondarycolor,
scale only axis,
xmin=1,
xmax=20,
ymin=0,
ymax=7,
ytick={0, 1, ..., 8},
yticklabels={0,1.0,2.0,3.0,4.0,5.0,6.0,7.0,8.0},
axis x line*=none,
axis y line*=right,
hide x axis,
smooth,
tension=0.5,
legend style={at={(1, 1)}, font=\fontsize{7pt}{7pt}\selectfont, 
anchor=north east, legend columns=3, fill=white, draw=none, 
nodes={scale=1.0, transform shape}, column sep=2pt},
legend cell align={left}
]
\pgfplotsset{every outer y axis 
line/.style={font=\fontsize{6}{6}\selectfont, xshift=0.6cm}, every 
tick/.style={xshift=0.6cm}, every y tick 
label/.style={font=\fontsize{4}{4}\selectfont, xshift=0.6cm}}

\addplot [mark=none, color=secondarycolor, mark options={scale=0.5}, line 
width=1pt]
table[row sep=crcr]{
1	6.9187\\
2	4.3540\\
3	2.9282\\
4	2.0767\\
5	1.3823\\
6	0.9187\\
7	0.5806\\
8	0.3391\\
9	0.1540\\
10	0.0105\\
11	0.0000\\
12	0.0007\\
13	0.0002\\
14	0.0007\\
15	0.0004\\
16	0.0004\\
17	0.0000\\
18	0.0001\\
19	0.0001\\
20	0.0000\\
}; % \addlegendentry{$\losshw$}
\end{axis}
\end{tikzpicture}\vspace{-3mm}%
  \caption{\revision{%
  		\ourmethod's step-wise regularization of pruning objectives
  		for \vgg on \cifar[10] with a target sparsity of $\perc{99}$.}}
  \label{fig:gamma-control}
  \vspace*{-3mm}
\end{wrapfigure}
% .
% In our experiments presented in \cref{sec:experiment}, we use \num{0.01}
% as the step-size for adjusting the regularization parameter.
\cref{fig:gamma-control} shows the relation of  $\gamma$ and the two
different losses, \lossrob against \pgd-10 and \losshw, on 
the y-axis over the epochs on the x-axis, exemplarily for weight 
pruning \vgg (learned on \cifar[10]) with a target 
compression~\mbox{$\compression_t=0.01$}, that is,
removing \perc{99} of the parameters.
% .
The model starts off from the 
uniform strategy with $\compression_{init}=0.1$,
%\mbox{($\compression_{w}=0.1$}, \mbox{$\compression_{c}=0.5$)} 
%used for initialization, 
learning to be more robust while being only slightly
penalized by \losshw. As the model has been adversarially pre-trained,
\lossrob is low already. Over time, we put more focus on
\losshw up until we have reached our target compression ($\gamma=0.11$
in this example), causing it to decrease while \lossrob increases.
% .
Afterwards, the model aims for higher robustness under the found
$\losshw$ penalty, balancing both objectives. 
\revision{
In \cref{sec:gamma-adapt}, we study the influence of different step
sizes and elaborate on the progression of the model's natural
accuracy and adversarial robustness in \cref{app:stganalyse}.
% (\cf~\cref{fig:gamma-adapt}) ~(\cf~\cref{fig:learn-process}).
}

\section{Evaluation}
\label{sec:experiment}

We resume to demonstrate \ourmethod's effectivity in pruning neural
networks by comparing to existing state-of-the-art methods in
\cref{sec:compare-sota}. In \cref{sec:influence-factor}, we then perform
an ablation study regarding the learnable parameters used by our method.
In particular, we show the influence of learning compression
quotas~\rates and importance scores~\scores, separately, to demonstrate
that both are indeed needed for successful, holistic pruning. In
\cref{sec:stg-compare} and \cref{app:stganalyse}, we then discuss the
non-uniform nature of \ourmethod's pruning strategy and compare to novel
extensions of \hydra and \radmm to non-uniform compression.
% .
In the appendix, we additionally present the extended 
comparison to related work~(\cref{app:related-work}), 
%discuss \ourmethod's initialization and the influence of non-uniform 
%strategies on relate work~(\cref{app:stganalyse}, 
elaborate on the pruned model's parameter 
distribution~(\cref{app:paramsdist}), 
show \ourmethod's performance on naturally trained 
models~(\cref{app:natprune}), and extend \ourmethod to channel 
pruning~(\cref{app:chprune}).

\paragraph{Experimental setup} We evaluate \ourmethod on two small-scale
datasets, \cifar[10]~\citep{CIFAR} and \svhn~\citep{Netzer2011SVHN}.
While the first is balanced, the second is not.
Consequently, we use accuracy~(\acc) as performance measure for the
former, and the balanced accuracy~(\bacc) for the
latter~\citep{Brodersen2010BACC}. Each dataset is learned with a
\vgg~\citep{Simonyan2015Very} and \resnet[18]~\citep{He2016Resnet}
model. Additionally, we show the performance on the large-scale,
balanced \imagenet dataset~\citep{Deng2009ImageNet} trained with a
\resnet[50] network.
% .
We apply $\gamma=0.01$ for small-scale datasets, and increase $\gamma$
to \num{0.1} for \imagenet to guarantee the arrival at the target
compression rate $\compression_t$. In the pruning and fine-tuning phase
of \ourmethod, we train for \num{20}~epochs and \num{100}~epochs,
respectively.

%For training the models, we employ adversarial training with different
%losses: \pgd~\citep{Madry2018Towards},
%\trades~\citep{Zhang2019Theoretically}, and \mart~\citep{Wang2020MART}
%for the small-scale datasets and \freeat~\citep{Shafahi2019Adversarial}
%with \num{4}~replays for the large-scale dataset.
% .
%For validating the pruned networks' adversarial robustness, we use
%\pgd~\citep{Madry2018Towards}, \cwinf~\citep{Carlini2017Towards},
%\Autopgd and \Autoattack~\citep{Croce2020AutoAttack}.
%\cref{app:expsetup} provides further details on the experimental setup.
%

\begin{revisionblock}%
\paragraph{Adversarial training}
We adversarially train on the small-scale datasets, \cifar[10] and
\svhn, with \pgdat~\citep{Madry2018Towards},
\tradesat~\citep{Zhang2019Theoretically}, and \martat
\citep{Wang2020MART}. For each, we use $l_{\infty}$ \pgd-10 attacks with
random initialization and a perturbation strength
$\strength=\sfrac{8}{255}$ with step size $\stepsize=\sfrac{2}{255}$.
We adopt stochastic gradient descent with a cosine
learning-rate schedule~\citep{Loshchilov2016SGDR} and a weight decay
of \num{0.0005}~\citep{Pang2021BagAT}.
For \trades and \mart, we follow the originally proposed 
regularization of $\regfac=6.0$ and $\regfac=5.0$, respectively.
% .
To accelerate the training process for the large-scale \imagenet
experiments, we resort to \freeat~\citep{Shafahi2019Adversarial} with
\num{4}~replays, reducing the pruning and fine-tuning phase to \num{5} 
epochs and \num{25} epochs, and use a perturbation strength
$\strength=\sfrac{4}{255}$ with step size $\stepsize=\sfrac{1}{255}$.
% , therewith the \num{20}~epochs for training used in small-scale
% datasets are reduced to merely \num{5}~epochs.
% All networks after adversarial pre-training are listed in
% \cref{tab:pretrain}.

% via  the PyTorch library Torchattacks~\citep{Kim2020torchattacks} for
% better robustness validation.

\paragraph{Adversarial robustness evaluation}
To validate the pruned networks' adversarial robustness, we use
\cwinf~\citep{Carlini2017Towards} optimized by \pgd and \pgd-10 with the
same settings used for training. Furthermore, we implement
\Autopgd(APGD)~\citep{Croce2020AutoAttack} with cross-entropy~(CE) loss,
\num{50}~steps, and \num{5}~restarts, and
\Autoattack(AA)~\citep{Croce2020AutoAttack} using the standard ensemble 
out of \Autopgd with CE-loss, $\text{Targeted-\Autopgd}_\text{DLR}$,
\fab~\citep{Croce2020FAB}, and \squareattack~\citep{Croce2019SquareA}.
All attacks are carried out on the respective complete test dataset.
\end{revisionblock}

%\begin{table}[tbh]
%	\centering
%	\caption{Adversarially pre-trained \vgg and \resnet[18] on 
%		\cifar[10] and \svhn.}
%	\tablesize
%	\input{tables/pretrain.tex}
%	\label{tab:pretrain}
%\end{table}

%For \svhn, since neither the training set nor the test 
%set in \svhn is class-wisely balanced as shown in \cref{tab:svhn_dist}, 
%we use the balanced accuracy~\citep{Brodersen2010BACC} as the 
%evaluation metric. Accordingly, a completely damaged model with random 
%prediction for \svhn is indicated by a balanced accuracy of 10\%.
%e.g. CIFAR10 test set with 10,000 images.

%%%% IMAGENET SETUP %%%% INCORPORATE THIS IN A.1
% % .
% As adversarial training can be a computational burden on large-scale
% datasets, we employ \freeat~\citep{Shafahi2019Adversarial} to accelerate
% training on \imagenet.
% We start from a robust \resnet[50] network that is pre-trained for
% \num{90}~epochs. And in the following pruning and fine-tuning stage,
% keep using the same adversarial setting as in pre-train stage, i.e.
% perturbation strength $\strength=\sfrac{4}{255}$ with step-size
% $\stepsize=\sfrac{1}{255}$.
% Note that, in pruning stage, we select step size \num{0.1} to
% dynamically adapt the regularization factor $\gamma$ such that it
% ensures the arrival of the target compression.

\newcommand{\ourmethodR}{\mbox{\ourmethod-\rates}\xspace}
\newcommand{\ourmethodS}{\mbox{\ourmethod-\scores}\xspace}

\begin{table}[!b]
	\centering
	\vspace*{-3mm}
	\caption{Optimizing either compression
	rates~(\ourmethodR) or importance scores~(\ourmethodS). Natural
	accuracy and \pgd-10 adversarial robustness are presented left and
	right of the \texttt{/} character.}
	\label{tab:rate-score}
	{
\newcommand{\mycsvreader}[4]{%
	\csvreader[
	head to column names,
	filter = \equal{\arch}{#2} \and \equal{\at}{#3} \and 
	\equal{\rate}{#4},
%	late after line=\\
	]%
	{#1}%
	{}
	{
%		& \csuse{\at}
%		& \oa
%		& \aa
%		& \rchoabf
%		& \rchaabf
%		& \schoabf
%		& \schaabf
%		& \hchoabf
%		& \hchaabf
		& \rwoabf
		& \rwaabf
		& \swoabf
		& \swaabf
		& \hwoabf
		& \hwaabf
	}
}

\begin{influencetable0}{\linewidth}
	\midrule
	\multirow{3}{*}{\resnet[18]} %{\rotatebox{90}{ResNet}}
	& \pgd
	\mycsvreader{results/rate_score_cifar10.csv}{resnet}{pgd}{0.01}
	\mycsvreader{results/rate_score_cifar10.csv}{resnet}{pgd}{0.001}
	\\
	& \trades
	\mycsvreader{results/rate_score_cifar10.csv}{resnet}{trades}{0.01}
	\mycsvreader{results/rate_score_cifar10.csv}{resnet}{trades}{0.01}
	\\
	& \mart
	\mycsvreader{results/rate_score_cifar10.csv}{resnet}{mart}{0.01}
	\mycsvreader{results/rate_score_cifar10.csv}{resnet}{mart}{0.01}
	\\
	\midrule %\mymidrule
%		&
	\multirow{3}{*}{\vgg} %{\rotatebox{90}{\vgg}}
	& \pgd
	\mycsvreader{results/rate_score_cifar10.csv}{vgg}{pgd}{0.01}
	\mycsvreader{results/rate_score_cifar10.csv}{vgg}{pgd}{0.001}
	\\
	& \trades
	\mycsvreader{results/rate_score_cifar10.csv}{vgg}{trades}{0.01}
	\mycsvreader{results/rate_score_cifar10.csv}{vgg}{trades}{0.001}
	\\
	& \mart
	\mycsvreader{results/rate_score_cifar10.csv}{vgg}{mart}{0.01}
	\mycsvreader{results/rate_score_cifar10.csv}{vgg}{mart}{0.001}
	\\
%	\midrule
%	&
%	\multirow{3}{*}{\wrn}
%	\mycsvreader{results/rate_score_cifar10.csv}{wrn}{pgd} 
%	\mycsvreader{results/rate_score_cifar10.csv}{wrn}{trades} 
%	\mycsvreader{results/rate_score_cifar10.csv}{wrn}{mart}
%	%
%	\midrule
%	%
%	\multirow{9}{*}{\rotatebox{90}{\svhn}} & 
%	\multirow{3}{*}{\resnet[18]} 
%	\mycsvreader{results/rate_score_svhn.csv}{resnet}{pgd} & 
%	\mycsvreader{results/rate_score_svhn.csv}{resnet}{trades} & 
%	\mycsvreader{results/rate_score_svhn.csv}{resnet}{mart}
%	\cmidrule{2-13} &
%	\multirow{3}{*}{\vgg} 
%	\mycsvreader{results/rate_score_svhn.csv}{vgg}{pgd} & 
%	\mycsvreader{results/rate_score_svhn.csv}{vgg}{trades} & 
%	\mycsvreader{results/rate_score_svhn.csv}{vgg}{mart}
%	\cmidrule{2-13} &
%	\multirow{3}{*}{\wrn}
%	\mycsvreader{results/rate_score_svhn.csv}{wrn}{pgd} & 
%	\mycsvreader{results/rate_score_svhn.csv}{wrn}{trades} & 
%	\mycsvreader{results/rate_score_svhn.csv}{wrn}{mart}
	\bottomrule
\end{influencetable0}} 
\end{table}

\subsection{Ablation Study}
\label{sec:influence-factor}

\ourmethod features two important components: Learning compression
quotas and learning what connections to prune. While we do argue that
both aspects are crucial for effective adversarial robust pruning, in
this section, we set out to analyze whether this indeed is the case.
We perform an ablation study regarding the compression quota to be
learned~\rates and determining importance scores~\scores.
% , to investigate the effectiveness of either one by running
% experiments with \resnet[18] and \vgg learned on \cifar[10].
For the experiments reported in \cref{tab:rate-score}, we thus either
learn compression quotas or importance scores, while
keeping the other one constant---consequently, we also do not balance pruning
objectives here.
% .
Note that for the optimization on compression quotas~\rates, importance
scores~\scores remain as initialized by \cref{eq:scores-init-c},
% (Step~2 in~\cref{alg:harp-alg}) .
while during investigating the influence of scores~\scores, we use
uniform compression.
% That latter is essentially equivalent to \hydrax.
We denote our method with compression-rate optimization only as
\ourmethodR, and with importance-score optimization only as \ourmethodS.

As already observed in~\cref{fig:overview}, moderate compression manages
to maintain natural accuracy and adversarial robustness very well, such
that in this study, we focus on \perc{99} and \perc{99.9} sparsity.
% .
Interestingly, \ourmethodR appears to be particular beneficial for
moderate to high compression, while \ourmethodS is increasingly
important for aggressive pruning. In particular for \vgg with its
cascading network architecture, this tendency is apparent for different
pre-training strategies. However, at either sparsity level the
combination of both, that is, optimizing compression rates~\rates
\emph{and} importance scores~\scores, allows \ourmethod to excel.

\subsection{Robust Pruning with \ourmethod}
\label{sec:compare-sota}

In this section, we report results of our comparative evaluation with
\radmm~\citep{Ye2019Adversarial}, \hydra~\citep{Sehwag2020HYDRA},
\bcsp~\citep{Ozdenizci2021Bayesian}, and \mad~\citep{Lee2022MAD} in
pruning a neural network's weights. In particular, for aggressive
compression yielding high sparsity, \ourmethod excels and outperforms
prior work. For the comparison to \radmm and \hydra, we perform a
systematic evaluation with \cifar[10], \svhn, and \imagenet. For \bcsp
and \mad, we revert to the respective settings reported in the original
papers.

% Subsequently, we conduct experiments with each of the
% related approaches in their respective settings.

\begin{table}[!b]
	\centering\vspace*{-2mm}
	\caption{Comparing \ourmethod with \radmm and \hydra on 
	models learned on \cifar[10].}
	\centering
	\newcommand{\mycsvreader}[4]{%
	\csvreader[
	head to column names,
	head to column names prefix = COL,
	filter = \equal{\COLarch}{#1} \and \equal{\COLat}{#2} \and 
	\equal{\COLrate}{#3} \and \equal{\COLacc}{#4}
	]%
	{results/cifar10_weight_errorbar.csv}%
	{}
	{
		& \COLradmmbf
		& \COLradmmstd
		& \COLhydrabf
		& \COLhydrastd
		& \COLbcspbf
		& \COLbcspstd
		& \COLharpbf
		& \COLharpstd
	}
}

\newcommand{\mymidrule}{\cmidrule{2-21}}

\begin{cifarweighttable}{\linewidth}
	\midrule
	\multirow{15}{*}{\rotatebox{90}{\resnet[18]}}
	& \multirow{5}{*}{\pgd}
	%& \multirow{4}{*}{\pgd}
	& -- & \num{82.89}
	\mycsvreader{resnet}{pgd}{0.99}{nat} &
	\mycsvreader{resnet}{pgd}{0.999}{nat} \\
%	& & \fgsm
%	\mycsvreader{resnet}{pgd}{0.99}{fgsm} &
%	\mycsvreader{resnet}{pgd}{0.999}{fgsm} \\
	& & \pgd & \num{50.05}
	\mycsvreader{resnet}{pgd}{0.99}{pgd} &
	\mycsvreader{resnet}{pgd}{0.999}{pgd} \\
	& & \cwinf & \num{47.95}
	\mycsvreader{resnet}{pgd}{0.99}{cw} &
	\mycsvreader{resnet}{pgd}{0.999}{cw} \\
	& & \autopgd & \num{47.78}
	\mycsvreader{resnet}{pgd}{0.99}{autopgd} &
	\mycsvreader{resnet}{pgd}{0.999}{autopgd} \\
	& & \autoattack	& \num{45.30}
	\mycsvreader{resnet}{pgd}{0.99}{autoattack} &
	\mycsvreader{resnet}{pgd}{0.999}{autoattack} \\
	\mymidrule
	%%%%%%%%%%%%%%%%%%%%%%%
	%%%%%%%%%%%%%%%%%%%%%%%
	& \multirow{5}{*}{\trades}
	%& \multirow{4}{*}{\trades}
	& -- & \num{81.30}
	\mycsvreader{resnet}{trades}{0.99}{nat} & 
	\mycsvreader{resnet}{trades}{0.999}{nat} \\
%	& & \fgsm
%	\mycsvreader{resnet}{trades}{0.99}{fgsm} &
%	\mycsvreader{resnet}{trades}{0.999}{fgsm} \\
	& & \pgd & \num{53.21}
	\mycsvreader{resnet}{trades}{0.99}{pgd} &
	\mycsvreader{resnet}{trades}{0.999}{pgd} \\
	& & \cwinf & \num{50.60}
	\mycsvreader{resnet}{trades}{0.99}{cw} &
	\mycsvreader{resnet}{trades}{0.999}{cw} \\
	& & \autopgd & \num{51.88}
	\mycsvreader{resnet}{trades}{0.99}{autopgd} &
	\mycsvreader{resnet}{trades}{0.999}{autopgd} \\
	& & \autoattack	& \num{49.48}
	\mycsvreader{resnet}{trades}{0.99}{autoattack} &
	\mycsvreader{resnet}{trades}{0.999}{autoattack} \\
	\mymidrule
	%%%%%%%%%%%%%%%%%%%%%%%
	%%%%%%%%%%%%%%%%%%%%%%%
	& \multirow{5}{*}{\mart}
	%& \multirow{4}{*}{\mart}
	& -- & \num{80.16}
	\mycsvreader{resnet}{mart}{0.99}{nat} & 
	\mycsvreader{resnet}{mart}{0.999}{nat} \\
%	& & \fgsm
%	\mycsvreader{resnet}{mart}{0.99}{fgsm} &
%	\mycsvreader{resnet}{mart}{0.999}{fgsm} \\
	& & \pgd & \num{53.72}
	\mycsvreader{resnet}{mart}{0.99}{pgd} &
	\mycsvreader{resnet}{mart}{0.999}{pgd} \\
	& & \cwinf & \num{48.53}
	\mycsvreader{resnet}{mart}{0.99}{cw} &
	\mycsvreader{resnet}{mart}{0.999}{cw} \\
	& & \autopgd & \num{51.59}
	\mycsvreader{resnet}{mart}{0.99}{autopgd} &
	\mycsvreader{resnet}{mart}{0.999}{autopgd} \\
	& & \autoattack & \num{46.39}
	\mycsvreader{resnet}{mart}{0.99}{autoattack} &
	\mycsvreader{resnet}{mart}{0.999}{autoattack} \\
	%%%%%%%%%%%%%%%%%%%%%%%%%%%%%%%%%%%%%%%%%%%%%%%%%%%%%%%%%%%%
	%%%%%%%%%%%%%%%%%%%%%%%%%%%%%%%%%%%%%%%%%%%%%%%%%%%%%%%%%%%%
	%%%%%%%%%%%%%%%%%%%%%%%%%%%%%%%%%%%%%%%%%%%%%%%%%%%%%%%%%%%%
	\midrule
	\multirow{15}{*}{\rotatebox{90}{\vgg}}
	& \multirow{5}{*}{\pgd}
	%& \multirow{4}{*}{\pgd}
	& -- & \num{79.68}
	\mycsvreader{vgg}{pgd}{0.99}{nat} &
	\mycsvreader{vgg}{pgd}{0.999}{nat} \\
%	& & \fgsm
%	\mycsvreader{vgg}{pgd}{0.99}{fgsm} &
%	\mycsvreader{vgg}{pgd}{0.999}{fgsm} \\
	& & \pgd & \num{47.60}
	\mycsvreader{vgg}{pgd}{0.99}{pgd} &
	\mycsvreader{vgg}{pgd}{0.999}{pgd} \\
	& & \cwinf & \num{45.23}
	\mycsvreader{vgg}{pgd}{0.99}{cw} &
	\mycsvreader{vgg}{pgd}{0.999}{cw} \\
	& & \autopgd & \num{45.05}
	\mycsvreader{vgg}{pgd}{0.99}{autopgd} &
	\mycsvreader{vgg}{pgd}{0.999}{autopgd} \\
	& & \autoattack & \num{42.12}
	\mycsvreader{vgg}{pgd}{0.99}{autoattack} &
	\mycsvreader{vgg}{pgd}{0.999}{autoattack} \\
	\mymidrule
	%%%%%%%%%%%%%%%%%%%%%%%
	%%%%%%%%%%%%%%%%%%%%%%%
	& \multirow{5}{*}{\trades}
	%& \multirow{4}{*}{\trades}
	& -- & \num{80.18}
	\mycsvreader{vgg}{trades}{0.99}{nat} &
	\mycsvreader{vgg}{trades}{0.999}{nat} \\
%	& & \fgsm
%	\mycsvreader{vgg}{trades}{0.99}{fgsm} &
%	\mycsvreader{vgg}{trades}{0.999}{fgsm} \\
	& & \pgd & \num{49.72}
	\mycsvreader{vgg}{trades}{0.99}{pgd} &
	\mycsvreader{vgg}{trades}{0.999}{pgd} \\
	& & \cwinf & \num{46.59}
	\mycsvreader{vgg}{trades}{0.99}{cw} &
	\mycsvreader{vgg}{trades}{0.999}{cw} \\
	& & \autopgd & \num{47.89}
	\mycsvreader{vgg}{trades}{0.99}{autopgd} &
	\mycsvreader{vgg}{trades}{0.999}{autopgd} \\
	& & \autoattack & \num{45.09}
	\mycsvreader{vgg}{trades}{0.99}{autoattack} &
	\mycsvreader{vgg}{trades}{0.999}{autoattack} \\
	\mymidrule
	%%%%%%%%%%%%%%%%%%%%%%%
	%%%%%%%%%%%%%%%%%%%%%%%
	& \multirow{5}{*}{\mart}
	%& \multirow{4}{*}{\mart}
	& -- & \num{73.44}
	\mycsvreader{vgg}{mart}{0.99}{nat} &
	\mycsvreader{vgg}{mart}{0.999}{nat} \\
%	& & \fgsm
%	\mycsvreader{vgg}{mart}{0.99}{fgsm} &
%	\mycsvreader{vgg}{mart}{0.999}{fgsm} \\
	& & \pgd & \num{51.51}
	\mycsvreader{vgg}{mart}{0.99}{pgd} &
	\mycsvreader{vgg}{mart}{0.999}{pgd} \\
	& & \cwinf & \num{44.38}
	\mycsvreader{vgg}{mart}{0.99}{cw} &
	\mycsvreader{vgg}{mart}{0.999}{cw} \\
	& & \autopgd & \num{49.56}
	\mycsvreader{vgg}{mart}{0.99}{autopgd} &
	\mycsvreader{vgg}{mart}{0.999}{autopgd} \\
	& & \autoattack & \num{42.20}
	\mycsvreader{vgg}{mart}{0.99}{autoattack} &
	\mycsvreader{vgg}{mart}{0.999}{autoattack} \\
	\bottomrule
\end{cifarweighttable}
	\label{tab:compare-cifar-w}
\end{table}
\begin{table}[!h]
	\centering
	\caption{Comparing \ourmethod with \radmm and \hydra on 
			models learned on \svhn.}
	\newcommand{\mycsvreader}[4]{%
	\csvreader[
	head to column names,
	head to column names prefix = COL,
	filter = \equal{\COLarch}{#1} \and \equal{\COLat}{#2} \and 
	\equal{\COLrate}{#3} \and \equal{\COLacc}{#4}
	]%
	{results/svhn_weight_errorbar.csv}%
	{}
	{
		& \COLradmmbf
		& \COLradmmstd
		& \COLhydrabf
		& \COLhydrastd
		& \COLbcspbf
		& \COLbcspstd
		& \COLharpbf
		& \COLharpstd
	}
}

\newcommand{\mymidrule}{\cmidrule{2-21}}

\begin{cifarweighttable}{\linewidth}
	\midrule
	\multirow{15}{*}{\rotatebox{90}{\resnet[18]}}
	& \multirow{5}{*}{\pgd}
	%& \multirow{4}{*}{\pgd}
	& -- & \num{91.92}
	\mycsvreader{resnet}{pgd}{0.99}{nat} &
	\mycsvreader{resnet}{pgd}{0.999}{nat} \\
%	& & \fgsm
%	\mycsvreader{resnet}{pgd}{0.99}{fgsm} &
%	\mycsvreader{resnet}{pgd}{0.999}{fgsm} \\
	& & \pgd & \num{56.82}
	\mycsvreader{resnet}{pgd}{0.99}{pgd} &
	\mycsvreader{resnet}{pgd}{0.999}{pgd} \\
	& & \cwinf & \num{52.41}
	\mycsvreader{resnet}{pgd}{0.99}{cw} &
	\mycsvreader{resnet}{pgd}{0.999}{cw} \\
	& & \autopgd & \num{47.67}
	\mycsvreader{resnet}{pgd}{0.99}{autopgd} &
	\mycsvreader{resnet}{pgd}{0.999}{autopgd} \\
	& & \autoattack & \num{43.56}
	\mycsvreader{resnet}{pgd}{0.99}{autoattack} &
	\mycsvreader{resnet}{pgd}{0.999}{autoattack} \\
	\mymidrule
	%%%%%%%%%%%%%%%%%%%%%%%
	%%%%%%%%%%%%%%%%%%%%%%%
	& \multirow{5}{*}{\trades}
	%& \multirow{4}{*}{\trades}
	& -- & \num{90.70}
	\mycsvreader{resnet}{trades}{0.99}{nat} & 
	\mycsvreader{resnet}{trades}{0.999}{nat} \\
%	& & \fgsm
%	\mycsvreader{resnet}{trades}{0.99}{fgsm} &
%	\mycsvreader{resnet}{trades}{0.999}{fgsm} \\
	& & \pgd & \num{57.12}
	\mycsvreader{resnet}{trades}{0.99}{pgd} &
	\mycsvreader{resnet}{trades}{0.999}{pgd} \\
	& & \cwinf & \num{51.40}
	\mycsvreader{resnet}{trades}{0.99}{cw} &
	\mycsvreader{resnet}{trades}{0.999}{cw} \\
	& & \autopgd & \num{45.60}
	\mycsvreader{resnet}{trades}{0.99}{autopgd} &
	\mycsvreader{resnet}{trades}{0.999}{autopgd} \\
	& & \autoattack & \num{42.08}
	\mycsvreader{resnet}{trades}{0.99}{autoattack} &
	\mycsvreader{resnet}{trades}{0.999}{autoattack} \\
	\mymidrule
	%%%%%%%%%%%%%%%%%%%%%%%
	%%%%%%%%%%%%%%%%%%%%%%%
	& \multirow{5}{*}{\mart}
	%& \multirow{4}{*}{\mart}
	& -- & \num{90.84}
	\mycsvreader{resnet}{mart}{0.99}{nat} & 
	\mycsvreader{resnet}{mart}{0.999}{nat} \\
%	& & \fgsm
%	\mycsvreader{resnet}{mart}{0.99}{fgsm} &
%	\mycsvreader{resnet}{mart}{0.999}{fgsm} \\
	& & \pgd & \num{59.53}
	\mycsvreader{resnet}{mart}{0.99}{pgd} &
	\mycsvreader{resnet}{mart}{0.999}{pgd} \\
	& & \cwinf & \num{51.48}
	\mycsvreader{resnet}{mart}{0.99}{cw} &
	\mycsvreader{resnet}{mart}{0.999}{cw} \\
	& & \autopgd & \num{51.25}
	\mycsvreader{resnet}{mart}{0.99}{autopgd} &
	\mycsvreader{resnet}{mart}{0.999}{autopgd} \\
	& & \autoattack & \num{42.93}
	\mycsvreader{resnet}{mart}{0.99}{autoattack} &
	\mycsvreader{resnet}{mart}{0.999}{autoattack} \\
	%%%%%%%%%%%%%%%%%%%%%%%%%%%%%%%%%%%%%%%%%%%%%%%%%%%%%%%%%%%%
	%%%%%%%%%%%%%%%%%%%%%%%%%%%%%%%%%%%%%%%%%%%%%%%%%%%%%%%%%%%%
	%%%%%%%%%%%%%%%%%%%%%%%%%%%%%%%%%%%%%%%%%%%%%%%%%%%%%%%%%%%%
	\midrule
	\multirow{15}{*}{\rotatebox{90}{\vgg}}
	& \multirow{5}{*}{\pgd}
	%& \multirow{4}{*}{\pgd}
	& -- & \num{91.38}
	\mycsvreader{vgg}{pgd}{0.99}{nat} &
	\mycsvreader{vgg}{pgd}{0.999}{nat} \\
%	& & \fgsm
%	\mycsvreader{vgg}{pgd}{0.99}{fgsm} &
%	\mycsvreader{vgg}{pgd}{0.999}{fgsm} \\
	& & \pgd & \num{52.22}
	\mycsvreader{vgg}{pgd}{0.99}{pgd} &
	\mycsvreader{vgg}{pgd}{0.999}{pgd} \\
	& & \cwinf & \num{47.74}
	\mycsvreader{vgg}{pgd}{0.99}{cw} &
	\mycsvreader{vgg}{pgd}{0.999}{cw} \\
	& & \autopgd & \num{47.72}
	\mycsvreader{vgg}{pgd}{0.99}{autopgd} &
	\mycsvreader{vgg}{pgd}{0.999}{autopgd} \\
	& & \autoattack & \num{43.59}
	\mycsvreader{vgg}{pgd}{0.99}{autoattack} &
	\mycsvreader{vgg}{pgd}{0.999}{autoattack} \\
	\mymidrule
	%%%%%%%%%%%%%%%%%%%%%%%
	%%%%%%%%%%%%%%%%%%%%%%%
	& \multirow{5}{*}{\trades}
	%& \multirow{4}{*}{\trades}
	& -- & \num{88.66}
	\mycsvreader{vgg}{trades}{0.99}{nat} &
	\mycsvreader{vgg}{trades}{0.999}{nat} \\
%	& & \fgsm
%	\mycsvreader{vgg}{trades}{0.99}{fgsm} &
%	\mycsvreader{vgg}{trades}{0.999}{fgsm} \\
	& & \pgd & \num{52.90}
	\mycsvreader{vgg}{trades}{0.99}{pgd} &
	\mycsvreader{vgg}{trades}{0.999}{pgd} \\
	& & \cwinf & \num{47.06}
	\mycsvreader{vgg}{trades}{0.99}{cw} &
	\mycsvreader{vgg}{trades}{0.999}{cw} \\
	& & \autopgd & \num{49.87}
	\mycsvreader{vgg}{trades}{0.99}{autopgd} &
	\mycsvreader{vgg}{trades}{0.999}{autopgd} \\
	& & \autoattack & \num{44.62}
	\mycsvreader{vgg}{trades}{0.99}{autoattack} &
	\mycsvreader{vgg}{trades}{0.999}{autoattack} \\
	\mymidrule
	%%%%%%%%%%%%%%%%%%%%%%%
	%%%%%%%%%%%%%%%%%%%%%%%
	& \multirow{5}{*}{\mart}
	%& \multirow{4}{*}{\mart}
	& -- & \num{87.45}
	\mycsvreader{vgg}{mart}{0.99}{nat} &
	\mycsvreader{vgg}{mart}{0.999}{nat} \\
%	& & \fgsm
%	\mycsvreader{vgg}{mart}{0.99}{fgsm} &
%	\mycsvreader{vgg}{mart}{0.999}{fgsm} \\
	& & \pgd & \num{52.20}
	\mycsvreader{vgg}{mart}{0.99}{pgd} &
	\mycsvreader{vgg}{mart}{0.999}{pgd} \\
	& & \cwinf & \num{44.98}
	\mycsvreader{vgg}{mart}{0.99}{cw} &
	\mycsvreader{vgg}{mart}{0.999}{cw} \\
	& & \autopgd & \num{46.30}
	\mycsvreader{vgg}{mart}{0.99}{autopgd} &
	\mycsvreader{vgg}{mart}{0.999}{autopgd} \\
	& & \autoattack & \num{40.43}
	\mycsvreader{vgg}{mart}{0.99}{autoattack} &
	\mycsvreader{vgg}{mart}{0.999}{autoattack} \\
	\bottomrule
\end{cifarweighttable}
	\label{tab:compare-svhn-w}
\end{table}\vspace*{-2mm}

\paragraph{Comparison to \radmm} The method by \citet{Ye2019Adversarial}
is designed to robustly prune neural networks on fine-grained (weights)
and coarse-grained granularity (channels/filters). We focus on pruning
weights at this point and present further results on pruning channels in
\cref{app:compare-ch-radmm}.
\cref{tab:compare-cifar-w,tab:compare-svhn-w,tab:compare-imgnet}
summarize the results for pruning models on \cifar[10], \svhn, and
\imagenet, respectively.
% .
% Note, that the overall compression is comparable between granularity
% settings as the more coarse-grained structure of channel pruning, of
% course, removes more connections at once than weight pruning. Taking
% \vgg as an example, pruning to a channel sparsity of \perc{90},
% leaving the first/input layer untouched, removes slightly more than
% \perc{99} of the network's weights.
% \textcolor{red}{Pruning \perc{90} of the channels, removes~\perc{99}
% of the network's weights.} .
\radmm performs well for pruning a network's weights to a
sparsity of \perc{99} for \cifar[10] as well as \svhn. However, its
performance drops drastically when increasing sparsity to \perc{99.9}.
For \svhn, the models are even completely damaged, yielding a balanced
accuracy of \perc{10}. Also for \imagenet, \radmm significantly harms
the model robustness and natural performance.
% .
\ourmethod, in turn, exceeds the performance of \radmm distinctively.
Models pruned by our method preserve natural accuracy and robustness
significantly better for high compression rates: At a sparsity of
\perc{99}, the natural accuracy and the adversarial robustness is at a
comparable level to the original model on \cifar[10] and \svhn. Even for
aggressive pruning to a sparsity of \perc{99.9}, the models show
substantial resistance against adversarial inputs across training
methods and considered attackers.
%.
On \imagenet, \ourmethod outperforms \radmm in moderate and aggressive
pruning with \num{20} percentage points for natural accuracy and
\num{10} percentage points for robustness against various attacks.
%
%% Note that 
%% uniform strategy strongly harms the input information, for instance 
%% sparsity \perc{90} on the first layer is equivalent to remove all 
%% channels, therefore we maintain the first layer being not pruned in 
%% both \hydra and \radmm channel pruning experiments. 
%
%
%%\begin{table}[!htbp]
%%	\centering
%%	%	\caption{Comparing \ourmethod with \radmm on weight and channel 
%%	%pruning.}
%%	%	\input{tables/cifar10_results_radmm.tex}
%%	\caption{Comparing \ourmethod with \radmm on weight pruning.}
%%	\input{tables/results_radmm.tex}
%%	\label{tab:compare-radmm}
%%\end{table}
%
%%\begin{table}[!htbp]
%%	\centering\vspace*{-2mm}
%%	%	\captionsetup{width=0.8\linewidth}
%%	%	\caption{Comparing \ourmethod with \hydra on weight and channel 
%%	%pruning.}
%%	%	\input{tables/cifar10_results_hydra.tex}
%%	\caption{Comparing \ourmethod with \hydra on weight pruning.}
%%	\input{tables/results_hydra.tex}
%%	\label{tab:compare-hydra}\vspace*{-1mm}
%%\end{table}

\paragraph{Comparison to \hydra} 
%.
% While \hydra mainly focuses on weight pruning it can also be used for
% pruning filters or channels.
% As a matter of fact, the authors show that their approach drops
% significantly for aggressive filter pruning at a sparsity of \perc{90}
% to merely \perc{18.3} and \perc{16.7} of natural accuracy and
% adversarial robustness, respectively, of \vgg learned on \cifar[10],
% which we can confirm by our experiments. Moreover, observe that this
% extends to other adversarial training losses and architectures as
% well.
% The authors have overlooked that for filter pruning the output layer,
% of course, must not be pruned as otherwise entire classes can not be
% predicted. For channel pruning, the same holds true for the first
% layer~\citep{Gale2019State}.
% .

% we thus use the original implementation by \citet{Sehwag2020HYDRA} for
% weight pruning and modify it for channel pruning to preserve the first
% layer as \ourmethod does. The latter we denote as \hydrax.
% .
% Additionally, we test the regularization-based adversarial-training
% losses, \trades~\citep{Zhang2019Theoretically} and
% \mart~\citep{Wang2020MART}.
% Indeed, \hydrax can drastically improve over the results reported by
% \citet{Sehwag2020HYDRA} across the different adversarial pre-training
% options.
% % .
% Moreover, \hydra surpasses \radmm in weight pruning, most of the time,
% but stills falls behind in channel pruning.
% .
\citet{Sehwag2020HYDRA} demonstrate remarkable performance on robustly
pruning a model's weights, which we are able to confirm in our
experiments. The results are presented in
\cref{tab:compare-cifar-w,tab:compare-svhn-w,tab:compare-imgnet} for
\cifar[10], \svhn, and \imagenet.
% .
\hydra consistently improves upon \radmm for the sparsity levels
considered in our experiments. However, while \hydra maintains its
natural accuracy and robustness for a sparsity of \perc{99} well, for
\perc{99.9} its performance degrades heavily (similarly to \radmm).
% .
Differently, \ourmethod is able to maintain both accuracy metrics
significantly better for aggressive pruning. In most cases, our method
doubles the accuracy of \hydra at a sparsity of \perc{99.9}, yielding an
unprecedentedly competitive performance for such highly compressed
networks.
% .
Also, the additionally gained robustness stemming from using \martat
rather than \tradesat nicely transfers to models pruned by \ourmethod,
while \hydra cannot benefit from this advantage at the same level.
% .
For \imagenet, \ourmethod is also able to outperform \hydra noticeably
and has the highest effectiveness in suppressing the model performance
recession. However, the drop in performance from \perc{90} to \perc{99}
sparsity is comparable to \hydra and remains significant. Limiting the
decay even further for large-scale datasets remains a challenge.

\begin{table}[!t]
	\centering\vspace*{-2mm}
	\caption{Comparing \ourmethod with \radmm and \hydra on \resnet[50] 
	models for \imagenet.}
	%\newcommand{\mycsvreader}[3]{%
	%	\csvreader[
	%	head to column names,
	%	head to column names prefix = COL,
	%	filter = \equal{\COLattack}{#1} \and \equal{\COLpreg}{#2} \and 
	%	\equal{\COLprate}{#3}
	%	]%
	%	{results/imagenet_results.csv}%
	%	{}
	%	{
		%		& \COLradmmbf
		%		& \COLhydrabf
		%		& \COLharpbf
		%	}
	%}
%
%\centering
%\begin{imagenettable}{\linewidth}
%	\midrule
%%	\multirow{5}{*}{\rotatebox{90}{\resnet[50]}}
%%	% \multirow{7}{*}{\rotatebox{90}{Channel Prune}} & 
%%%	\multirow{4}{*}{\makecell{\bfseries \freeat\\ 
		%%%	\num{60.25}\,/\,\num{32.82}}}
%%	& 
%	--
%	& \num{60.25}
%	\mycsvreader{nat}{weight}{0.9} 
%	\mycsvreader{nat}{weight}{0.99}\\
%%	& 
%	\pgd
%	& \num{32.82} 	
%	\mycsvreader{pgd}{weight}{0.9} 
%	\mycsvreader{pgd}{weight}{0.99}\\
%%	& 
%	\cwinf
%	& \num{30.67}
%	\mycsvreader{cw}{weight}{0.9} 
%	\mycsvreader{cw}{weight}{0.99}\\
%%	& 
%	\autopgd
%	& \num{31.54}
%	\mycsvreader{autopgd}{weight}{0.9}
%	\mycsvreader{autopgd}{weight}{0.99}\\
%%	& 
%	\autoattack
%	& \num{28.79}
%	\mycsvreader{autoattack}{weight}{0.9}
%	\mycsvreader{autoattack}{weight}{0.99}\\
%	\bottomrule
%\end{imagenettable}

\newcommand{\mycsvreader}[3]{%
	\csvreader[
	head to column names,
	head to column names prefix = COL,
	filter = \equal{\COLattack}{#1} \and \equal{\COLpreg}{#2} \and 
	\equal{\COLprate}{#3}
	]%
	{results/imagenet_weight_errorbar.csv}%
	{}
	{
		& \COLradmmbf
		& \COLradmmstd
		& \COLhydrabf
		& \COLhydrastd
		& \COLharpbf
		& \COLharpstd
	}
}

\centering
\begin{imgneterrorbar}{\linewidth}
	\midrule
	--
	& \num{60.25}
	\mycsvreader{nat}{weight}{0.9} 
	\mycsvreader{nat}{weight}{0.99}\\
	%	& 
	\pgd
	& \num{32.82} 	
	\mycsvreader{pgd}{weight}{0.9} 
	\mycsvreader{pgd}{weight}{0.99}\\
	%	& 
	\cwinf
	& \num{30.67}
	\mycsvreader{cw}{weight}{0.9} 
	\mycsvreader{cw}{weight}{0.99}\\
	%	& 
	\autopgd
	& \num{31.54}
	\mycsvreader{autopgd}{weight}{0.9}
	\mycsvreader{autopgd}{weight}{0.99}\\
	%	& 
	\autoattack
	& \num{28.79}
	\mycsvreader{autoattack}{weight}{0.9}
	\mycsvreader{autoattack}{weight}{0.99}\\
	\bottomrule
\end{imgneterrorbar}\vspace*{-2mm}
	\label{tab:compare-imgnet}
\end{table}
\vspace*{-3mm}

\paragraph{Comparison to \bcsp}
Similar to our approach, \bcsp~\citep{Ozdenizci2021Bayesian} strives for
a non-uniform compression strategy. To do so, they introduce ``Bayesian
Connectivity Sampling'' for selecting important network
% .
elements from a global view. In contrast to \ourmethod, it however does
not follow the staged pruning-pipeline of \citet{Han2015Learning} but
achieves robust pruning via an end-to-end sparse training-procedure.
\revision{In \cref{fig:overview}, we present results of \bcsp for a \vgg
	model trained with \pgdat on \cifar[10], and explicitly investigate 
	its performance at sparsity \perc{99} and \perc{99.9} on \cifar[10] 
	and \svhn in \cref{tab:compare-cifar-w,tab:compare-svhn-w}.} While 
	it does not allow for
channel pruning, \bcsp outperforms both \hydra and \radmm in pruning
\vgg's weights, while remains inferior to \hydra in pruning \resnet[18].
% uses \pgd-10 attacks \textcolor{red}{with a perturbation strength
	% $\strength=\sfrac{8}{255}$ with step-size 
	%$\stepsize=\sfrac{2}{255}$}
% and works for weight pruning only.
% .
\ourmethod, however, surpasses \bcsp significantly with increasing model
sparsity and yields a functional model at \perc{99.9} sparsity, while
\bcsp does~not.
\revision{In~\cref{sec:stg-compare}, we analyze the strategies and
	attribute the observed collapse to a heavily pruned input layer. More
	comparative results and experiments on \resnet[18] are
	presented in \cref{app:related-work}.}
%tab:bcsp-compare} and \cref{fig:bcsp-compare-resnet}
%in the appendix, respectively.}
%
\paragraph{Comparison to \mad}
\citet{Lee2022MAD} propose to compute an adversarial saliency map of a
robustly pre-trained model via ``Kronecker-Factored Approximate
Curvature''~\citep{Martens2015KFAC} and mask out weights that are least
relevant for robustness.
Unfortunately, we have not been able to fully reproduce the \mad's
results. We thus present the numbers reported by the authors of \mad
along-side results for \ourmethod in the experimental setup described in
the original publication and adjust pre-training accordingly for
\cifar[10] and \svhn. \citet{Lee2022MAD} use a \pgd
step-size of \num{0.0069} for adversarial pre-training over \num{60}
epochs and a \num{2e-4} weight decay. At moderate compression ($<$
\perc{90} sparsity) \revision{shown in \cref{tab:mad-compare}}, \mad and
\ourmethod are on a par. Neither one harms the performance of the neural
network for \cifar[10] and \svhn.
% .
\begin{wraptable}{r}{.53\linewidth}
	\centering
	\tablesize
	\vspace*{-2mm}
	\caption{Pruning \perc{90} weights on pre-trained robust neural 
		networks via \ourmethod and \mad.}
	%\vspace*{-2mm}
	\newcolumntype{Y}{>{\centering\arraybackslash}X}

\sisetup{mode=text}
\setlength{\tabcolsep}{1.5pt}
\begin{tabular}{
		l
		l
		S[table-format=2.2]
		S[table-format=2.2]
		S[table-format=2.2]
		S[table-format=2.2]
		S[table-format=2.2]
		S[table-format=2.2]
	}
	\toprule
	{\multirow{2}{*}{\bf Model}}
	& {\multirow{2}{*}{\bf Attack}}
	& \multicolumn{3}{c}{\bf \cifar[10]}
	& \multicolumn{3}{c}{\bf \svhn}
	\\ \cmidrule(r){3-5} \cmidrule(){6-8}
	&
	& {\bf Pre-train}	& {\bf \mad}	& {\bf \ourmethod} 
	& {\bf Pre-train}	& {\bf \mad}	& {\bf \ourmethod}
	\\ \midrule
	\multirow{2}{*}{\vgg}
	& --
	& 81.32 			& 81.44			& \bf 82.10
	& 93.19				& \bf 92.41		& \bf 92.38
	\\
	& \pgd-10
	& 50.21				& \bf 51.75		& 50.94
	& 56.03				& \bf 60.42		& 58.97
	\\
	\midrule
	\multirow{2}{*}{\resnet[18]}
	& --
	& 84.22 			& 82.73			& \bf 84.76
	& 93.67				& \bf 92.63		& \bf 92.80
	\\
	& \pgd-10
	& 53.33				& \bf 52.98		& \bf 53.46
	& 59.62				& \bf 60.61		& \bf 60.37
	\\
	\bottomrule
\end{tabular}

	\label{tab:mad-compare}
	\vspace*{-3mm}
\end{wraptable}
% .
% use the provide pre-trained \vgg and \resnet[18]
% models\footnote{\url{https://github.com/ByungKwanLee/Masking-Adversarial-Damage}}
% on \cifar[10] and \svhn.
% .
% As indicated in \cref{fig:mad-compare}, showing the natural accuracy
% and adversarial robustness for both methods side-by-side, this changes
% for \perc{99} sparsity. While our method is able to still maintain
% robustness, \mad drops in performance. The natural accuracy remains at
% a comparable level. Please note, that \citet{Lee2022MAD} only report
% isolated measurement points for the natural accuracy.
\revision{As indicated in \cref{fig:mad-compare}, showing the \mbox{\pgd-10}
adversarial robustness on \cifar[10] and \svhn, this changes for
\perc{99} sparsity. While our method maintains
robustness, \mad drops in performance significantly.}
% .
In particular, it is noteworthy that at a sparsity of \perc{99.9}
\ourmethod achieves similar adversarial robustness than \mad at a
sparsity of~\perc{99}, meaning, the network pruned by \mad is an order
of magnitude larger.

%We hence yield on-par adversarial robustness for a
%a network that is compressed by an order of magnitude stronger.

\begin{figure}[h]
  %\vspace*{-2mm}
  \centering
	% \captionsetup{font={small}, width=\linewidth}
% \caption{\revision{Comparing \ourmethod with \radmm (blue), \hydra
% (green) and \mad (gray). All pre-trained robust models are originally
% from \mad. Solid lines and dotted lines represent the natural accuracy
% and \pgd-10 adversarial robustness respectively on \cifar[10].}}
  \caption{\revision{Comparing \pgd-10 adversarial robustness 
  of \radmm, \hydra, \mad, and \ourmethod on \vgg models learned on 
  \cifar[10]. All pre-trained models are provided by \citet{Lee2022MAD}.}}
  \vspace*{-1mm}
  \begin{subfigure}[h]{.45\linewidth}
    \centering
    \caption{\cifar[10]}
    \begin{tikzpicture}
\begin{semilogxaxis}[
%log ticks with fixed point,
height=0.5\linewidth,
width=0.8\linewidth,
legend pos=outer north east,
enlarge x limits=0.02,
enlarge y limits=0.05,
grid=both,
grid style={dashed,gray!30},
smooth,
tension=0.0,
xlabel={Sparsity},
ylabel={Accuracy [\%]},
xmin=0.001,
xmax=1,
ymin=30,
ymax=60,
x dir=reverse,
xtick={0.001, 0.01, 0.1, 1.0},
ytick={30, 40, 50, 60}, 
xticklabels={99.9\%, 99\%, 90\%, 0\%},
yticklabels={30, 40, 50, 60},
yticklabel style = {font=\fontsize{6}{6}\selectfont},
xticklabel style = {font=\fontsize{6}{6}\selectfont},
ylabel style = {font=\fontsize{7}{7}\selectfont, yshift=-4ex},
xlabel style = {font=\fontsize{7}{7}\selectfont, yshift=1ex},
%x tick label as interval
scale only axis,
legend image post style={scale=0.6},
legend style={at={(0.02, 0.05)}, font=\fontsize{9}{9}\selectfont, 
anchor=south west, 
legend columns=2, fill=white, draw=none, nodes={scale=0.6, transform 
shape}, column sep=2pt, line width=0.8pt},
legend cell align={left}
]

%%%%%%%%%%%%% VGG16 %%%%%%%%%%%%%

\addplot [mark=square*, color=secondarycolor, mark 
options={scale=0.6, solid}, line 
width=0.8pt]
coordinates {
	(1, 50.2)
	(0.7, 50.9)
	(0.5, 50.8)
	(0.3, 50.7)
	(0.1, 50.9)
	(0.01, 49.7)
	(0.005, 48.4)
	(0.001, 40.1)
}; \addlegendentry{\ourmethod}

\addplot [mark=diamond*, color=madcolor, mark 
options={scale=0.6, solid}, line width=0.8pt]
coordinates {
	(1,	50.2)
	(0.7, 51.5)
	(0.5, 51.1)
	(0.3, 51.5)
	(0.1, 51.8)
	(0.01, 40.5)
}; \addlegendentry{\mad}

%\addplot [mark=square*, color=secondarycolor, mark 
%options={scale=0.6, solid}, line 
%width=0.8pt]
%coordinates {
%	(0.001, 81.3)
%	(0.7, 81.7)
%	(0.5, 81.9)
%	(0.3, 82.1)
%	(0.1, 82.1)
%	(0.1, 79.1)
%%	(5, 76.5)
%}; %\addlegendentry{\ourmethod \vgg}

%\addplot [only marks, mark=diamond*, color=madcolor, 
%mark options={scale=0.6, solid}, line width=0.8pt]
%coordinates {
%	(0.001,	81.3)
%	(0.1, 81.4)
%}; %\addlegendentry{\mad \vgg}

\addplot [mark=*, color=tertiarycolor, mark 
options={scale=0.6, solid}, line width=0.8pt]
coordinates {
	(1, 50.2)
	(0.7, 50.2)
	(0.5, 50.2)
	(0.3, 49.9)
	(0.1, 47.4)
	(0.01, 37.8)
}; \addlegendentry{\hydra}

\addplot [mark=triangle*, color=primarycolor, mark 
options={scale=0.6, solid}, line width=0.8pt]
coordinates {
	(1, 50.2)
	(0.7, 48.4)
	(0.5, 48.9)
	(0.3, 48.3)
	(0.1, 47.4)
	(0.01, 33.7)
}; \addlegendentry{\radmms}

\addplot [mark=square*, color=secondarycolor, mark 
options={scale=0.6, solid}, line 
width=0.8pt]
coordinates {
	(1, 50.2)
	(0.7, 50.9)
	(0.5, 50.8)
	(0.3, 50.7)
	(0.1, 50.9)
	(0.01, 49.7)
	(0.005, 48.4)
	(0.001, 40.1)
}; %\addlegendentry{\ourmethod}

\end{semilogxaxis}

\end{tikzpicture}
  \end{subfigure}
  \hfill
  \begin{subfigure}[h]{.45\linewidth}
    \centering
    \caption{\svhn}
    \begin{tikzpicture}
\begin{semilogxaxis}[
%log ticks with fixed point,
height=0.5\linewidth,
width=0.8\linewidth,
legend pos=outer north east,
enlarge x limits=0.02,
enlarge y limits=0.08,
grid=both,
grid style={dashed,gray!30},
smooth,
tension=0.0,
xlabel={Sparsity},
ylabel={Accuracy [\%]},
xmin=0.001,
xmax=1,
ymin=30,
ymax=60,
x dir=reverse,
xtick={0.001, 0.01, 0.1, 1.0},
ytick={30, 40, 50, 60}, 
xticklabels={99.9\%, 99\%, 90\%, 0\%},
yticklabels={30, 40, 50, 60},
yticklabel style = {font=\fontsize{6}{6}\selectfont},
xticklabel style = {font=\fontsize{6}{6}\selectfont},
ylabel style = {font=\fontsize{7}{7}\selectfont, yshift=-4ex},
xlabel style = {font=\fontsize{7}{7}\selectfont, yshift=1ex},
%x tick label as interval
scale only axis,
legend image post style={scale=0.6},
legend style={at={(0.02, 0.05)}, font=\fontsize{9}{9}\selectfont, 
	anchor=south west, 
	legend columns=2, fill=white, draw=none, nodes={scale=0.6, transform 
		shape}, column sep=2pt, line width=0.8pt},
legend cell align={left}
]

%%%%%%%%%%%%% VGG16 %%%%%%%%%%%%%

%\addplot [mark=square*, color=secondarycolor, mark 
%options={scale=0.6, solid}, line 
%width=0.8pt]
%coordinates {
%	(1, 93.2)
%	(1.43, 92.3)
%	(2, 92.4)
%	(3.33, 92.2)
%	(10, 92.4)
%	(100, 91.5)
%	(500, 90.3)
%}; \addlegendentry{\ourmethod}

%\addplot [mark=diamond*, color=madcolor, 
%mark options={scale=0.6, solid}, line width=0.8pt]
%coordinates {
%	(10,	92.4)
%	(10, 92.4)
%}; \addlegendentry{\mad}

%\addplot [only marks, mark=diamond*, color=madcolor, 
%mark options={scale=0.6, solid}, line width=0.8pt]
%coordinates {
%	(1,	93.2)
%	(10, 92.4)
%};

\addplot [mark=square*, color=secondarycolor, mark 
options={scale=0.6, solid}, line 
width=0.8pt]
coordinates {
	(1, 56.0)
	(0.7, 58.4)
	(0.5, 58.2)
	(0.3, 58.3)
	(0.1, 58.5)
	(0.01, 56.2)
	(0.005, 55.1)
	(0.001, 47.3)
}; \addlegendentry{\ourmethod}

\addplot [mark=diamond*, color=madcolor, mark 
options={scale=0.6, solid}, line width=0.8pt]
coordinates {
	(1,	56.0)
	(0.7, 58.9)
	(0.5, 58.5)
	(0.3, 58.8)
	(0.1, 58.4)
	(0.01, 51.8)
}; \addlegendentry{\mad}

\addplot [mark=*, color=tertiarycolor, mark 
options={scale=0.6, solid}, line width=0.8pt]
coordinates {
	(1, 50.2)
	(0.7, 50.2)
	(0.5, 50.2)
	(0.3, 49.9)
	(0.1, 47.4)
	(0.01, 37.8)
}; \addlegendentry{\hydra}

\addplot [mark=triangle*, color=primarycolor, mark 
options={scale=0.6, solid}, line width=0.8pt]
coordinates {
	(1, 50.2)
	(0.7, 48.4)
	(0.5, 48.9)
	(0.3, 48.3)
	(0.1, 47.4)
	(0.01, 33.7)
}; \addlegendentry{\radmms}

\addplot [mark=square*, color=secondarycolor, mark 
options={scale=0.6, solid}, line 
width=0.8pt]
coordinates {
	(1, 56.0)
	(0.7, 58.4)
	(0.5, 58.2)
	(0.3, 58.3)
	(0.1, 58.5)
	(0.01, 56.2)
	(0.005, 55.1)
	(0.001, 47.3)
}; %\addlegendentry{\ourmethod}

\end{semilogxaxis}

\end{tikzpicture}
  \end{subfigure}
  \label{fig:mad-compare}
  \vspace*{-2mm}
\end{figure}
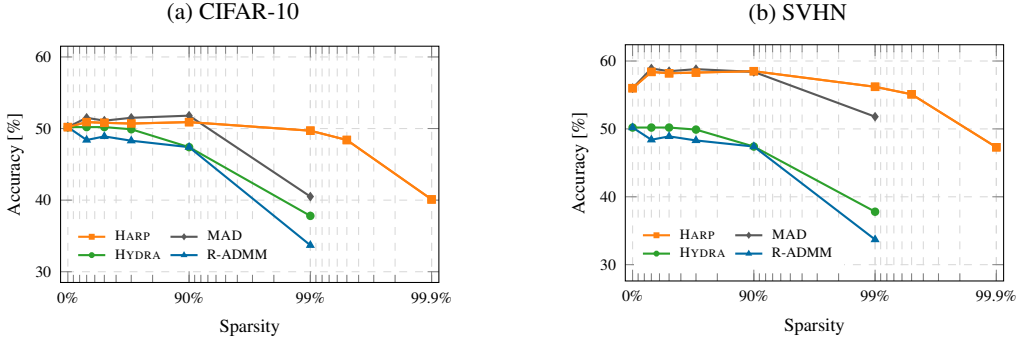
\begin{table}[!b]
	\caption{Comparing performance improvement of \radmm and \hydra by 
		using \erk and \lamp and by \ourmethod on \cifar[10]. 
		Natural accuracy and \pgd-10 robustness are presented left 
		and	right of the \texttt{/} character.}
	\tablesize
	\newcommand{\sotareader}[3]{%
	\csvreader[
	head to column names,
	filter = \equal{\method}{#1} \and \equal{\arch}{#2} \and 
	\equal{\prate}{#3},
	]%
	{results/lamp_erk_compare_cifar10.csv}%
	{}
	{	
		& \unioabf
		& \uniaabf
		& \erkoabf
		& \erkaabf
		& \lampoabf
		& \lampaabf
	}
}

\newcommand{\harpreader}[3]{%
	\csvreader[
	head to column names,
	filter = \equal{\method}{#1} \and \equal{\arch}{#2} \and 
	\equal{\prate}{#3},
	%	late after line=\\
	]%
	{results/lamp_erk_compare_cifar10.csv}%
	{}
	{	
		& \harpoabf
		& \harpaabf
	}
}

\begin{stgtable}{\linewidth}
	\midrule
	\multirow{2}{*}{\resnet[18]}
	& \perc{99}
	\sotareader{radmm}{resnet}{0.01}
	\sotareader{hydra}{resnet}{0.01}
	\harpreader{hydra}{resnet}{0.01}
	\\
	& \perc{99.9}
	\sotareader{radmm}{resnet}{0.001}
	\sotareader{hydra}{resnet}{0.001}
	\harpreader{hydra}{resnet}{0.001}
	\\ \midrule
	\multirow{2}{*}{\vgg}
	& \perc{99}
	\sotareader{radmm}{vgg}{0.01}
	\sotareader{hydra}{vgg}{0.01}
	\harpreader{hydra}{vgg}{0.01}
	\\
	& \perc{99.9}
	\sotareader{radmm}{vgg}{0.001}
	\sotareader{hydra}{vgg}{0.001}
	\harpreader{hydra}{vgg}{0.001}
	\\
	\bottomrule
\end{stgtable}
	\label{tab:stg-compare}
\end{table}

\subsection{Strategy Analysis}
\label{sec:stg-compare}

%\begin{wraptable}{r}{0.47\linewidth}
%  \vspace*{-4mm}
%  \tablesize
%  \input{tables/stg-compare.tex}
%  \vspace*{-2mm}
%\end{wraptable}

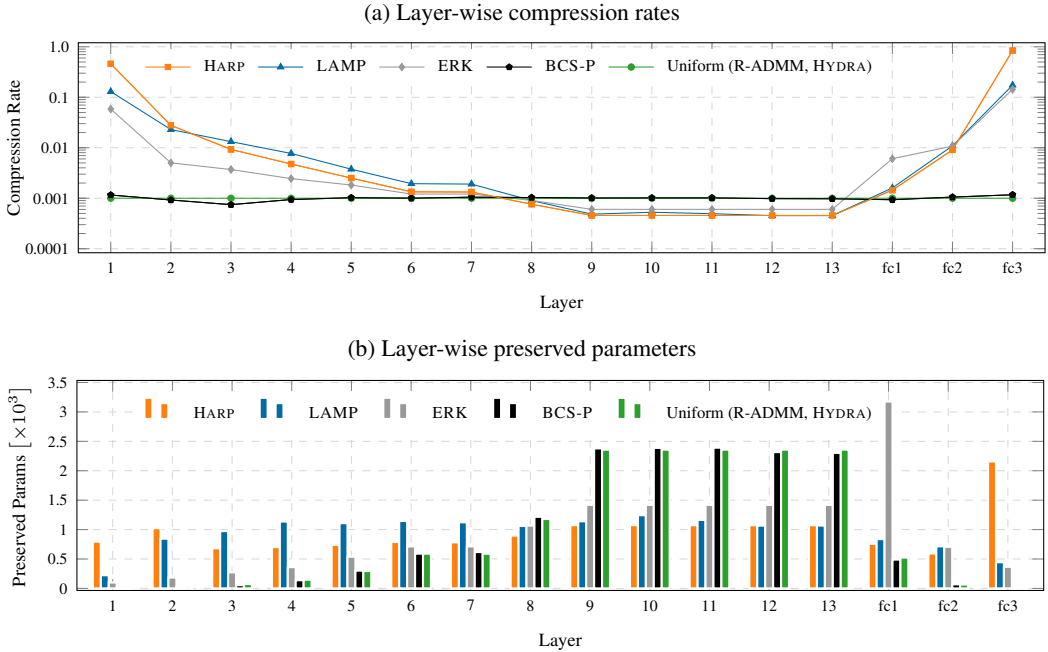
\begin{figure}[!h]
	%	\caption{\revision{Strategies of \ourmethod, \lamp, and \erk for 
			%			\vgg with a \perc{99.9} sparsity on \cifar[10].}}
	\caption{\revision{Strategy comparison for pruning a
			\vgg with target \perc{99.9} sparsity on \cifar[10].}}
	\vspace*{-1mm}
	\begin{subfigure}[h]{\linewidth}
		\centering
		\caption{Layer-wise compression rates}
		\vspace*{-1.5mm}
		\pgfplotstableread[col sep=space]{%
layer	uniform0001	erk0001	lamp0001	harp0001	bcsp0001
1	0.0010	0.0584490895	0.1296296120	0.4594291151	0.0011574074
2	0.0010	0.0050184727	0.0229220986	0.0277645625	0.0009223090
3	0.0010	0.0037028193	0.0132107139	0.0092659546	0.0007459852
4	0.0010	0.0024482012	0.0077107549	0.0047599999	0.0009426541
5	0.0010	0.0018242598	0.0037604570	0.0025094135	0.0010240343
6	0.0010	0.0012105107	0.0019429326	0.0013384445	0.0010019938
7	0.0010	0.0012105107	0.0019056797	0.0013269510	0.0010477702
8	0.0010	0.0009045005	0.0008994341	0.0007611368	0.0010316637
9	0.0010	0.0006018877	0.0004836321	0.0004555447	0.0010087755
10	0.0010	0.0006018877	0.0005272627	0.0004556927	0.0010125902
11	0.0010	0.0006018877	0.0004937649	0.0004552769	0.0010138618
12	0.0010	0.0006018877	0.0004522800	0.0004552965	0.0009820726
13	0.0010	0.0006018877	0.0004526973	0.0004554314	0.0009765625
14	0.0010	0.0060596466	0.0015983582	0.0014460140	0.0009346008
15	0.0010	0.0107727051	0.0109100342	0.0090687536	0.0010528564
16	0.0010	0.1433593631	0.1738281250	0.8430150747	0.0011718750
}\vggstg

\begin{tikzpicture}
\begin{semilogyaxis}[
log ticks with fixed point,
height=0.2\textwidth,
width=0.92\textwidth,
legend pos=outer north east,
enlarge x limits=0.036,
enlarge y limits=0.02,
grid=major, 
grid style={dashed,gray!30},
xlabel={Layer},
ylabel={Compression Rate},  % $\left[ \times 10^3 \right]$
xmin=1,
xmax=16,
ymin=0.0001,
ymax=1.0,
xtick={1,2,...,16},
ytick={0.0001, 0.001, 0.01, 0.1, 1.0},
xticklabels={1,2,3,4,5,6,7,8,9,10,11,12,13, fc1, fc2, fc3},
yticklabels={0.0001, 0.001, 0.01, 0.1, 1.0},
yticklabel style = {font=\tiny, yshift=0.0ex},
xticklabel style = {font=\tiny},
ylabel style = {font=\fontsize{7}{7}\selectfont, yshift=-2.5ex},
xlabel style = {font=\fontsize{7}{7}\selectfont, yshift=0.5ex},
scale only axis,
legend style={at={(0.06,0.97)}, font=\legendsize, anchor=north 
west, 
legend columns=5, fill=white, draw=white, 
nodes={scale=0.9, 
transform shape}, column sep=3pt},
legend cell align={left}
]

\addplot [mark=square*, color=secondarycolor, mark 
options={scale=0.5}] table[x index = 0, y index=4] {\vggstg};
% harp

\addplot [mark=triangle*, color=primarycolor, mark 
options={scale=0.7}] table[x index = 0, y index=3] {\vggstg};
% lamp_0001

\addplot [mark=diamond*, color=black!40, mark 
options={scale=0.7}] table[x index = 0, y index=2] {\vggstg};
% erk_0001

\addplot [mark=pentagon*, color=black, mark 
options={scale=0.6}] table[x index = 0, y index=5] {\vggstg};
% bcsp_0001

\addplot [mark=*, color=tertiarycolor, mark 
options={scale=0.6}] table[x index = 0, y index=1] {\vggstg};
% uniform_0001

\addplot [mark=pentagon*, color=black, mark 
options={scale=0.6}] table[x index = 0, y index=5] {\vggstg};
% bcsp_0001

\addplot [mark=square*, color=secondarycolor, mark 
options={scale=0.5}] table[x index = 0, y index=4] {\vggstg};
% harp

%\addlegendentry{\ourmethod};
%\addlegendentry{\lamp};
%\addlegendentry{\erk};
\legend{\ourmethod, \lamp, \erk, \bcsp, Uniform~\text{(\radmms, \hydra)}}

\end{semilogyaxis}
\end{tikzpicture}
		\label{fig:stg-overview-vgg-cr}\vspace*{-2mm}
	\end{subfigure}
	\vfill
	\begin{subfigure}[h]{\linewidth}
		\centering
		\caption{Layer-wise preserved parameters}
		\vspace*{-1.5mm}
		\pgfplotstableread[col sep=space]{%
layer	uni0001	erk0001	lamp0001	harp0001	bcsp0001
1	2	101	224	794	2
2	37	185	845	1024	34
3	74	273	974	683	55
4	147	361	1137	702	139
5	295	538	1109	740	302
6	590	714	1146	789	591
7	590	714	1124	783	618
8	1180	1067	1061	898	1217
9	2359	1420	1141	1075	2380
10	2359	1420	1244	1075	2389
11	2359	1420	1165	1074	2392
12	2359	1420	1067	1074	2317
13	2359	1420	1068	1074	2304
14	524	3177	838	758	490
15	66	706	715	594	69
16	3	367	445	2158	3
}\vggparams

\begin{tikzpicture}
\begin{axis}[
ybar=0pt,
height=0.2\textwidth,
width=0.92\textwidth,
legend pos=outer north east,
enlarge x limits=0.04,
enlarge y limits=0.01,
bar width=3,
grid=major, 
grid style={dashed,gray!30},
xlabel={Layer},
ylabel={Preserved Params $\left[ \times 10^3 \right]$},
xmin=1,
xmax=16,
ymin=0.0,
ymax=3500,
xtick={1,2,...,16},
xtick align=inside,
ytick={0, 500, 1000, 1500, 2000, 2500, 3000, 3500},
xticklabels={1,2,3,4,5,6,7,8,9,10,11,12,13, fc1, fc2, fc3},
yticklabels={0, 0.5, 1, 1.5, 2, 2.5, 3, 3.5},
yticklabel style = {font=\tiny, yshift=0.0ex},
xticklabel style = {font=\tiny},
ylabel style = {font=\fontsize{7}{7}\selectfont, yshift=-3ex},
xlabel style = {font=\fontsize{7}{7}\selectfont, yshift=0.5ex},
scale only axis,
legend style={at={(0.06,0.95)}, font=\legendsize, anchor=north 
west, 
legend columns=5, fill=white, draw=white, 
nodes={scale=0.9, 
transform shape}, column sep=7.3pt},
legend cell align={left}
]

\addplot [draw=white, thick, fill=secondarycolor, mark 
options={scale=0.5}] table[x index = 0, y index=4] {\vggparams};
% harp

\addplot [draw=white, thick, fill=primarycolor, mark 
options={scale=0.5}] table[x index = 0, y index=3] {\vggparams};
% lamp_0001

\addplot [draw=white, thick, fill=black!40, thick, mark 
options={scale=0.5}] table[x index = 0, y index=2] {\vggparams};
% erk_0001

\addplot [draw=white, thick, fill=black, mark 
options={scale=0.5}] table[x index = 0, y index=5] {\vggparams};
% bcsp

\addplot [draw=white, thick, fill=tertiarycolor, thick, mark 
options={scale=0.5}] table[x index = 0, y index=1] {\vggparams};
% uni_0001

%\addlegendentry{\ourmethod};
%\addlegendentry{\lamp};
%\addlegendentry{\erk};
\legend{\ourmethod, \lamp, \erk, \bcsp, Uniform~\text{(\radmms, \hydra)}}
	\end{axis}
\end{tikzpicture}
		\label{fig:stg-overview-vgg}\vspace*{-2mm}
	\end{subfigure}
	\vspace*{-5mm}
	\label{fig:compare-cifar10-stg}
\end{figure}

Also in conventional network pruning, non-uniform compression strategies
have been proven effective, for instance, \erk by
\citet{Evci2020Rigging} and \lamp by \citet{Lee2021Layer}. While we are
the first to show an equivalent benefit for adversarial robust pruning,
in this section, we investigate (a)~if these approaches can also be
applied to adversarial robust pruning, and (b)~if they find similar
compression strategies in comparison to \ourmethod.
% .
Both \erk and \lamp focus on model parameters and support weight pruning
only. We adjust \hydra to accept different compression rates per layer
and assign the strategies determined by \erk and \lamp. The results for
pruning a \pgdat pre-trained \vgg model are presented in
\cref{tab:stg-compare}. \lamp in particular produces very promising
compression strategies that, however, are surpassed by \ourmethod.
%.
%An extension to channel pruning, unfortunately, is not trivial.
For \imagenet learned with \resnet[50], in turn, using \lamp and \erk
with \hydra impacts performance negatively, unlike \ourmethod
\mbox{(\cf \cref{tab:compare-imgnet-stg}} in \cref{app:stganalyse}).
% .
In \cref{fig:compare-cifar10-stg}, we present the non-uniform
compression strategies determined by the different approaches as (a)
compression rates and (b) overall preserved parameters per layer.
Additionally, we show a uniform compression yielding \perc{99.9}
sparsity as reference. Note, that due to the layer's difference in size
the absolute number of preserved parameters varies for the uniform
strategy in \cref{fig:stg-overview-vgg} also.
% .
It is interesting to see that \lamp puts more focus on layers in the
middle sacrificing input/output information, while \ourmethod tends to
focus on front and back layer stronger. \erk, in turn, even maintains
large portions of \code{fc1}.
% .
\revision{Interestingly, \bcsp yields an almost uniform strategy
although it strives for non-uniformity. Even worse, only two out of
\num{1728} parameters in \vgg's input layer are preserved at \perc{99.9}
sparsity, severely impacting performance.}

%
% them to further investigate the weight pruning performance \hydra. The
% table on the right side summarized the results of \hydra weight pruning
% of \vgg on \cifar[10] by using \erk strategies and \lamp strategies.
% Results exhibits the necessity of non-uniformity in \hydra.
% Whereas, \ourmethod remains the prominant pruning performance.
% \cref{fig:stg-overview} visualizes the distribution of parameters
% preservation in different strategies. \erk preserves model parameters in
% the middle layers. Differently, \lamp and \ourmethod generates a rather
% uniform stratetgy, where, however, \ourmethod intends to preserve more
% input and output information, i.e. less compression on first and last
% layer. In contrast, both \lamp and \erk preserves more middle parameters
% and sacrifices more on input/output information.

%\begin{figure}[tbh]
%	\caption{Compression strategies of \ourmethod, \lamp, and \erk for 
%pruning 
%		weights a \perc{99.9} sparsity.}% of \pgdat pre-trained \vgg 
%		%model.} 
%	\input{figures/compare-lamp-erk.tex}
%	\label{fig:stg-overview}\vspace*{-2mm}
%\end{figure}

\section{Conclusions}

For pruning neural networks it is crucial to decide \emph{how many and
which parameters to prune}. This does not only affect natural accuracy,
but also the robustness against adversarial input manipulations.
Maintaining adversarial robustness is crucial for safety-critical
applications, such as autonomous driving and edge AI.
% .
Our method, \ourmethod, incorporates a global view of the network's
compression and scores for gauging the importance of network connections
into a dynamically regularized loss formulation that allows to
significantly outperform related work. We are the first to reach
competitive performance for highly aggressive pruning, aiming at
%\perc{90} channel sparsity and 
up to \perc{99.9} network sparsity.
% .
Therewith, we show that learning a global, but layer-specific and, thus,
non-uniform compression strategy is at least as important as deciding on
what connections to prune.

\paragraph{Limitations} Similarly to the majority of related work in the
field, our method is of empirical nature and, thus, can benefit from
future work on theoretic analyses of how and why our pruning approach
arrives at a particular compression strategy. In our evaluation, we show
that \ourmethod's compression is close to the theoretically founded
strategies of \lamp, raising the hope that a rigorous theoretic
justification is possible.
% .
Ideally this extends to a theoretically determined, one-shot calculation
that spares the computational effort of the optimization problem
considered for \ourmethod.

\clearpage

\section*{Acknowledgements} The authors thank the anonymous reviewers
for their valuable suggestions, and gratefully acknowledge funding by
the Helmholtz Association (HGF) within topic ``46.23 Engineering Secure
Systems" and by SAP S.E. under project DE-2020-021.

\section*{Ethics Statement} Deep neural networks are used in a wide
variety of applications. While these applications may have ethical
implications, our method itself does not. HARP aims to compress an
existing model, maintaining accuracy and resistance against input
manipulation attacks. As such, our method is agnostic to the underlying
application. Moreover, we do not reveal any vulnerabilities in the
robust model compression, do not use any personalized data, and do not
involve human subjects in our~experiments.

% Deep neural networks are widely adopted in many mission-critical areas 
% while facing challenges of both the hardware resource constraint and the 
% threat from adversarial attacks. In this paper, we propose \ourmethod to 
% address the preservation of adversarial robustness in holistic model 
% pruning. This work has no ethical issues since our method aims to 
% maintain the resistance against adversarial threats in the model 
% compression and we do not reveal any new vulnerabilities in the robust 
% model compression. However, all considered robust pruning methods 
% including \ourmethod present the gap of highest reachable compression 
% sparsity between harmlessly pruning a naturally and an adversarially 
% pre-trained model. Future work is worthy to investigate the reason for 
% the existence of such gap.

\section*{Reproducibility Statement}
For the sake of reproducibility and to foster future research, we make
the implementations of \ourmethod for holistic adversarially robust
pruning publicly available~at:
% .
\begin{center}
	\projecturl\\
\end{center}

%Additionally, we provide the source code in the supplementary package 
%in subfolder \code{harp-prune/}. 

\bibliography{bib/references.bib}
\bibliographystyle{iclr2023_conference}

%%%%%%%%%%%%%%%%%%%%%%%%%%%%%%%%%%%%%%%%%%%%%%%%%%%%%%%%%%%%

\clearpage
\appendix
\section{Appendix}

\revision{We proceed to analyze the influence of different step sizes to
regularize \ourmethod (\cref{sec:gamma-adapt}), before we extend upon
the empirical evidence of our methods performance:
We broaden our comparison to related work by inspecting the performance
of \bcsp and \mad on \resnet[18] and additionally considering
DNR~(\cref{app:related-work})}. We also extend upon
the analysis of \ourmethod's pruning strategies (\cref{app:stganalyse}),
inspect the parameter distribution of the pruned models
(\cref{app:paramsdist}, and evaluate \ourmethod's performance based on
natural training (\cref{app:natprune}).
% .
% Then, we broaden the strategy analysis by considering the influence of
% non-uniform strategies on \ourmethod's initialization and their impact
% on the end-to-end training process and the learning for \imagenet.
% We analyze the layer-wise parameter distribution of \vgg pruned by
% \hydra, \radmm,~and~\ourmethod, .
% .
Moreover, we demonstrate an extension of our method to
channel/structural pruning (\cref{app:chprune}) \revision{ and present a
comparison of training consumption for all methods in our evaluation
(\cref{app:runtime}).
}

\begin{table}[t]
	\centering
	\caption{\revision{Overview of used method abbreviations.}}
	\tablesize
	\newcolumntype{Y}{>{\centering\arraybackslash}X}
\newcommand{\mymidrule}{\cmidrule(lr){1-2}\cmidrule{3-5}\cmidrule(l){6-6}}

\setlength{\tabcolsep}{8pt}
\begin{tabular}{lll}
    \toprule
    \bf Category & \bf Used Abbreviation  & \bf Reference \\
    \midrule
    \multirow{4}{*}{Attacks}
    & \pgd                                 & \citet{Madry2018Towards}         \\
    & \cwinf                               & \citet{Carlini2017Towards}        \\
    & \autopgd, \Autopgd                   & \citet{Croce2020AutoAttack}      \\
    & \autoattack, \Autoattack             & \citet{Croce2020AutoAttack}      \\
    \midrule
    \multirow{3}{*}{Adversarial Training}
    & \pgdat                               & \citet{Madry2018Towards}         \\
    & \tradesat                            & \citet{Zhang2019Theoretically}   \\
    & \martat                              & \citet{Wang2020MART}             \\
    \midrule
    \multirow{2}{*}{Non-uniform compression}
    & \erk                                 & \citet{Evci2020Rigging}          \\
    & \lamp                                & \citet{Lee2021Layer}             \\
    \midrule
    \multirow{6}{*}{Adversarially robust pruning}
    & \hydra                               & \citet{Sehwag2020HYDRA}          \\
    & \radmm, \radmms                      & \citet{Ye2019Adversarial}        \\
    & \bcsp                                & \citet{Ozdenizci2021Bayesian}    \\
    & \mad                                 & \citet{Lee2022MAD}               \\
    & DNR                                  & \citet{Kundu2021DNR}             \\
    & \ourmethod (Ours)                    & \citet{Zhao2023Holistic}         \\
    \bottomrule
\end{tabular}

	\label{tab:abbr-summary}
\end{table}

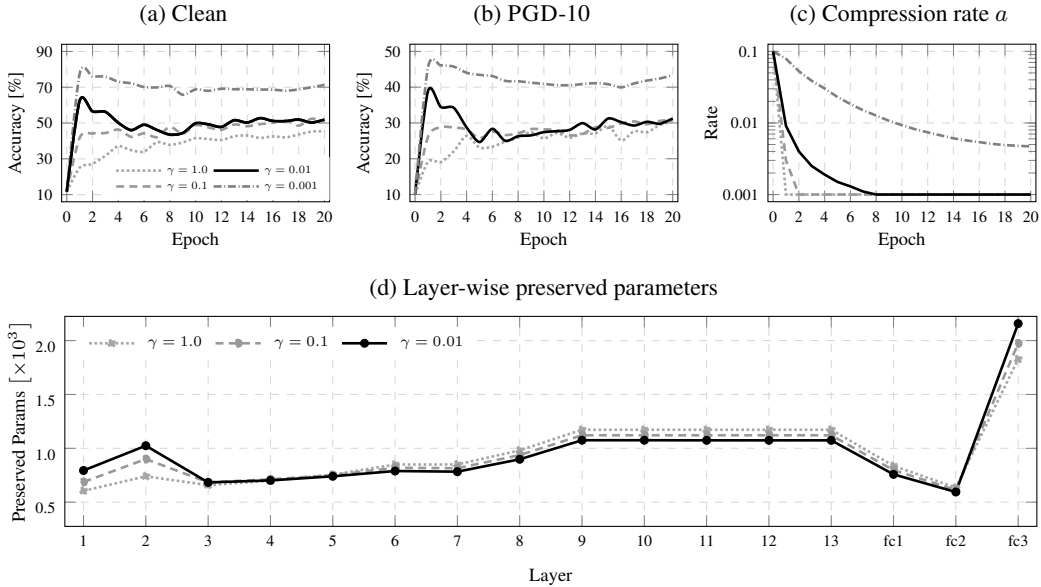
\begin{figure}[!b]
	\caption{\revision{Different $\gamma$ step-size for pruning
			\perc{99.9} weights of a \vgg by \ourmethod on \cifar[10].}}
	\label{fig:gamma-adapt}
	\begin{subfigure}{0.3\linewidth}
		\centering
		\captionsetup{margin={7mm,0mm}}
		\caption{Clean}
		\vspace*{-1mm}
		\pgfplotstableread[col sep=space]{%
epoch	g1000	g0100	g0010	g0001
0	11.5800	11.5800	11.58	11.58
1	25.0400	41.7900	62.46	77.49
2	27.2100	44.1700	56.51	76.09
3	31.6900	44.4100	56.25	75.87
4	37.0800	46.3800	50.17	73.12
5	34.8100	42.3000	46.03	72.20
6	34.0400	44.2500	49.06	70.11
7	39.3000	42.1200	46.13	69.91
8	37.8100	47.9100	43.53	70.66
9	39.1700	44.0500	44.42	65.79
10	41.5300	48.8100	49.83	68.78
11	41.0500	47.5300	49.23	67.98
12	40.5100	46.2000	47.83	69.17
13	42.6300	49.1700	51.58	68.88
14	43.0700	48.2000	50.24	68.86
15	41.6800	49.3800	52.74	68.70
16	42.5300	50.1600	51.16	68.67
17	42.0200	51.5400	51.08	68.10
18	43.5600	48.5200	51.93	68.80
19	45.2900	52.4100	50.27	70.01
20	45.3200	50.8000	51.93	71.27
}\gammaacc

\begin{tikzpicture}
	\begin{axis}[
		height=0.5\linewidth,
		width=0.85\linewidth,
		legend pos=outer north east,
		enlarge x limits=0.02,
		enlarge y limits=0.05,
		grid=major, 
		grid style={dashed,gray!30},
		xlabel={Epoch},
		ylabel={Accuracy [\%]},  % $\left[ \times 10^3 \right]$
		xmin=0,
		xmax=20,
		ymin=10,
		ymax=90,
		xtick={0,2,...,20},
		ytick={10, 30, ..., 90},
		xticklabels={0,2,4,6,8,10,12,14,16,18,20},
		yticklabels={10, 30, 50, 70, 90},
		yticklabel style = {font=\tiny, yshift=0.0ex},
		xticklabel style = {font=\tiny},
		ylabel style = {font=\fontsize{7}{7}\selectfont, yshift=-4ex},
		xlabel style = {font=\fontsize{7}{7}\selectfont, yshift=1.5ex},
		scale only axis,
		smooth,
		tension=0.5,
		legend style={at={(1.0,0.01)}, font=\legendsize, anchor=south 
		east, legend columns=2, fill=white, draw=white, 
		nodes={scale=0.6, transform shape}, column sep=0.5pt},
		legend cell align={left}
		]
		
		\addplot [mark=none, densely dotted, color=gray!70, mark 	
		options={scale=0.2}, line width=1.0pt] table[x index = 0, y 
		index=1] {\gammaacc};
		% 1.0
		
%		\addplot [mark=none, densely dotted, color=red, mark 
%		options={scale=0.5}, line width=1.0pt] table[x index = 0, y 
%		index=2] {\gammaacc};
%		% 0.5
%		
%		\addplot [mark=none, densely dash dot, color=red, mark 
%		options={scale=0.5}, line width=1.0pt] table[x index = 0, y 
%		index=3] {\gammaacc};
%		% 0.2

		\addplot [mark=none, color=black, mark 
		options={scale=0.5}, line width=1.0pt] table[x index = 0, y 
		index=3] {\gammaacc};
		% 0.01
		
		\addplot [mark=none, densely dashed, color=gray!80, mark 
		options={scale=0.5}, line width=1.0pt] table[x index = 0, y 
		index=2] {\gammaacc};
		% 0.1
		
%		\addplot [mark=none, densely dotted, color=cyan, mark 
%		options={scale=0.5}, line width=1.0pt] table[x index = 0, y 
%		index=5] {\gammaacc};
%		% 0.05
%		
%		\addplot [mark=none, densely dash dot, color=cyan, mark 
%		options={scale=0.5}, line width=1.0pt] table[x index = 0, y 
%		index=6] {\gammaacc};
%		% 0.02
%		
%		\addplot [mark=none, densely dotted, color=gray, mark 
%		options={scale=0.5}, line width=1.0pt] table[x index = 0, y 
%		index=8] {\gammaacc};
%		% 0.05
%		
%		\addplot [mark=none, densely dash dot, color=gray, mark 
%		options={scale=0.5}, line width=1.0pt] table[x index = 0, y 
%		index=9] {\gammaacc};
%		% 0.002
		
		\addplot [mark=none, densely dash dot, color=gray, mark 
		options={scale=0.5}, line width=1.0pt] table[x index = 0, y 
		index=4] {\gammaacc};
		% 0.001
		
		\addplot [mark=none, color=black, mark 
		options={scale=0.5}, line width=1.0pt] table[x index = 0, y 
		index=3] {\gammaacc};
		% 0.01

		\legend{
			$\gamma=1.0$, 
%			$\gamma=0.5$, 
%			$\gamma=0.2$, 
			$\gamma=0.01$,
%			$\gamma=0.05$,
%			$\gamma=0.02$,
			$\gamma=0.1$,
			$\gamma=0.001$
		}
		
	\end{axis}
\end{tikzpicture}
		\label{fig:gamma-acc}
	\end{subfigure}
	\hskip 10pt
	\begin{subfigure}{0.3\linewidth}
		\centering
		\captionsetup{margin={7mm,0mm}}
		\caption{\pgd-10}
		\vspace*{-1mm}
		\pgfplotstableread[col sep=space]{%
epoch	g1000	g0100	g0010	g0001
0	10.04	10.04	10.04	10.04
1	19.14	26.07	38.87	45.47
2	19.08	28.81	34.46	46.09
3	22.07	28.72	34.24	45.74
4	26.39	28.19	28.77	44.02
5	23.11	25.73	24.67	43.43
6	23.37	27.57	28.36	43.10
7	24.79	26.63	25.01	41.75
8	25.92	27.12	26.30	41.65
9	27.4	28.29	26.53	41.25
10	25.67	28.22	27.44	40.91
11	27.19	28	27.69	40.57
12	25.87	26.56	28.06	40.54
13	27.08	26.91	29.95	40.86
14	27.07	28.81	28.19	41.12
15	29.21	28.65	31.25	40.81
16	25.09	29.82	30.22	39.95
17	27.67	30.44	29.28	41.03
18	27.42	29.65	30.32	41.84
19	29.44	30.59	29.78	42.40
20	30.41	30.81	31.26	43.38
}\gammapgd

\begin{tikzpicture}
\begin{axis}[
	height=0.5\linewidth,
	width=0.85\linewidth,
	legend pos=outer north east,
	enlarge x limits=0.02,
	enlarge y limits=0.05,
	grid=major, 
	grid style={dashed,gray!30},
	xlabel={Epoch},
	ylabel={Accuracy [\%]},  % $\left[ \times 10^3 \right]$
	xmin=0,
	xmax=20,
	ymin=10,
	ymax=50,
	xtick={0,2,...,20},
	ytick={10, 20, ..., 50},
	xticklabels={0,2,4,6,8,10,12,14,16,18,20},
	yticklabels={10, 20, 30, 40, 50},
	yticklabel style = {font=\tiny, yshift=0.0ex},
	xticklabel style = {font=\tiny},
	ylabel style = {font=\fontsize{7}{7}\selectfont, yshift=-4ex},
	xlabel style = {font=\fontsize{7}{7}\selectfont, yshift=1.5ex},
	scale only axis,
	smooth,
	tension=0.5,
	legend style={at={(0.99,0.01)}, font=\legendsize, anchor=south 
		east, legend columns=1, fill=white, draw=white, 
		nodes={scale=0.6, transform shape}, column sep=3pt},
	legend cell align={left}
	]
	
	\addplot [mark=none, densely dotted, color=gray!70, mark 	
	options={scale=0.5}, line width=1.0pt] table[x index = 0, y 
	index=1] {\gammapgd};
	% 1.0
	
%	\addplot [mark=none, densely dotted, color=red, mark 
%	options={scale=0.5}, line width=1.0pt] table[x index = 0, y 
%	index=2] {\gammapgd};
%	% 0.5
%	
%	\addplot [mark=none, densely dash dot, color=red, mark 
%	options={scale=0.5}, line width=1.0pt] table[x index = 0, y 
%	index=3] {\gammapgd};
%	% 0.2
	
	\addplot [mark=none, densely dashed, color=gray!80, mark 
	options={scale=0.5}, line width=1.0pt] table[x index = 0, y 
	index=2] {\gammapgd};
	% 0.1
	
	%	\addplot [mark=none, densely dotted, color=cyan, mark 
	%	options={scale=0.5}, line width=1.0pt] table[x index = 0, y 
	%	index=5] {\gammapgd};
	%	% 0.05
	%	
	%	\addplot [mark=none, densely dash dot, color=cyan, mark 
	%	options={scale=0.5}, line width=1.0pt] table[x index = 0, y 
	%	index=6] {\gammapgd};
	%	% 0.02
	
	\addplot [mark=none, color=black, mark 
	options={scale=0.5}, line width=1.0pt] table[x index = 0, y 
	index=3] {\gammapgd};
	% 0.01
	
%	\addplot [mark=none, densely dotted, color=gray, mark 
%	options={scale=0.5}, line width=1.0pt] table[x index = 0, y 
%	index=8] {\gammapgd};
%	% 0.005
%	
%	\addplot [mark=none, densely dash dot, color=gray, mark 
%	options={scale=0.5}, line width=1.0pt] table[x index = 0, y 
%	index=8] {\gammapgd};
%	% 0.002

	\addplot [mark=none, densely dash dot, color=gray, mark 
	options={scale=0.5}, line width=1.0pt] table[x index = 0, y 
	index=4] {\gammapgd};
	% 0.001
	
%	\legend{
%		$\gamma=0.1$, 
%		$\gamma=0.05$, 
%		$\gamma=0.02$
%	}
		
	\end{axis}
\end{tikzpicture}
		\label{fig:gamma-pgd}
	\end{subfigure}
	\hskip 10pt
	%  \begin{subfigure}{0.3\linewidth}
		%    \centering
		%    \caption{\losshw}
		%    \input{figures/gamma-hw.tex}
		%    \label{fig:gamma-hw}
		%  \end{subfigure}
	\begin{subfigure}{0.3\linewidth}
		\centering
		\captionsetup{margin={12mm,0mm}}
		\caption{Compression rate $\compression$}
		\vspace*{-1mm}
		\pgfplotstableread[col sep=space]{%
epoch	g1000	g0100	g0010	g0001
0	0.1	0.1	0.1000	0.1000
1	0.001	0.003	0.0091	0.0787
2	0.001	0.001	0.0040	0.0528
3	0.001	0.001	0.0025	0.0387
4	0.001	0.001	0.0019	0.0303
5	0.001	0.001	0.0015	0.0230
6	0.001	0.001	0.0013	0.0184
7	0.001	0.001	0.0011	0.0150
8	0.001	0.001	0.0010	0.0127
9	0.001	0.001	0.0010	0.0107
10	0.001	0.001	0.0010	0.0093
11	0.001	0.001	0.0010	0.0082
12	0.001	0.001	0.0010	0.0074
13	0.001	0.001	0.0010	0.0067
14	0.001	0.001	0.0010	0.0061
15	0.001	0.001	0.0010	0.0057
16	0.001	0.001	0.0010	0.0054
17	0.001	0.001	0.0010	0.0051
18	0.001	0.001	0.0010	0.0049
19	0.001	0.001	0.0010	0.0048
20	0.001	0.001	0.0010	0.0047
}\gammarate

\begin{tikzpicture}
 \begin{semilogyaxis}[
	log ticks with fixed point,
	height=0.5\linewidth,
	width=0.85\linewidth,
	legend pos=outer north east,
	enlarge x limits=0.02,
	enlarge y limits=0.05,
	grid=major, 
	grid style={dashed,gray!30},
	xlabel={Epoch},
	ylabel={Rate},  % $\left[ \times 10^3 
	%\right]$
	xmin=0,
	xmax=20,
	ymin=0.001,
	ymax=0.1,
	xtick={0,2,...,20},
	ytick={0.001, 0.01, 0.1},
	xticklabels={0,2,4,6,8,10,12,14,16,18,20},
	yticklabels={0.001, 0.01, 0.1},
	yticklabel style = {font=\tiny, yshift=0.0ex},
	xticklabel style = {font=\tiny},
	ylabel style = {font=\fontsize{7}{7}\selectfont, yshift=-3ex},
	xlabel style = {font=\fontsize{7}{7}\selectfont, yshift=1.5ex},
	scale only axis,
%	smooth,
%	tension=0.1,
	legend style={at={(0.99,0.99)}, font=\legendsize, anchor=north 
		east, legend columns=1, fill=white, draw=white, 
		nodes={scale=0.6, transform shape}, column sep=2pt},
	legend cell align={left}
	]
	
	\addplot [mark=none, densely dotted, color=gray!70, mark 	
	options={scale=0.5}, line width=1.0pt] table[x index = 0, y 
	index=1] {\gammarate};
	% 1.0
	
%	\addplot [mark=none, densely dotted, color=red, mark 
%	options={scale=0.5}, line width=1.0pt] table[x index = 0, y 
%	index=2] {\gammarate};
%	% 0.5
%	
%	\addplot [mark=none, densely dash dot, color=red, mark 
%	options={scale=0.5}, line width=1.0pt] table[x index = 0, y 
%	index=3] {\gammarate};
%	% 0.2
	
	\addplot [mark=none, densely dashed, color=gray!80, mark 
	options={scale=0.5}, line width=1.0pt] table[x index = 0, y 
	index=2] {\gammarate};
	% 0.1
	
%	\addplot [mark=none, densely dotted, color=cyan, mark 
%	options={scale=0.5}, line width=1.0pt] table[x index = 0, y 
%	index=5] {\gammarate};
%	% 0.05
%	
%	\addplot [mark=none, densely dash dot, color=cyan, mark 
%	options={scale=0.5}, line width=1.0pt] table[x index = 0, y 
%	index=6] {\gammarate};
%	% 0.02
	
%	\addplot [mark=none, densely dotted, color=gray, mark 
%	options={scale=0.5}, line width=1.0pt] table[x index = 0, y 
%	index=8] {\gammarate};
%	% 0.005
%	
%	\addplot [mark=none, densely dash dot, color=gray, mark 
%	options={scale=0.5}, line width=1.0pt] table[x index = 0, y 
%	index=9] {\gammarate};
%	% 0.002
	
	\addplot [mark=none, densely dash dot, color=gray, mark 
	options={scale=0.5}, line width=1.0pt] table[x index = 0, y 
	index=4] {\gammarate};
	% 0.001
	
	\addplot [mark=none, color=black, mark 
	options={scale=0.5}, line width=1.0pt] table[x index = 0, y 
	index=3] {\gammarate};
	% 0.01
	
%		\legend{
%			$\gamma=0.01$, 
%			$\gamma=0.005$, 
%			$\gamma=0.002$,
%		}
		
	\end{semilogyaxis}
\end{tikzpicture}
		\label{fig:gamma-prate}
	\end{subfigure}
	\vskip -5pt
	\begin{subfigure}[h]{\linewidth}
		\centering
		\captionsetup{margin={5mm, 0mm}}
		\caption{Layer-wise preserved parameters}
		\vspace*{-1mm}
		\pgfplotstableread[col sep=space]{%
layer	g1000	g0500	g0200	g0100	g0050	g0020	g0010	g0005
1	605	718	662	687	705	764	794	804
2	740	850	816	900	942	974	1024	1116
3	658	671	670	684	673	682	683	772
4	699	695	726	712	708	715	702	770
5	755	745	759	751	749	739	740	764
6	849	827	833	819	815	799	789	773
7	850	828	828	815	810	794	783	755
8	981	955	952	937	927	914	898	858
9	1172	1143	1139	1121	1110	1092	1075	1044
10	1172	1144	1140	1121	1110	1092	1075	1044
11	1172	1144	1140	1121	1110	1092	1074	1018
12	1172	1144	1140	1121	1110	1092	1074	1018
13	1172	1144	1140	1121	1110	1092	1074	1019
14	837	815	812	797	787	773	758	725
15	635	619	622	613	610	601	594	597
16	1828	1852	1914	1974	2022	2081	2158	2213
}\gammastg

\begin{tikzpicture}
	\begin{axis}[
		height=0.2\linewidth,
		width=0.925\linewidth,
		legend pos=outer north east,
		enlarge x limits=0.02,
		enlarge y limits=0.15,
		grid=major, 
		grid style={dashed,gray!30},
		xlabel={Layer},
		ylabel={Preserved Params $\left[ \times 10^3 \right]$},
		xmin=1,
		xmax=16,
		ymin=500,
		ymax=2000,
		xtick={1,2,...,16},
		ytick={500, 1000, 1500, 2000},
		xticklabels={1,2,3,4,5,6,7,8,9,10,11,12,13, fc1, fc2, fc3},
		yticklabels={0.5, 1.0, 1.5, 2.0, 2.5},
		yticklabel style = {font=\tiny, yshift=-0.5ex},
		xticklabel style = {font=\tiny},
		ylabel style = {font=\fontsize{7}{7}\selectfont, yshift=-4ex},
		xlabel style = {font=\fontsize{7}{7}\selectfont, yshift=0.5ex},
		scale only axis,
		legend style={at={(0.015,0.95)}, font=\legendsize, anchor=north 
			west, legend columns=3, fill=white, draw=white, 
			nodes={scale=0.8, transform shape}, column sep=3pt},
		legend cell align={left}
		]
		
		\addplot [mark=*, densely dotted, color=gray!70, mark 	
		options={scale=0.6}, line width=1.0pt] table[x index = 0, y 
		index=1] {\gammastg};
		% 1.0
		
		\addplot [mark=*, densely dashed, color=gray!80, mark 
		options={scale=0.6}, line width=1.0pt] table[x index = 0, y 
		index=4] {\gammastg};
		% 0.1
		
%		\addplot [mark=none, color=black, mark 
%		options={scale=0.5}, line width=1.0pt] table[x index = 0, y 
%		index=7] {\gammastg};
%		% 0.01
%		
%		\addplot [mark=none, densely dotted, color=red, mark 
%		options={scale=0.5}, line width=1.0pt] table[x index = 0, y 
%		index=2] {\gammastg};
%		% 0.5
%		
%		\addplot [mark=none, densely dotted, color=cyan, mark 
%		options={scale=0.5}, line width=1.0pt] table[x index = 0, y 
%		index=5] {\gammastg};
%		% 0.05
%		
%		\addplot [mark=none, densely dotted, color=gray, mark 
%		options={scale=0.5}, line width=1.0pt] table[x index = 0, y 
%		index=8] {\gammastg};
%		% 0.005
%		
%		\addplot [mark=none, densely dash dot, color=red, mark 
%		options={scale=0.5}, line width=1.0pt] table[x index = 0, y 
%		index=3] {\gammastg};
%		% 0.2
%		
%		\addplot [mark=none, densely dash dot, color=cyan, mark 
%		options={scale=0.5}, line width=1.0pt] table[x index = 0, y 
%		index=6] {\gammastg};
%		% 0.02
		
		\addplot [mark=*, color=black, mark 
		options={scale=0.6}, line width=1.0pt] table[x index = 0, y 
		index=7] {\gammastg};
		% 0.01

		\legend{
			$\gamma=1.0$, 
			$\gamma=0.1$, 
			$\gamma=0.01$, 
%			$\gamma=0.5$,
%			$\gamma=0.05$,
%			$\gamma=0.05$,
%			$\gamma=0.2$,
%			$\gamma=0.02$
		}
		
	\end{axis}
\end{tikzpicture}
		\label{fig:gamma-stgs}
	\end{subfigure}
\end{figure}%\vspace*{-4mm}%
%\end{revisionblock}

%\begin{revisionblock}%
\subsection{Influence of dynamic $\gamma$ on \ourmethod's pruning}
\label{sec:gamma-adapt}

In this section, we present results for different regularization
step-sizes of \ourmethod when pruning \vgg to a sparsity of \perc{99.9}.
% .
As illustrated in~\cref{fig:gamma-acc,fig:gamma-pgd}, \ourmethod
exhibits strong oscillation for small step sizes ($\gamma\leq0.1$),
while larger steps ($\gamma=1.0$) restrict performance improvements. A
step size of $\gamma=0.01$, in turn, helps the model to better converge.
\ourmethod performs similarly for a slightly changed step size of
$\gamma=0.005$, but yielding the target compression is delayed.
% .
When lowering the step size further to \num{0.001}, the target
compression cannot be reached anymore as shown in
\cref{fig:gamma-prate}.
% .
Hence, small step sizes shorten the time for adversarial training but
cannot guarantee yielding the target compression.
% .
In~\cref{fig:gamma-stgs}, we summarize the generated strategies that
satisfy the target compression. With decreasing $\gamma$, \ourmethod
preserves layer in front and at the end stronger. Model parameters in
the middle layers are sacrificed to preserve higher performance.

\begin{revisionblock}
\subsection{Extended Comparison to Related Work}
\label{app:related-work}
% In~\cref{tab:bcsp-compare}, we summarize the experimental comparison
% between \ourmethod and \bcsp in different adversarial training
% approaches. Moreover, in~\cref{fig:bcsp-compare-svhn} and
% \cref{fig:mad-compare-svhn}, we conduct the experiments to compare
% \ourmethod's weight pruning with \bcsp and \mad on \svhn.
% .
\paragraph{Comparison to \bcsp}
%\cref{tab:bcsp-compare} summarizes the experiments using different
%adversarial training methods on \ourmethod and \bcsp. At sparsity
%\perc{90}, \vgg preserves a higher natural accuracy after \bcsp's
%pruning, whereas \ourmethod preserves higher adversarial robustness. At
%sparsity up to \perc{99}, however, \ourmethod overwhelms the robust
%pruning and achieves significantly higher performance than \bcsp.
%
\cref{fig:bcsp-compare-resnet} shows extended experiments for 
\resnet[18]. At sparsity \perc{99.9}, model collapse appears after 
\bcsp's pruning, particularly on \svhn.
%.
\cref{fig:bcsp-stg-resnet-svhn} visualizes pruning strategies for
\resnet[18] on \svhn. Similar to~\cref{fig:stg-overview-vgg-cr},
\bcsp exhibits an almost uniform pruning strategy, compressing
the input layer down to \num{0.0005}. This reduction refers to one out of
\num{1728}~parameters which clearly cannot yield any meaningful 
prediction.
Additionally, we extend to initialize \bcsp with \lamp non-uniform 
strategy~(cf. \cref{tab:stg-compare-bcsp}). Similar to the effect on 
\hydra, \bcsp significantly benefit from the non-uniform initialization. 
Especially, the plight of \bcsp at sparsity \perc{99.9} is solved. 
However, \hydra shows a better adaptability to \lamp's initialization. 
At the same time, \ourmethod remains overall the best performance on 
both \perc{99} and \perc{99.9} sparsity.

%\begin{table}[!h]
%  \centering
%  \vspace*{-2mm}
%  %\captionsetup{width=.95\linewidth}
%  \caption{\revision{Comparing \ourmethod with \bcsp for \vgg and
%    \resnet[18] learned on \cifar[10]. The natural accuracy and \pgd-10
%    robustness are presented left and right of the \texttt{/} 
%character.}}
%  \label{tab:bcsp-compare}
%  {\input{tables/bcsp-compare.tex}}
%\end{table}

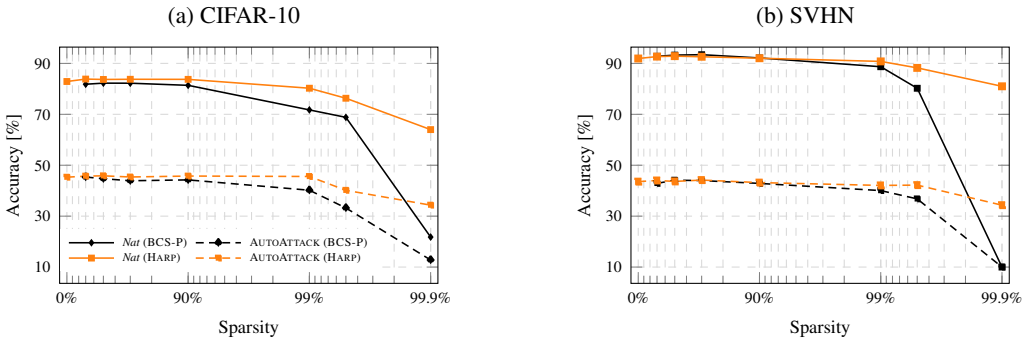
\begin{figure}[!htbp] \caption{\revision{Comparing \ourmethod with \bcsp
for pruning \resnet[18] weights on \cifar[10] and \svhn.}}
	\begin{subfigure}{0.45\linewidth}
		\caption{\cifar[10]}
		\begin{tikzpicture}
\begin{semilogxaxis}[
log ticks with fixed point,
height=0.5\linewidth,
width=0.8\linewidth,
legend pos=north east,
enlarge x limits=0.02,
enlarge y limits=0.08,
grid=both,
grid style={dashed,gray!30},
smooth,
tension=0.0,
xlabel={Sparsity},
ylabel={Accuracy [\%]},
xmin=0.001,
xmax=1,
ymin=10,
ymax=90,
x dir=reverse,
xtick={
	0.001, 0.01, 0.1, 1.0 
},
ytick={10, 30, 50, 70, 90},
xticklabels={99.9\%, 99\%, 90\%, 0\%},
yticklabels={10, 30, 50, 70, 90},
yticklabel style = {font=\fontsize{6}{6}\selectfont},
xticklabel style = {font=\fontsize{6}{6}\selectfont},
ylabel style = {font=\fontsize{7}{7}\selectfont, yshift=-4ex},
xlabel style = {font=\fontsize{7}{7}\selectfont, yshift=1ex},
%x tick label as interval
scale only axis,
legend style={at={(0.0,0.05)}, font=\fontsize{9}{9}\selectfont, 
anchor=south west, 
legend columns=2, fill=white, draw=none, nodes={scale=0.5, transform 
shape}, column sep=2pt},
legend cell align={left}
]

%\addplot [mark=*, color=tertiarycolor, mark options={scale=0.5}, 
%line width=0.6pt]
%coordinates {
%(1,	77.31)
%(2,	65.09)
%(3,	50.91)
%(4,	24.48)
%}; % \addlegendentry{\emph{Nat} (HYDRA)}

%\addplot [mark=*, color=primarycolor, mark options={scale=0.5}, line 
%width=0.6pt]
%coordinates {
%(1,	75.41)
%(2,	61.84)
%(3,	52.95)
%(4,	22.59)
%}; %\addlegendentry{\emph{Nat} (ADMM)}

\addplot [mark=diamond*, color=black, mark options={scale=0.5}, 
line width=0.6pt]
coordinates {
	(0.7, 81.82)
	(0.5, 82.24)
	(0.3, 82.22)
	(0.1, 81.38)
	(0.01, 71.73)
	(0.005, 68.81)
	(0.001, 21.78)
}; 

\addplot [mark=diamond*, densely dashed, color=black, mark 
options={scale=0.8}, line 
width=0.6pt]
coordinates {
	(0.7, 45.36)
	(0.5,    44.71)
	(0.3, 43.90)
	(0.1,   44.24)
	(0.01,  40.18)
	(0.005,  33.21)
	(0.001, 12.78)
};

%\addplot [mark=*, densely dashed, color=tertiarycolor, mark 
%options={scale=0.5}, 
%line 
%width=0.6pt]
%coordinates {
%	(1,	45.58)
%	(2,	39.80)
%	(3,	33.86)
%	(4,	21.40)
%}; %\addlegendentry{\emph{Adv} (HYDRA)}

%\addplot [mark=*, densely dashed, color=primarycolor, mark 
%options={scale=0.5}, line 
%width=0.6pt]
%coordinates {
%	(1,	44.37)
%	(2,	36.83)
%	(3, 34.64)
%	(4, 19.25)
%}; %\addlegendentry{\emph{Adv} (ADMM)}

\addplot [mark=square*, color=secondarycolor, mark options={scale=0.5}, 
line 
width=0.6pt]
coordinates {
	(1, 82.89)
	(0.7, 83.84)
	(0.5, 83.69)
	(0.3, 83.76)
	(0.1,	83.73)
	(0.01,	80.25)
	(0.005,	76.31)
	(0.001,	63.99)
}; 

\addplot [mark=square*, densely dashed, color=secondarycolor, mark 
options={scale=0.5}, line 
width=0.6pt]
coordinates {
	(1,		45.30)
	(0.7,	45.72)
	(0.5,		45.84)
	(0.3,	45.38)
	(0.1,	45.75)
	(0.01,	45.58)
	(0.005,	40.15)
	(0.001,	34.32)
};

\addlegendentry{\emph{Nat} (\bcsp)}
\addlegendentry{\emph{\Autoattack} (\bcsp)}
\addlegendentry{\emph{Nat} (\ourmethod)}
\addlegendentry{\emph{\Autoattack} (\ourmethod)}

\end{semilogxaxis}

\end{tikzpicture}
	\end{subfigure}
	\hfill
	\begin{subfigure}{0.45\linewidth}
		\caption{\svhn}
		\begin{tikzpicture}
\begin{semilogxaxis}[
log ticks with fixed point,
height=0.5\linewidth,
width=0.8\linewidth,
legend pos=north east,
enlarge x limits=0.02,
enlarge y limits=0.08,
grid=both,
grid style={dashed,gray!30},
smooth,
tension=0.0,
xlabel={Sparsity},
ylabel={Accuracy [\%]},
xmin=0.001,
xmax=1.0,
ymin=10,
ymax=90,
x dir=reverse,
xtick={
	0.001, 0.01, 0.1, 1.0 
},
ytick={10, 30, 50, 70, 90},
xticklabels={99.9\%, 99\%, 90\%, 0\%},
yticklabels={10, 30, 50, 70, 90},
yticklabel style = {font=\fontsize{6}{6}\selectfont},
xticklabel style = {font=\fontsize{6}{6}\selectfont},
ylabel style = {font=\fontsize{7}{7}\selectfont, yshift=-4ex},
xlabel style = {font=\fontsize{7}{7}\selectfont, yshift=1ex},
%x tick label as interval
scale only axis,
legend style={at={(0.0,0.0)}, font=\fontsize{9}{9}\selectfont, 
	anchor=south west, 
	legend columns=2, fill=white, draw=none, nodes={scale=0.5, transform 
		shape}, column sep=2pt},
legend cell align={left}
]

%\addplot [mark=*, color=tertiarycolor, mark options={scale=0.5}, 
%line width=0.6pt]
%coordinates {
%(1,	77.31)
%(2,	65.09)
%(3,	50.91)
%(4,	24.48)
%}; % \addlegendentry{\emph{Nat} (HYDRA)}

%\addplot [mark=*, color=primarycolor, mark options={scale=0.5}, line 
%width=0.6pt]
%coordinates {
%(1,	75.41)
%(2,	61.84)
%(3,	52.95)
%(4,	22.59)
%}; %\addlegendentry{\emph{Nat} (ADMM)}

%\addplot [mark=*, densely dashed, color=tertiarycolor, mark 
%options={scale=0.5}, 
%line 
%width=0.6pt]
%coordinates {
%	(1,	45.58)
%	(2,	39.80)
%	(3,	33.86)
%	(4,	21.40)
%}; %\addlegendentry{\emph{Adv} (HYDRA)}

%\addplot [mark=*, densely dashed, color=primarycolor, mark 
%options={scale=0.5}, line 
%width=0.6pt]
%coordinates {
%	(1,	44.37)
%	(2,	36.83)
%	(3, 34.64)
%	(4, 19.25)
%}; %\addlegendentry{\emph{Adv} (ADMM)}

\addplot [mark=square*, color=black, mark options={scale=0.5}, 
line 
width=0.6pt]
coordinates {
	(0.7,	92.78)
	(0.5, 	93.26)
	(0.3,	93.42)
	(0.1,	92.14)
	(0.01,	88.69)
	(0.005,	80.23)
	(0.001,	10.00)
}; 

\addplot [mark=square*, densely dashed, color=black, mark 
options={scale=0.5}, line 
width=0.6pt]
coordinates {
	(0.7,	42.87)
	(0.5, 	44.13)
	(0.3,	43.97)
	(0.1,	42.86)
	(0.01,	40.04)
	(0.005,	36.78)
	(0.001,	10.00)
};

%%%%%%%%%%%%%%%%%%%%%%%%%%%%%%%%%%%%%%%%%%%%%%%%%%%%%%%%%%%%%%%%%%%%

\addplot [mark=square*, color=secondarycolor, mark options={scale=0.6}, 
line width=0.6pt]
coordinates {
	(1, 91.92)
	(0.7, 92.67)
	(0.5, 92.85)
	(0.3, 92.61)
	(0.1,	92.03)
	(0.01,	90.78)
	(0.005,	88.24)
	(0.001,	81.00)
};

\addplot [mark=square*, densely dashed, color=secondarycolor, mark 
options={scale=0.6}, line width=0.6pt]
coordinates {
	(1,    43.56)
	(0.7, 44.01)
	(0.5,    43.64)
	(0.3, 44.17)
	(0.1,   43.18)
	(0.01,  42.07)
	(0.005,  42.19)
	(0.001, 34.31)
}; 

%\addlegendentry{\emph{Nat} (\bcsp)}
%\addlegendentry{\emph{\pgd-10} (\bcsp)}
%\addlegendentry{\emph{Nat} (\ourmethod)}
%\addlegendentry{\emph{\pgd-10} (\ourmethod)}

\end{semilogxaxis}

\end{tikzpicture}
	\end{subfigure}
	\label{fig:bcsp-compare-resnet}
\end{figure}

\begin{figure}[!h]
	\centering
	\caption{\revision{Strategy comparison for pruning a \resnet[18] 
	with target \perc{99.9} sparsity on \svhn.}}
	\vspace*{-1mm}
	\pgfplotstableread[col sep=space]{%
layer	uni0001	bcsp0001	harp0001
1	0.001	0.000578703703703704	0.237383127212524
2	0.001	0.000623914930555555	0.010349583812058
3	0.001	0.000813802083333333	0.0107299592345953
4	0.001	0.0009765625	0.0110683925449848
5	0.001	0.000868055555555556	0.010664171539247
6	0.001	0.00103081597222222	0.00580214243382216
7	0.001	0.000983344184027778	0.00323048117570579
8	0.001	0.001220703125	0.0346209481358528
9	0.001	0.000895182291666667	0.00332604348659515
10	0.001	0.000895182291666667	0.00323202298022807
11	0.001	0.00100029839409722	0.0018066824413836
12	0.001	0.00100877549913194	0.00101016834378243
13	0.001	0.001434326171875	0.011306463740766
14	0.001	0.00105624728732639	0.00100364908576012
15	0.001	0.0009307861328125	0.000986704253591597
16	0.001	0.000966389973958333	0.000580520078074187
17	0.001	0.00101131863064236	0.000358256715117022
18	0.001	0.00107574462890625	0.00334298657253385
19	0.001	0.00101343790690104	0.000358613498974591
20	0.001	0.00100665622287326	0.000358686258550733
21	0.001	0.0009765625	0.132312342524529
}\resnetstg

\begin{tikzpicture}
\begin{semilogyaxis}[
log ticks with fixed point,
height=0.2\textwidth,
width=0.92\textwidth,
legend pos=outer north east,
enlarge x limits=0.036,
enlarge y limits=0.02,
grid=major, 
grid style={dashed,gray!30},
xlabel={Layer},
ylabel={Compression Rate},  % $\left[ \times 10^3 \right]$
xmin=1,
xmax=21,
ymin=0.0001,
ymax=1.0,
xtick={1,2,...,21},
ytick={0.0001, 0.001, 0.01, 0.1, 1.0},
xticklabels={1, 2-1, 2-2, 2-3, 2-4, 3-1, 3-2, 3-short, 3-3, 3-4, 
4-1, 4-2, 4-short, 4-3, 4-4, 5-1, 5-2, 5-short, 5-3, 5-4, fc1},
yticklabels={0.0001, 0.001, 0.01, 0.1, 1.0},
yticklabel style = {font=\tiny, yshift=0.0ex},
xticklabel style = {font=\tiny},
ylabel style = {font=\fontsize{7}{7}\selectfont, yshift=-2.5ex},
xlabel style = {font=\fontsize{7}{7}\selectfont, yshift=0.5ex},
scale only axis,
legend style={at={(0.06,0.97)}, font=\legendsize, anchor=north 
west, 
legend columns=5, fill=white, draw=white, 
nodes={scale=0.9, 
transform shape}, column sep=3pt},
legend cell align={left}
]

\addplot [mark=square*, color=secondarycolor, mark 
options={scale=0.5}] table[x index = 0, y index=3] {\resnetstg};
% harp

\addplot [mark=*, color=tertiarycolor, mark 
options={scale=0.6}] table[x index = 0, y index=1] {\resnetstg};
% uniform_0001

\addplot [mark=pentagon*, color=black, mark 
options={scale=0.6}] table[x index = 0, y index=2] {\resnetstg};
% bcsp_0001

\addplot [mark=square*, color=secondarycolor, mark 
options={scale=0.5}] table[x index = 0, y index=3] {\resnetstg};
% harp

%\addlegendentry{\ourmethod};
%\addlegendentry{\lamp};
%\addlegendentry{\erk};
\legend{\ourmethod, Uniform, \bcsp}
\end{semilogyaxis}
\end{tikzpicture}
	\label{fig:bcsp-stg-resnet-svhn}
	\vspace*{-5mm}
\end{figure}

\begin{table}[!h]
%	\captionsetup{width=.85\linewidth}
	\caption{\lamp initialization on \hydra and \bcsp and their 
	comparison to \ourmethod on \cifar[10]. The natural accuracy 
	and \pgd-10 robustness are presented left and right of the 
	\texttt{/} character.}
	\tablesize
	\newcommand{\sotareader}[3]{%
	\csvreader[
	head to column names,
	filter = \equal{\method}{#1} \and \equal{\arch}{#2} \and 
	\equal{\prate}{#3},
	]%
	{results/lamp_bcsp_cifar10.csv}%
	{}
	{	
		& \unioabf
		& \uniaabf
		& \lampoabf
		& \lampaabf
	}
}

\newcommand{\harpreader}[3]{%
	\csvreader[
	head to column names,
	filter = \equal{\method}{#1} \and \equal{\arch}{#2} \and 
	\equal{\prate}{#3},
	%	late after line=\\
	]%
	{results/lamp_bcsp_cifar10.csv}%
	{}
	{	
		& \harpoabf
		& \harpaabf
	}
}

\begin{bcspstgtable}{\linewidth}
	\midrule
	\multirow{2}{*}{\resnet[18]}
	& \perc{99}
	\sotareader{hydra}{resnet}{0.01}
	\sotareader{bcsp}{resnet}{0.01}
	\harpreader{hydra}{resnet}{0.01}
	\\
	& \perc{99.9}
	\sotareader{hydra}{resnet}{0.001}
	\sotareader{bcsp}{resnet}{0.001}
	\harpreader{hydra}{resnet}{0.001}
	\\ \midrule
	\multirow{2}{*}{\vgg}
	& \perc{99}
	\sotareader{hydra}{vgg}{0.01}
	\sotareader{bcsp}{vgg}{0.01}
	\harpreader{bcsp}{vgg}{0.01}
	\\
	& \perc{99.9}
	\sotareader{hydra}{vgg}{0.001}
	\sotareader{bcsp}{vgg}{0.001}
	\harpreader{bcsp}{vgg}{0.001}
	\\
	\bottomrule
\end{bcspstgtable}
	\label{tab:stg-compare-bcsp}
\end{table}

\paragraph{Comparison to \mad}
We maintain the same training setting in~\cref{fig:mad-compare}. 
Similarly, \mad presents a 
promising performance on \svhn (cf~\cref{fig:mad-compare-svhn}) to 
further enhance adversarial resistance by moderately pruning. At the 
same time, \ourmethod presents robustness improvement as 
well in moderate pruning. In distinction from \mad, however, \ourmethod 
remains the stable performance on pruning a higher sparsity while \mad 
performs less efficiently at aggressive pruning.

\begin{figure}[!h]
	\centering
	% \captionsetup{font={small}, width=\linewidth}
	\caption{\revision{Comparing \pgd-10 adversarial robustness
			of \resnet[18] learned on \svhn and pruned by \radmm, 
			\hydra, \mad
			and \ourmethod. All pre-trained robust models are from 
			\mad.}}
	\vspace*{-1mm}
	\begin{subfigure}[h]{.45\linewidth}
		\centering
		\caption{\vgg}
		\begin{tikzpicture}
\begin{semilogxaxis}[
%log ticks with fixed point,
height=0.5\linewidth,
width=0.8\linewidth,
legend pos=outer north east,
enlarge x limits=0.02,
enlarge y limits=0.08,
grid=both,
grid style={dashed,gray!30},
smooth,
tension=0.0,
xlabel={Sparsity},
ylabel={Accuracy [\%]},
xmin=0.001,
xmax=1,
ymin=30,
ymax=60,
x dir=reverse,
xtick={0.001, 0.01, 0.1, 1.0},
ytick={30, 40, 50, 60}, 
xticklabels={99.9\%, 99\%, 90\%, 0\%},
yticklabels={30, 40, 50, 60},
yticklabel style = {font=\fontsize{6}{6}\selectfont},
xticklabel style = {font=\fontsize{6}{6}\selectfont},
ylabel style = {font=\fontsize{7}{7}\selectfont, yshift=-4ex},
xlabel style = {font=\fontsize{7}{7}\selectfont, yshift=1ex},
%x tick label as interval
scale only axis,
legend image post style={scale=0.6},
legend style={at={(0.02, 0.05)}, font=\fontsize{9}{9}\selectfont, 
	anchor=south west, 
	legend columns=2, fill=white, draw=none, nodes={scale=0.6, transform 
		shape}, column sep=2pt, line width=0.8pt},
legend cell align={left}
]

%%%%%%%%%%%%% VGG16 %%%%%%%%%%%%%

%\addplot [mark=square*, color=secondarycolor, mark 
%options={scale=0.6, solid}, line 
%width=0.8pt]
%coordinates {
%	(1, 93.2)
%	(1.43, 92.3)
%	(2, 92.4)
%	(3.33, 92.2)
%	(10, 92.4)
%	(100, 91.5)
%	(500, 90.3)
%}; \addlegendentry{\ourmethod}

%\addplot [mark=diamond*, color=madcolor, 
%mark options={scale=0.6, solid}, line width=0.8pt]
%coordinates {
%	(10,	92.4)
%	(10, 92.4)
%}; \addlegendentry{\mad}

%\addplot [only marks, mark=diamond*, color=madcolor, 
%mark options={scale=0.6, solid}, line width=0.8pt]
%coordinates {
%	(1,	93.2)
%	(10, 92.4)
%};

\addplot [mark=square*, color=secondarycolor, mark 
options={scale=0.6, solid}, line 
width=0.8pt]
coordinates {
	(1, 56.0)
	(0.7, 58.4)
	(0.5, 58.2)
	(0.3, 58.3)
	(0.1, 58.5)
	(0.01, 56.2)
	(0.005, 55.1)
	(0.001, 47.3)
}; \addlegendentry{\ourmethod}

\addplot [mark=diamond*, color=madcolor, mark 
options={scale=0.6, solid}, line width=0.8pt]
coordinates {
	(1,	56.0)
	(0.7, 58.9)
	(0.5, 58.5)
	(0.3, 58.8)
	(0.1, 58.4)
	(0.01, 51.8)
}; \addlegendentry{\mad}

\addplot [mark=*, color=tertiarycolor, mark 
options={scale=0.6, solid}, line width=0.8pt]
coordinates {
	(1, 50.2)
	(0.7, 50.2)
	(0.5, 50.2)
	(0.3, 49.9)
	(0.1, 47.4)
	(0.01, 37.8)
}; \addlegendentry{\hydra}

\addplot [mark=triangle*, color=primarycolor, mark 
options={scale=0.6, solid}, line width=0.8pt]
coordinates {
	(1, 50.2)
	(0.7, 48.4)
	(0.5, 48.9)
	(0.3, 48.3)
	(0.1, 47.4)
	(0.01, 33.7)
}; \addlegendentry{\radmms}

\addplot [mark=square*, color=secondarycolor, mark 
options={scale=0.6, solid}, line 
width=0.8pt]
coordinates {
	(1, 56.0)
	(0.7, 58.4)
	(0.5, 58.2)
	(0.3, 58.3)
	(0.1, 58.5)
	(0.01, 56.2)
	(0.005, 55.1)
	(0.001, 47.3)
}; %\addlegendentry{\ourmethod}

\end{semilogxaxis}

\end{tikzpicture}
	\end{subfigure}
	\hfill
	\begin{subfigure}[h]{.45\linewidth}
		\centering
		\caption{\resnet[18]}
		\begin{tikzpicture}
\begin{semilogxaxis}[
%log ticks with fixed point,
height=0.5\linewidth,
width=0.8\linewidth,
legend pos=outer north east,
enlarge x limits=0.02,
enlarge y limits=0.08,
grid=both,
grid style={dashed,gray!30},
smooth,
tension=0.0,
xlabel={Sparsity},
ylabel={Accuracy [\%]},
xmin=0.001,
xmax=1,
ymin=30,
ymax=60,
x dir=reverse,
xtick={0.001, 0.01, 0.1, 1.0},
ytick={30, 40, 50, 60}, 
xticklabels={99.9\%, 99\%, 90\%, 0\%},
yticklabels={30, 40, 50, 60},
yticklabel style = {font=\fontsize{6}{6}\selectfont},
xticklabel style = {font=\fontsize{6}{6}\selectfont},
ylabel style = {font=\fontsize{7}{7}\selectfont, yshift=-4ex},
xlabel style = {font=\fontsize{7}{7}\selectfont, yshift=1ex},
%x tick label as interval
scale only axis,
legend image post style={scale=0.6},
legend style={at={(0.02, 0.05)}, font=\fontsize{9}{9}\selectfont, 
	anchor=south west, 
	legend columns=2, fill=white, draw=none, nodes={scale=0.6, transform 
		shape}, column sep=2pt, line width=0.8pt},
legend cell align={left}
]

%%%%%%%%%%%%% ResNet18 %%%%%%%%%%%%%

\addplot [mark=*, color=tertiarycolor, mark options={scale=0.6}, 
line width=0.6pt]
coordinates {
	(1, 53.3)
	(0.7, 54.7)
	(0.5, 54.8)
	(0.3, 54.6)
	(0.1, 52.6)
	(0.01, 45.1)
}; %\hydra

\addplot [mark=triangle*, color=primarycolor, mark options={scale=0.6, 
solid}, 
line width=0.8pt]
coordinates {
	(1, 53.3)
	(0.7, 50.2)
	(0.5, 50.3)
	(0.3, 51.3)
	(0.1, 50.9)
	(0.01, 40.3)
}; %\radmm

\addplot [mark=diamond*, color=madcolor, mark options={scale=0.6, 
solid}, line 
width=0.8pt]
coordinates {
	(1,	59.6)
	(0.7, 61.4)
	(0.5, 61.3)
	(0.3, 61.4)
	(0.1, 60.6)
	(0.01, 55.4)
}; %\mad

%\addplot [mark=square*, color=secondarycolor, mark options={scale=0.6}, 
%line width=0.6pt]
%coordinates {
%	(1, 93.7)
%	(1.43, 92.5)
%	(2, 92.5)
%	(3.33, 92.7)
%	(10, 92.4)
%	(100, 91.1)
%	(500, 83.4)
%}; %\ourmethod

\addplot [mark=square*, color=secondarycolor, mark 
options={scale=0.6}, 
line width=0.6pt]
coordinates {
	(1, 59.6)
	(0.7, 60.7)
	(0.5, 61.0)
	(0.3, 60.6)
	(0.1, 60.3)
	(0.01, 58.7)
	(0.005, 56.4)
	(0.001, 48.1)
}; %\ourmethod

%\addplot [only marks, mark=diamond*, color=madcolor, mark 
%options={scale=0.6,	solid}, line 
%width=0.8pt]
%coordinates {
%	%	(1,	93.7)
%	(10, 93.3)
%}; %\mad

\end{semilogxaxis}

\end{tikzpicture}
	\end{subfigure}
	\label{fig:mad-compare-svhn}
\end{figure}
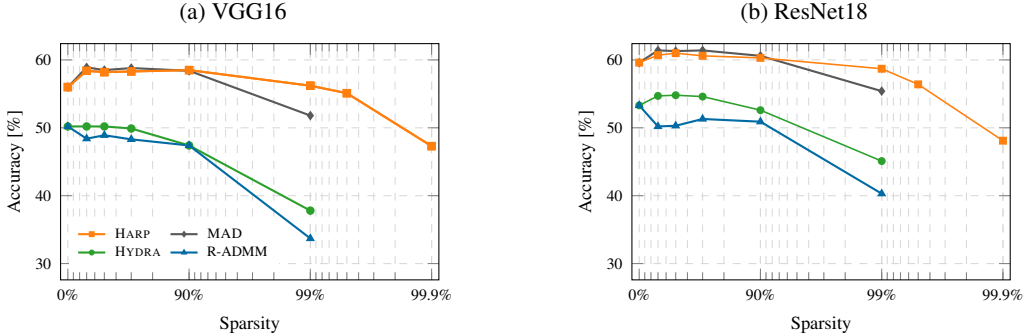

\paragraph{Comparison to~DNR}
\citet{Kundu2021DNR} propose a robust pruning framework, DNR, to learn
compact and robust neural network through dynamic network rewiring. 
%DNR yields good overall performance but its final compression rate 
%cannot be controlled due to the employed compactness regularization.
% .
To conduct the comparison, we follow the authors training
setting~\citep{Rakin2019ParamNoise} and use the authors evaluation
metrics. Results are reported in~\cref{tab:dnr-compare}.
% .
At a high compression ($50\times$) \ourmethod plays its strengths.
For \cifar[10], our method yields robustness scores comparable to DNR at
a compression of $\approx20\times$. The natural accuracy is slightly
inferior, though.
For the large-scale dataset, Tiny-\imagenet~\citep{Le2015TinyImageNet},
in turn, \ourmethod yields higher robustness than DNR for compression up
to $50\times$, but falls behind in natural accuracy again.
% .
In summary, \ourmethod yields higher robustness while DNR seems favors
natural accuracy.

\begin{table}[!h]
	\centering
	\tablesize
	\caption{\revision{Comparing \ourmethod with DNR in pruning neural network 
	weights.}}
	\newcolumntype{Y}{>{\centering\arraybackslash}X}

\sisetup{mode=text}
\setlength{\tabcolsep}{8pt}
\begin{tabular}{
		l
		l
		l
		S[table-format=2.2]
		S[table-format=2.2]
		S[table-format=2.2]
		S[table-format=2.2]
		S[table-format=2.2]
		S[table-format=2.2]
	}
\toprule
\bf Dataset
& \bf Model
& \bf Method
& {\bf Compression}
& {\bf Sparsity [\%]}
& {\bf Natural}
& {\bf \fgsm}
& {\bf \pgd-7}
\\ \midrule
\multirow{8}{*}{\cifar[10]}
& \multirow{4}{*}{\vgg}
& DNR
& 20.85~$\times$	& 95.4	& \bf 86.74	& 52.92		& 43.21	\\ 
\cmidrule(l){3-8}
& & \multirow{3}{*}{\ourmethod}
& 20.85~$\times$	& 95.4	& 85.09		& \bf 53.55	& \bf 45.34 \\
& &
& 50.00~$\times$	& 98.0	& 83.16		& 52.27		& 45.13		\\
& &
& 100.00~$\times$	& 99.0	& 81.35		& 50.83		& 42.75		\\
\cmidrule(l){2-8}
& \multirow{4}{*}{\resnet[18]}
& DNR
& 21.57~$\times$	& 95.2	& \bf 87.32	& 55.13		& 47.35	\\ 
\cmidrule(l){3-8}
& & \multirow{3}{*}{\ourmethod}
& 21.57~$\times$	& 95.2	& 86.97		& \bf 56.33	& \bf 49.31 \\
& &
& 50.00~$\times$	& 98.0	& 84.30		& 53.87		& 46.78		\\
& &
& 100.00~$\times$	& 99.0	& 82.35		& 52.06		& 45.47		\\
\midrule
\multirow{3}{*}{Tiny-ImageNet}
& \multirow{3}{*}{\vgg}
& DNR
& 20.63~$\times$	& 95.2	& \bf 51.71	& 18.21		& 14.46	\\ 
\cmidrule(l){3-8}
& & \multirow{2}{*}{\ourmethod}
& 20.63~$\times$	& 95.2	& 50.35		& \bf 19.06	& \bf 15.84 \\
& &
& 50.00~$\times$	& 98.0	& 50.21		& \bf 19.61	& \bf 16.24 \\
\bottomrule
\end{tabular}

	\label{tab:dnr-compare}
\end{table}
\end{revisionblock}

\subsection{Extended Strategy Analysis}
\label{app:stganalyse}

In this section, we investigate the impact of non-uniform 
strategies on robust weight pruning. Firstly, we compare different 
strategies used to initialize \ourmethod's pruning in order to 
present the insensibility on \ourmethod' layer-wise rate 
initialization.
In the next step, we analyze the end-to-end learning process with 
regard to the stability of training with and without using non-uniform 
strategies on \hydra. Secondly, we assess \hydra's performance by using 
\erk and \lamp for \imagenet, demonstrating that the positive impact of 
the strategies' non-uniformity is not consistent for 
different datasets. 

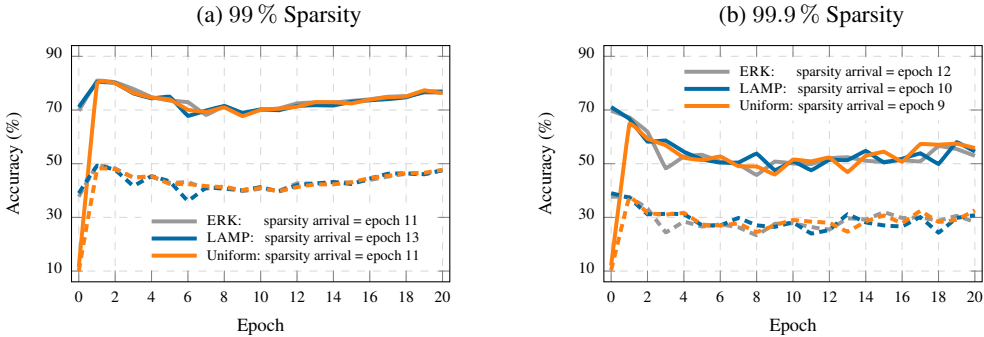
\begin{figure}[!h]
	\caption{Processes of weight pruning on 
		a robust \resnet for \cifar[10] by \ourmethod  
		with different initialization strategies: default 
		Uniform~(orange), \erk(gray) and \lamp(blue). Natural 
		performance and \pgd-10 adversarial robustness are presented as 
		solid lines and dashed lines, respectively.}
	\captionsetup[subfigure]{oneside, margin={12mm, 0mm}}
	\begin{subfigure}[h]{.45\linewidth}
		\caption{\perc{99} Sparsity}
		\vspace{-1mm}
		\pgfplotstableread[col sep=space]{%
epoch	nat	adv
0	11.58	10.04
1	79.74	47.75
2	79.97	47.9
3	76.34	44.83
4	74.59	45.37
5	73.98	42.45
6	70.07	42.41
7	69.26	41.65
8	70.97	41.22
9	67.73	40.16
10	70.06	40.95
11	70.46	39.71
12	71.23	41.3
13	72.91	42.27
14	72.88	42.31
15	72.46	43.06
16	73.84	44.68
17	74.77	45.5
18	75.04	46.46
19	77.34	46.62
20	76.27	47.7
}\uniinit

\pgfplotstableread[col sep=space]{%
epoch		nat	adv
0	69.68	37.73
1	80.89	49.40
2	80.34	48.22
3	77.85	44.86
4	74.80	45.14
5	73.48	42.87
6	72.93	43.14
7	68.18	40.69
8	71.43	41.54
9	68.47	39.76
10	70.05	40.69
11	70.56	39.91
12	72.48	42.68
13	72.84	42.55
14	72.79	43.15
15	73.41	43.00
16	73.97	43.92
17	74.95	45.21
18	75.15	46.25
19	76.50	46.05	
20	76.44	47.64
}\erkinit

\pgfplotstableread[col sep=space]{%
epoch		nat	adv
0	71.08	39.00
1	80.35	49.25
2	80.09	47.69
3	76.19	41.79
4	74.35	45.32
5	74.95	43.35
6	67.85	36.21
7	69.79	41.05
8	71.50	40.69
9	68.95	40.12
10	70.23	41.27
11	69.91	39.65
12	71.52	42.00
13	71.79	42.61
14	71.67	43.14
15	73.07	42.44
16	73.61	44.47
17	74.05	46.26
18	74.72	46.39
19	76.78	46.10
20	76.88	47.57
}\lampinit

\begin{tikzpicture}
\begin{axis}[
height=0.5\textwidth,
width=0.8\textwidth, % Scale the plot to \linewidth
enlarge x limits=0.02,
enlarge y limits=0.05,
grid=major, 
grid style={dashed,gray!30},
xlabel= Epoch, % Set the labels
ylabel= Accuracy (\%),
xmin=0,
xmax=20,
ymin=10,
ymax=90,
ytick={10, 30, 50, 70, 90},
xtick={0, 2, ..., 20},
yticklabels={10, 30, 50, 70, 90},
xticklabels={0, 2, 4, 6, 8, 10, 12, 14, 16, 18, 20},
smooth,
tension=0.1,
yticklabel style = {font=\tiny, yshift=0.3ex},
xticklabel style = {font=\tiny},
ylabel style = {font=\fontsize{7}{7}\selectfont, yshift=-3ex},
xlabel style = {font=\fontsize{7}{7}\selectfont, yshift=1.0ex},
legend style={at={(0.95,0.05)}, font=\fontsize{11}{11}\selectfont, 
anchor=south east, 
	legend columns=1, fill=white, draw=white, 
	nodes={scale=0.5, transform shape}, column sep=2pt},
legend cell align={left},
scale only axis,
every axis plot/.append style={line width=0.7pt},
]

%%%% Natural %%%%
\addplot+[mark=none, color=black!40, line 
width=1.5pt] table[x index = 0, y index=1] 
{\erkinit};

\addplot+[mark=none, color=primarycolor, line 
width=1.5pt] table[x index = 0, y index=1] 
{\lampinit};

\addplot+[mark=none, color=secondarycolor, solid, line 
width=1.5pt] table[x index = 0, y index=1] 
{\uniinit};

%%%% PGD10 %%%%

\addplot+[mark=none, color=black!40, densely dashed, line 
width=1.5pt] table[x index = 0, y index=2] 
{\erkinit};

\addplot+[mark=none, color=primarycolor, densely dashed, line 
width=1.5pt] table[x index = 0, y index=2] 
{\lampinit};

\addplot+[mark=none, color=secondarycolor, densely dashed, line 
width=1.5pt] table[x index = 0, y index=2] 
{\uniinit};

%\addplot+[mark=none, color=black!80, dashed, mark 
%options={scale=0.5}, line width=1.5pt]
%coordinates{
%	(20,	-10)
%	(20,	100)
%};

\addlegendentry{\erk: ~~~~~sparsity arrival = epoch 11};
\addlegendentry{\lamp: ~~sparsity arrival = epoch 13};
\addlegendentry{Uniform: sparsity arrival = epoch 11};
%\addlegendentry{\emph{Nat} (\erk)};
%\addlegendentry{\emph{\pgd-10} (\erk)};
%\addlegendentry{\emph{Nat} (\lamp)};
%\addlegendentry{\emph{\pgd-10} (\lamp)};
%\addlegendentry{\emph{Nat} (Uniform)};
%\addlegendentry{\emph{\pgd-10} (Uniform)};

\end{axis}

\end{tikzpicture}
		\label{fig:init-resnet-w001}
	\end{subfigure}
	\hskip 20pt
	\begin{subfigure}[h]{.45\linewidth}
		\caption{\perc{99.9} Sparsity}
		\vspace{-1mm}
		\pgfplotstableread[col sep=space]{%
0		12.07	10.3
1		64.67	37.59
2		59.26	31.66
3		56.85	31.03
4		52.21	31.65
5		51.31	27.32
6		52.68	26.97
7		49.07	27.74
8		48.97	24.52
9		46	27.09
10		51.54	29.02
11		50.86	28.41
12		52.32	27.97
13		46.9	24.7
14		52.91	28.48
15		54.45	30.7
16		50.72	27.88
17		57.29	32.38
18		57.06	28.33
19		57.48	29.12
20		55.84	32.63
}\uniinit

\pgfplotstableread[col sep=space]{%
epoch		nat	adv
0	69.70	37.71
1	67.23	37.61
2	61.87	33.26
3	48.36	24.48
4	52.68	28.39
5	53.30	26.58
6	50.79	27.38
7	49.78	26.29
8	45.82	23.39
9	50.78	27.61
10	50.07	27.78
11	49.83	26.40
12	52.06	25.47
13	52.40	29.62
14	51.08	29.22
15	50.70	31.91
16	51.32	29.77
17	50.86	30.03
18	56.55	28.78
19	55.45	30.54
20	52.91	28.25
}\erkinit

\pgfplotstableread[col sep=space]{%
epoch		nat	adv
0	71.08	39.00
1	66.73	37.28
2	58.24	31.17
3	58.62	31.33
4	54.45	31.16
5	51.58	27.09
6	50.41	27.00
7	50.42	29.76
8	53.74	27.04
9	47.56	26.47
10	50.93	28.10
11	47.58	24.00
12	51.44	25.36
13	51.31	31.13
14	54.74	28.12
15	50.45	27.01
16	51.85	26.64
17	53.86	30.06
18	49.95	24.36
19	57.99	29.80
20	54.50	30.79
}\lampinit

\begin{tikzpicture}
\begin{axis}[
height=0.5\textwidth,
width=0.8\textwidth, % Scale the plot to \linewidth
enlarge x limits=0.02,
enlarge y limits=0.05,
grid=major, 
grid style={dashed,gray!30},
xlabel= Epoch, % Set the labels
ylabel= Accuracy (\%),
xmin=0,
xmax=20,
ymin=10,
ymax=90,
ytick={10, 30, 50, 70, 90},
xtick={0, 2, ..., 20},
yticklabels={10, 30, 50, 70, 90},
xticklabels={0, 2, 4, 6, 8, 10, 12, 14, 16, 18, 20},
smooth,
tension=0.1,
yticklabel style = {font=\tiny, yshift=0.3ex},
xticklabel style = {font=\tiny},
ylabel style = {font=\fontsize{7}{7}\selectfont, yshift=-3ex},
xlabel style = {font=\fontsize{7}{7}\selectfont, yshift=1.0ex},
legend style={at={(0.95,0.95)}, font=\fontsize{11}{11}\selectfont, 
	anchor=north east, 
	legend columns=1, fill=white, draw=white, 
	nodes={scale=0.5, transform shape}, column sep=2pt},
legend cell align={left},
scale only axis,
every axis plot/.append style={line width=0.7pt},
]

%%%% Natural %%%%
\addplot+[mark=none, color=black!40, line 
width=1.5pt] table[x index = 0, y index=1] 
{\erkinit};

\addplot+[mark=none, color=primarycolor, line 
width=1.5pt] table[x index = 0, y index=1] 
{\lampinit};

\addplot+[mark=none, color=secondarycolor, solid, line 
width=1.5pt] table[x index = 0, y index=1] 
{\uniinit};

%%%% PGD10 %%%%

\addplot+[mark=none, color=black!40, densely dashed, line 
width=1.5pt] table[x index = 0, y index=2] 
{\erkinit};

\addplot+[mark=none, color=primarycolor, densely dashed, line 
width=1.5pt] table[x index = 0, y index=2] 
{\lampinit};

\addplot+[mark=none, color=secondarycolor, densely dashed, line 
width=1.5pt] table[x index = 0, y index=2] 
{\uniinit};

%\addplot+[mark=none, color=black!80, dashed, mark 
%options={scale=0.5}, line width=1.5pt]
%coordinates{
%	(20,	-10)
%	(20,	100)
%};

\addlegendentry{\erk: ~~~~~sparsity arrival = epoch 12};
\addlegendentry{\lamp: ~~sparsity arrival = epoch 10};
\addlegendentry{Uniform: sparsity arrival = epoch 9};
%\addlegendentry{\emph{Nat} (\erk)};
%\addlegendentry{\emph{\pgd-10} (\erk)};
%\addlegendentry{\emph{Nat} (\lamp)};
%\addlegendentry{\emph{\pgd-10} (\lamp)};
%\addlegendentry{\emph{Nat} (Uniform)};
%\addlegendentry{\emph{\pgd-10} (Uniform)};

\end{axis}

\end{tikzpicture}
		\label{fig:init-resnet-w0001}
	\end{subfigure}\vspace*{-2mm}
	\label{fig:stg-init}
\end{figure}

\paragraph{Analysis of \ourmethod's strategy 
initialization}\label{app:init-analysis}
As demonstrated by \citet{Sehwag2020HYDRA}, score initialization is key 
to realize a improvement on the weight pruning. 
In \cref{fig:stg-init}, we provide the learning processes of 
\ourmethod's pruning at sparsity that are initialized by \erk, \lamp 
and the default uniform strategy.
%Here, we , and 
%evidence from experiments, that the learning on layer compression 
%rates 
%and masking scores concurrently is not sensible to different 
%strategies' initialization. \cref{fig:stg-init} provides the pruning 
%phases of \ourmethod with default uniform, \erk and \lamp strategies 
%in 
%sparsity \perc{90} for initializing layer compression rates. 
In comparison with \erk and \lamp, uniform strategy leads to a 
dramatic performance degradation after initialization. However, the 
training on scores $\scores_{W}$ and rates \rates shows the cape of 
recovering the network performance, resulting in a largely similar 
network performance to \erk and \lamp after the first epoch training. 
In the subsequent epochs, the learning curves on different 
initializations are essentially overlap.
%.
By further evaluating the final pruned networks after fine-tuning, 
results in \cref{tab:stg-init} show that the difference between uniform 
adn non-uniform strategies are negligible for the layer compression 
rate initialization.
%using non-uniform strategies 
%for the initialization does not produce significant difference in 
%comparison with using uniform strategy. 
Confirming, that \ourmethod is insensitive to different strategies 
initialization. Therewith, we rely on the uniform strategy as the 
default setting in our~experiments.
\begin{table}[tbh]
	\centering
	\caption{Comparing different strategies initialization on 
	\ourmethod for \cifar[10].}
	\tablesize
	\newcolumntype{Y}{>{\centering\arraybackslash}X}

\tablesize
\sisetup{mode=text}
\begin{tabular}{
wl{8mm}
wc{10mm}
S[table-format=2.2]@{~/~}S[table-format=2.2]
S[table-format=2.2]@{~}S[table-format=2.2]@{~/~}S[table-format=2.2]@{~}S[table-format=2.2]
S[table-format=2.2]@{~}S[table-format=2.2]@{~/~}S[table-format=2.2]@{~}S[table-format=2.2]
}
\toprule
\bf Model	
& \bf Sparsity	
& \multicolumn{2}{c}{\bf Uniform}	
& \multicolumn{4}{c}{\bf \erk}
& \multicolumn{4}{c}{\bf \lamp}
\\ \midrule
\multirow{2}{*}{\resnet[18]}
& \perc{99}	
& 80.25 & 50.36	
& 79.42 & \text{($-$0.83)} & 50.43 & \text{($+$0.07)}
& 80.48 & \text{($+$0.23)} & 50.33 & \text{($-$0.03)}
\\
& \perc{99.9}	
& 63.99 & 39.39	
& 63.61 & \text{($-$0.38)} & 39.20 & \text{($-$0.19)}
& 63.64 & \text{($-$0.35)} & 39.43 & \text{($+$0.04)}
\\ \midrule
\multirow{2}{*}{\vgg}
& \perc{99}	
& 78.50 & 48.71	
& 78.38 & \text{($-$0.12)} & 48.40 & \text{($+$0.31)}
& 78.82 & \text{($+$0.32)} & 48.69 & \text{($-$0.02)}
\\
& \perc{99.9}	
& 59.33 & 37.46	
& 59.79 & \text{($+$0.46)} & 37.72 & \text{($+$0.26)}
& 59.85 & \text{($+$0.52)} & 37.65 & \text{($+$0.19)}
\\ \bottomrule
\end{tabular}

	\label{tab:stg-init}
\end{table}

\begin{figure}[!tbh]
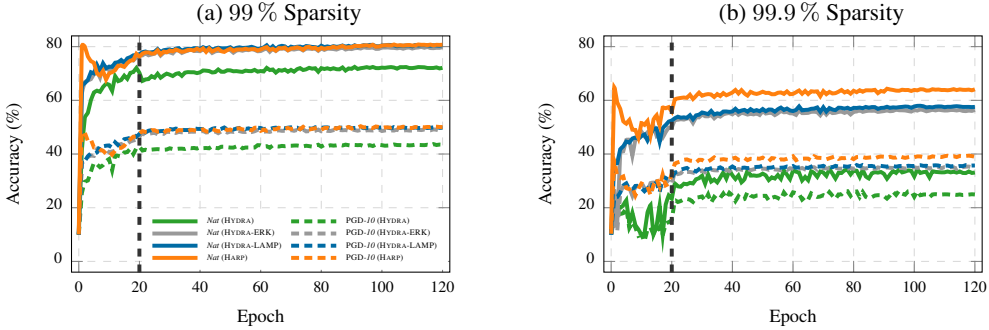

	\caption{End-to-end learning processes of weight pruning on 
		a pre-trained robust \resnet by using \hydra(green), 
		\hydra-\erk(gray), \hydra-\lamp(blue), and 
		\ourmethod(orange).}
	\captionsetup[subfigure]{oneside, margin={12mm, 0mm}}
	\begin{subfigure}[h]{.45\linewidth}
		\caption{\perc{99} Sparsity}
		\vspace{-2mm}
		\input{figures/pcurve-resnet-w001-lamp-erk.tex}
		\label{fig:end2end-resnet-w001}
	\end{subfigure}
	\hskip 20pt
	\begin{subfigure}[h]{.45\linewidth}
		\caption{\perc{99.9} Sparsity}
		\vspace{-2mm}
		\input{figures/pcurve-resnet-w0001-lamp-erk.tex}
		\label{fig:end2end-resnet-w0001}
	\end{subfigure}
	\label{fig:learn-process}
\end{figure}

\paragraph{Comparison of end-to-end compression pipelines}
In \cref{fig:learn-process}, we presents a comparison between 
different strategies on \hydra and \ourmethod across the 
end-to-end compression pipeline. For pruning \perc{99} weights, the 
potential of the original \hydra is impaired by the use of a uniform 
strategy in the pruning phase, limiting the potential for improving via 
fine-tuning.
%the original \hydra starts being 
%limited by the uniform strategy already in the pruning phase, which 
%restricts the improvement potential by fine-tuning. 
At the same time, the non-uniform strategies \erk and \lamp prove their 
positive impact in the pruning stage. Without fine-tuning, they achieve 
already better results than \hydra. In \ourmethod's pruning, we observe 
a down-up performance curve, indicating that the network performance 
recovers after the robust networks is compressed. Further improved 
by fine-tuning, \ourmethod achieves slightly better results than \hydra 
with \erk and \lamp. In pruning a \perc{99.9} sparsity, \hydra produces 
a dramatic decrease in performance that can, however, be compensated by 
using \lamp and \erk. Using \ourmethod during pruning results in
the highest network performance, approximating the final result 
after~fine-tuning.
%.

\paragraph{Comparing non-uniform strategies for \imagenet}
Presented in \cref{tab:stg-compare}, \hydra is significantly improved 
by applying \erk and \lamp. In \cref{tab:compare-imgnet-stg}, we 
further investigate their impact on the large-scale dataset \imagenet. 
Compared to original \hydra, using \erk and \lamp strategies leads to a 
performance degradation, and with \lamp yields more harm than 
\erk at sparsity \perc{99}. \cref{fig:stg-overview-resnet50} visualizes 
the strategies of sparsity \perc{99} for \imagenet experiments. Similar 
to \cref{fig:compare-cifar10-stg}, \lamp and \erk tend to preserve 
parameters, leading to a global parameter distribution that 
approximates a uniform shape. However, since \ourmethod preserves 
more parameters in the former layers and the last fully connected 
layer, more input and output information are preserved. 
\begin{table}[!htbp]
%	\vspace*{-1mm}
	\centering
	\caption{Comparing \ourmethod weight pruning with \hydra, and 
	\hydra with \erk and \lamp for \imagenet. Values in brackets 
	show the performance change by using \erk and \lamp on \hydra.}
	\newcommand{\mycsvreader}[3]{%
	\csvreader[
	head to column names,
	head to column names prefix = COL,
	filter = \equal{\COLattack}{#1} \and \equal{\COLpreg}{#2} \and 
	\equal{\COLprate}{#3}
	]%
	{results/imagenet_results.csv}%
	{}
	{
		& \COLhydrastgbf
		& \COLhydraerkbf
		& \COLhydraerkdelta
		& \COLhydralampbf
		& \COLhydralampdelta
		& \COLharpstgbf
	}
}

\begin{imagenetstgtable}{\linewidth}
	\midrule
	% \multirow{7}{*}{\rotatebox{90}{Channel Prune}} & 
%	\multirow{4}{*}{\makecell{\bfseries \freeat\\ 
%	\num{60.25}\,/\,\num{32.82}}}
	--
	& \num{60.25}
	\mycsvreader{nat}{weight}{0.1} 
	\mycsvreader{nat}{weight}{0.01}\\
	\pgd
	& \num{32.82} 	
	\mycsvreader{pgd}{weight}{0.1} 
	\mycsvreader{pgd}{weight}{0.01}\\
	\cwinf
	& \num{30.67}
	\mycsvreader{cw}{weight}{0.1} 
	\mycsvreader{cw}{weight}{0.01}\\
	\autopgd
	& \num{31.54}
	\mycsvreader{autopgd}{weight}{0.1}
	\mycsvreader{autopgd}{weight}{0.01}\\
	\autoattack
	& \num{28.79}
	\mycsvreader{autoattack}{weight}{0.1}
	\mycsvreader{autoattack}{weight}{0.01}\\
	\bottomrule
%	\multirow{7}{*}{\rotatebox{90}{Weight Prune}} 
%	& \multirow{3}{*}{\perc{90}} 
%	\mycsvreader{hydrabf}{weight}{0.1} \\
%	& \mycsvreader{radmmbf}{weight}{0.1} \\ \cmidrule{3-7}
%	& \mycsvreader{harpbf}{weight}{0.1}
%	\\ \cmidrule{2-7}
%	& \multirow{3}{*}{\perc{99}} 
%	\mycsvreader{hydrabf}{weight}{0.01} \\ 
%	& \mycsvreader{radmmbf}{weight}{0.01} \\ \cmidrule{3-7}
%	& \mycsvreader{harpbf}{weight}{0.01}
%	\\ \midrule

\end{imagenetstgtable}
	\label{tab:compare-imgnet-stg}
%	\vspace*{-3mm}
\end{table}
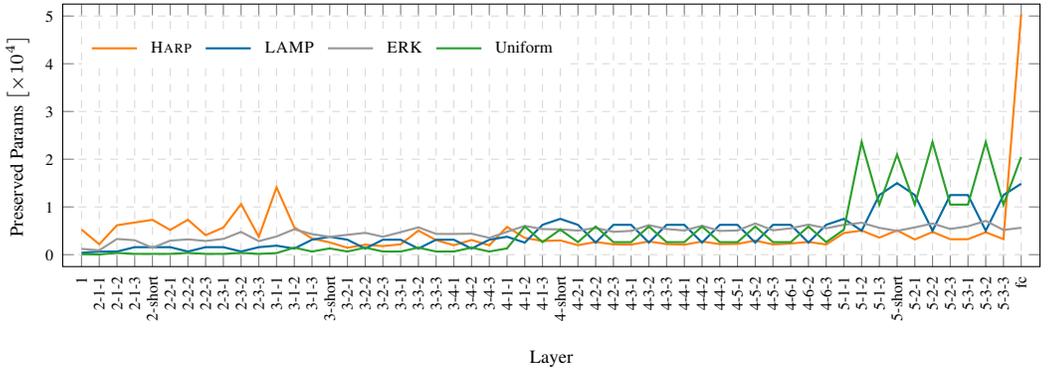
\begin{figure}[!htbp]
	\centering
	%\captionsetup{margin={8mm, 0mm}}
	\caption{Strategies of \ourmethod, \lamp, and \erk for 
		pruning a \perc{99} sparsity on \imagenet.}
	\pgfplotstableread[col sep=space]{%
layer	uni001	harp001	erk001	lamp001
1	94	5265	395	1215
2	41	2178	634	909
3	369	6162	653	3285
4	164	6733	1570	3023
5	164	7297	1570	1440
6	164	5173	1570	2938
7	369	7317	653	3204
8	164	4079	1570	2870
9	164	5684	1570	3303
10	369	10591	653	4758
11	164	3745	1570	2829
12	328	14091	1881	3807
13	1475	5731	1277	5322
14	655	3382	3129	4299
15	1311	2568	3753	3743
16	655	1449	3129	4155
17	1475	2079	1277	4571
18	655	1786	3129	3751
19	655	2175	3129	4702
20	1475	5044	1277	5730
21	655	3043	3129	4345
22	655	1971	3129	4332
23	1475	3078	1277	4398
24	655	1944	3129	3515
25	1311	5828	3753	4795
26	5898	3518	2525	6021
27	2621	2858	6249	5356
28	5243	2986	7497	5296
29	2621	1984	6249	5054
30	5898	2658	2525	5598
31	2621	2158	6249	4785
32	2621	2108	6249	4982
33	5898	2773	2525	6184
34	2621	2214	6249	5381
35	2621	2086	6249	5061
36	5898	2759	2525	6053
37	2621	2203	6249	4973
38	2621	2262	6249	5069
39	5898	2949	2525	6535
40	2621	2148	6249	5104
41	2621	2338	6249	5515
42	5898	2664	2525	6300
43	2621	2137	6249	5531
44	5243	4533	7497	6258
45	23593	5049	5021	6682
46	10486	3566	12488	5617
47	20972	5079	14984	4981
48	10486	3178	12488	5718
49	23593	4761	5021	6523
50	10486	3231	12488	5382
51	10486	3239	12488	5944
52	23593	4696	5021	7080
53	10486	3245	12488	5173
54	20480	50393	14857	5634
}\resnetparams

\begin{tikzpicture}
\begin{axis}[
height=0.25\textwidth,
width=0.93\textwidth,
legend pos=outer north east,
enlarge x limits=0.02,
enlarge y limits=0.05,
grid=major, 
grid style={dashed,gray!30},
xlabel={Layer},
ylabel={Preserved Params $\left[ \times 10^4 \right]$},
xmin=1,
xmax=54,
ymin=0.0,
ymax=50000,
scaled ticks=false,
xtick={1,2,...,54},
ytick={0, 10000, 20000, 30000, 40000, 50000},
xticklabels={
	1, 
	2-1-1, 2-1-2, 2-1-3, 2-short, 
	2-2-1, 2-2-2, 2-2-3, 
	2-3-1, 2-3-2, 2-3-3, 
	3-1-1, 3-1-2, 3-1-3, 3-short,
	3-2-1, 3-2-2, 3-2-3,
	3-3-1, 3-3-2, 3-3-3,
	3-4-1, 3-4-2, 3-4-3,
	4-1-1, 4-1-2, 4-1-3, 4-short,
	4-2-1, 4-2-2, 4-2-3,
	4-3-1, 4-3-2, 4-3-3,
	4-4-1, 4-4-2, 4-4-3, 
	4-5-1, 4-5-2, 4-5-3,
	4-6-1, 4-6-2, 4-6-3,
	5-1-1, 5-1-2, 5-1-3, 5-short,
	5-2-1, 5-2-2, 5-2-3,
	5-3-1, 5-3-2, 5-3-3,
	fc
},
yticklabels={0, 1, 2, 3, 4, 5},
yticklabel style = {font=\tiny, yshift=0.0ex},
xticklabel style = {font=\tiny, rotate=90},
ylabel style = {font=\fontsize{7}{7}\selectfont, yshift=-4ex},
xlabel style = {font=\fontsize{7}{7}\selectfont, yshift=-3ex},
%x tick label as interval
scale only axis,
legend style={at={(0.02,0.9)}, font=\legendsize, anchor=north 
	west, 
	legend columns=4, fill=white, draw=white, 
	nodes={scale=0.9, 
		transform shape}, column sep=3pt},
legend cell align={left}
]

\addplot [draw=white, thick, color=secondarycolor, mark 
options={scale=0.5}] table[x index = 0, y index=2] {\resnetparams};
% harp_001

\addplot [draw=white, thick, color=primarycolor, mark 
options={scale=0.5}] table[x index = 0, y index=3] {\resnetparams};
% lamp_001

\addplot [draw=white, thick, color=black!40, thick, mark 
options={scale=0.5}] table[x index = 0, y index=4] {\resnetparams};
% erk_001

\addplot [draw=white, thick, color=tertiarycolor, mark 
options={scale=0.5}] table[x index = 0, y index=1] {\resnetparams};
% uni_001

%\addlegendentry{\ourmethod};
%\addlegendentry{\lamp};
%\addlegendentry{\erk};
\legend{\ourmethod, \lamp, \erk, Uniform}
\end{axis}
\end{tikzpicture}
	\label{fig:stg-overview-resnet50}
	\vspace*{-5mm}
	\vspace*{-2mm}
\end{figure}

\begin{figure}[!t]
	\centering
	\captionsetup{width=1.05\linewidth}
	\caption{Parameter distribution of \vgg pruned to
		\perc{99} sparsity with \mbox{\pgdat.}}
	\begin{subfigure}{\linewidth}
		\caption{\cifar[10]}
		\includegraphics[scale=0.23]{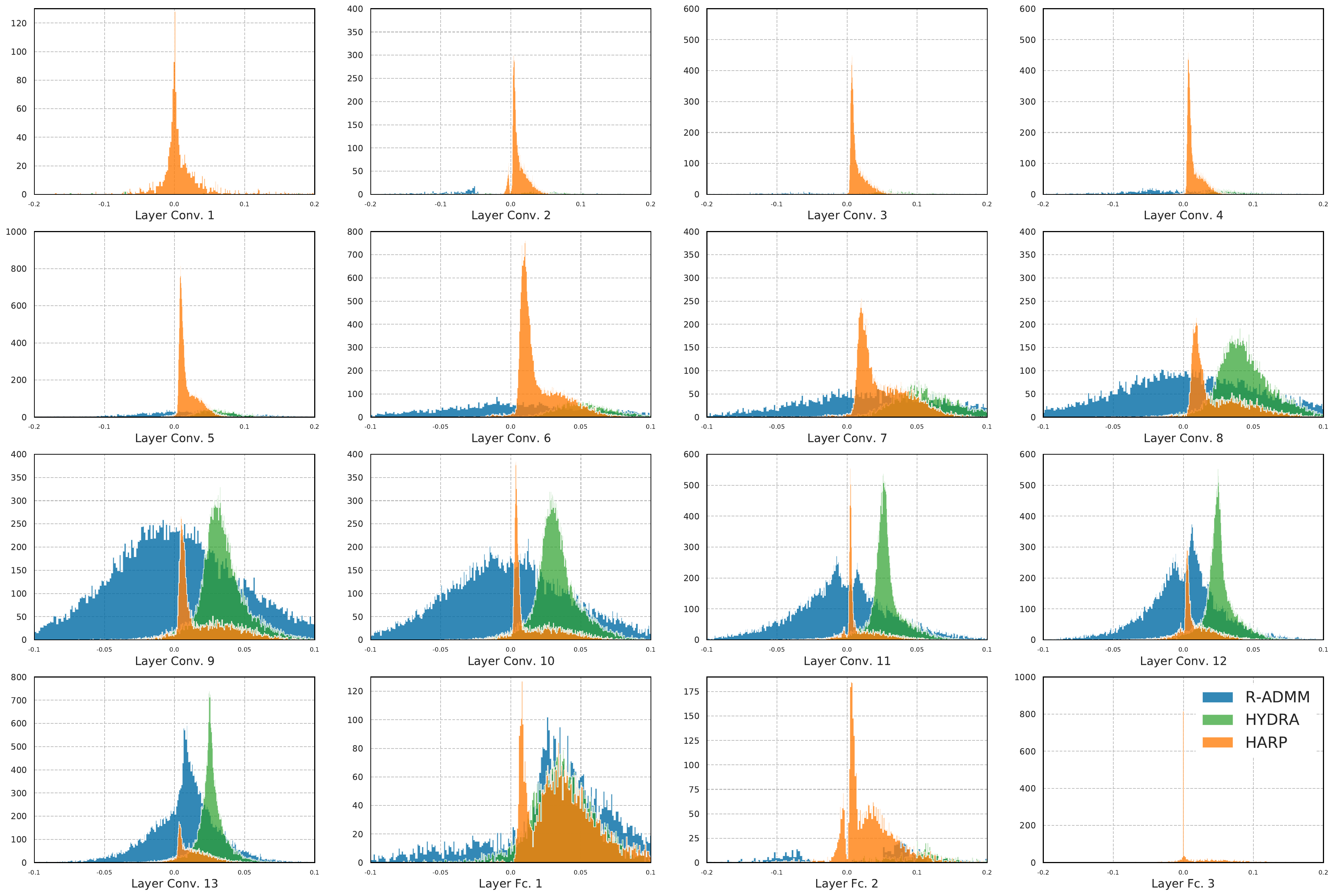}
		\label{fig:vgg-cifar-wdist}
	\end{subfigure}
	\vskip 5pt
	\begin{subfigure}{\linewidth}
		\caption{\svhn}
		\includegraphics[scale=0.23]{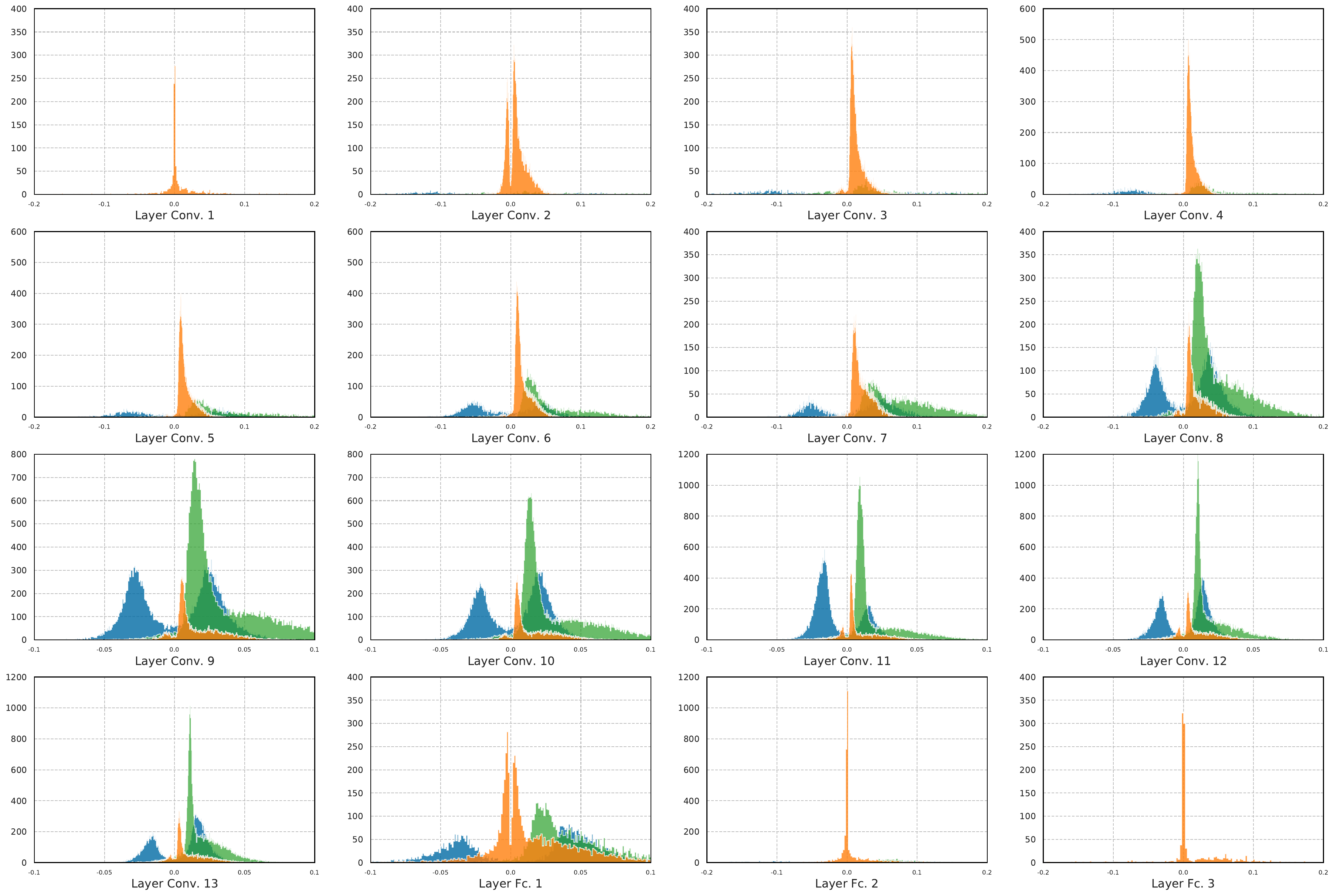}
		\label{fig:vgg-svhn-wdist}
	\end{subfigure}
\end{figure}

\subsection{Parameter Distribution}
\label{app:paramsdist}

\ourmethod prunes neural networks by incorporating connection
importance-scores for learning pruning masks and the layers' compression
rates into the optimization procedure. The pruned models particularly
benefit from the non-uniformity of \ourmethod's compression strategy,
yielding high natural accuracy and high adversarial robustness.
% .
\citet{Ye2019Adversarial} argue that wider parameter distributions
promise higher robustness, while \citet{Sehwag2020HYDRA} show that 
parameters close to zero are crucial to preserve the
adversarial robustness.
% .
In this section, we take \vgg learned on \cifar[10] and \svhn as 
examples to analyze the difference in parameter distributions along the 
network's layers and compare \ourmethod with \radmm and \hydra.

%\begin{figure}[!t]
%	\centering
%	\captionsetup{width=1.05\linewidth}
%	\caption{Parameter distribution of \vgg (\cifar[10]) pruned to
%	\perc{99} weight sparsity with \mbox{\pgdat.}}
%	\includegraphics[scale=0.23]{figures/layerdist-vgg16-w001.pdf}
%	\label{fig:vgg-chdist}
%\end{figure}

\paragraph{Parameter distribution when pruning weights} 
\cref{fig:vgg-cifar-wdist} and \cref{fig:vgg-svhn-wdist} visualize the 
layer-wise weight distribution of
\vgg pruned by \radmm~(blue), \hydra~(green), and \ourmethod~(orange) 
on datasets \cifar[10] and \svhn, respectively.
Concatenated architectures such as \vgg possess relatively high sparsity
in the middle layers~\citep{Lee2021Layer, Evci2020Rigging}. This
phenomenon is also captured by \ourmethod, which learns to preserve the
first six layers and the last layer.
% .
Another intriguing but interesting observation concerns non-zero
parameters: \radmm tends to favor middle layers that contain many zero
parameters. In contrast, \hydra tries to push the model's weights
``away'' from zero, which reduces sparsity and thus achieves higher
robustness than \radmm. 
% .
Pruning weights with \ourmethod, however, converges to a different
distribution. All but the first and last layers, have weights with
(mostly) smaller magnitude than for models pruned with \hydra---few of
them, however, remain zero-valued. For instance, in the penultimate
layer, preserved weights are close to zero and the distribution is more
compact than those of \hydra and \radmm.
By comparing the parameter distribution of middle layers in \cifar[10] 
and \svhn, we observe that \hydra tends to have a rather close-zero 
distribution in pruning networks for \svhn than its behavior in 
\cifar[10]. Recall \cref{tab:compare-cifar-w} and 
\cref{tab:compare-svhn-w}, the 
performance gap between \hydra and \ourmethod in \svhn is smaller than 
in \cifar[10]. This implies an interesting phenomenon that encouraging 
a close-zero distribution yields a positive impact on pruning robust 
networks.
%
%\begin{figure}[!t]
%	\centering
%	\captionsetup{width=1.05\linewidth}
%	\caption{Parameter distribution of \vgg (\svhn) pruned to
%		\perc{99} sparsity with \mbox{\pgdat.}}
%	\includegraphics[scale=0.23]{figures/layerdist-vgg16-svhn-w001.pdf}
%	\label{fig:vgg-svhn-wdist}
%\end{figure}

\subsection{Performance with natural training}
\label{app:natprune}

We extend our experiments to pruning naturally pre-trained networks 
with datasets \cifar[10] and \svhn, and consider \hydra, \hydra with 
\lamp, and the original \lamp pruning in the comparison.
In \cref{tab:natprune}, there exists no significant difference 
between the pruning methods at sparsity \perc{90}. 
For \cifar[10], \hydra yields a larger performance drop at sparsity 
\perc{99}. In contrast, using \lamp directly or using \hydra with 
\lamp's strategy yields better results. With regards to the highest 
sparsity \perc{99.9}, \lamp pruning and \hydra with \lamp strategy have 
significantly better performance than original \hydra. Meanwhile, 
\ourmethod shows the highest compatibility to prune networks for 
\cifar[10]. Additionally, in pruning a sparsity \perc{99.9} for \svhn, 
\ourmethod presents the pruning performance that deviates from the best 
result by no more than \perc{0.5}. Conclusively, pruning naturally 
pre-trained networks benefits more from the layer-wise specific 
non-uniform strategies. At the same time, \ourmethod offers more 
significant improvement about concerning pruning robust networks than 
other approaches.
\begin{table}[!h]
	\centering
	\caption{Comparing pruning weights of naturally pre-trained 
		networks}
	\tablesize
	\newcommand{\mycsvreader}[3]{%
	\csvreader[
	head to column names,
	head to column names prefix = COL,
	filter = \equal{\COLarch}{#1} \and \equal{\COLdataset}{#2} \and 
	\equal{\COLsparsity}{#3}
	]%
	{results/nat_weight.csv}%
	{}
	{	
		& \perc{\COLsparsity}
		& \COLhydrabf
		& \COLlampbf
		& \COLhydralampbf
		& \COLharpbf
	}
}

\begin{nattable}{\linewidth}
	\cmidrule(r){1-7} \cmidrule(l){8-13}
	\multirow{3}{*}{\vgg}
	& \multirow{3}{*}{\num{93.68}}
	\mycsvreader{vgg}{cifar}{90}
	& \multirow{3}{*}{\num{95.71}}
	\mycsvreader{vgg}{svhn}{90}
	\\ %\cmidrule(lr){3-7} \cmidrule(lr){9-13}
	& \mycsvreader{vgg}{cifar}{99}
	& \mycsvreader{vgg}{svhn}{99}
	\\ %\cmidrule(lr){3-7} \cmidrule(lr){9-13}
	& \mycsvreader{vgg}{cifar}{99.9}
	& \mycsvreader{vgg}{svhn}{99.9}
	\\ \cmidrule(r){1-7} \cmidrule(l){8-13}
	\multirow{3}{*}{\resnet[18]}
	& \multirow{3}{*}{\num{95.13}}
	\mycsvreader{resnet}{cifar}{90}
	& \multirow{3}{*}{\num{96.13}}
	\mycsvreader{resnet}{svhn}{90}
	\\ %\cmidrule(lr){3-7} \cmidrule(lr){9-13}
	& \mycsvreader{resnet}{cifar}{99}
	& \mycsvreader{resnet}{svhn}{99}
	\\ %\cmidrule(lr){3-7} \cmidrule(lr){9-13}
	& \mycsvreader{resnet}{cifar}{99.9}
	& \mycsvreader{resnet}{svhn}{99.9}
	\\ \bottomrule
\end{nattable}
	\label{tab:natprune}
\end{table}

\subsection{Extension to Channel Pruning}
\label{app:chprune}

Structurally pruning neural networks has better compatibility with the 
hardware deployment, while its coarse pruning granularity leads to 
larger performance degradation than weight pruning, itself increasing 
the difficulty of pruning robust networks. In this section, 
we extend our method \ourmethod on pruning network layer channels. To 
map \ourmethod on channel pruning, we use \flops estimation 
(denoted as function $FLOPS(\cdot)$) to replace the sparsity in 
$\losshw$ (expressed as \cref{eq:hw-loss-flops}). 
Moreover, we determine $\compression_{t}$ by considering network global 
\flops reduction, in order for the pruned model to have the same 
size as a uniform strategy would have in \flops.

\begin{revisionblock}
\begin{equation}
\label{eq:hw-loss-flops}
\losshw(\prunedparams, \compression_{t}) := \max \left\{
  \frac{
    FLOPs\left( 
    \prunedparams %\npreservedparams 
    \right)
  }{
    \compression_{t}\cdot FLOPs\left(
    \params %\nparams 
    \right)%\,~~
  }
  -1
~,~ 0
\right\}
\end{equation}
\end{revisionblock}

Different from weight pruning, importance scores $\scores^{(l)}$ 
originates $\mathbb{R}^{c_i^{(\layer)}}$ for channel pruning. In 
addition, we use \cref{eq:scores-init-c} to initialize $\scores^{(l)}$ 
in channel pruning and follow the order of weight magnitude 
\citep{Ye2019Adversarial} by summing up the weights of each input 
channel (expressed as $\Csum(\cdot)$) and normalizing by the maximal 
channel-wise summation.

\begin{equation}
\labelx{eq:scores-init-c}
{
	\scores_{C}^{(\layer)} =  
	\left(
	\eta\cdot\cfrac{\Csum(|\params^{(\layer)}|)}{\max(\Csum(|\params^{(\layer)}|))}
	\right)_{c_i^{(\layer)}}
}
\end{equation}

\cref{alg:harp-alg} presents the complete implementation of \ourmethod.
Note that weight pruning is largely independent of the network 
architecture as it does not change the layer's size (we~merely zero out 
weights). However, pruning channels/filters requires special 
attention.
%.
When facing \resnet[-like] architectures, we 
do not train the compression rates of all shortcut layers directly but 
update them by assigning the rate of the connected input layer in the 
residual block instead. This way, the pruned input layer aligns with 
the channel dimensionality of connected pruned shortcut-layers.

\begin{figure*}[h]%\vspace{-3mm}
	\begin{tikzpicture}
	\node(CP) [] at (0, 0.1) {\includegraphics[width=30mm, trim=5mm 0 
		7mm 0, clip]{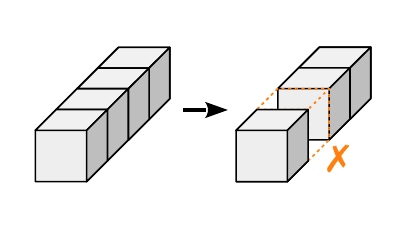}};
	\node(text) [below=2mm of CP, anchor=south] {
		\scriptsize Channel\,/\,Filter pruning
	};
	\node(plot1) [] at (4.5, 0.1) {% was:[right=2mm of CP] {
		\begin{tikzpicture}
\begin{axis}[
height=0.2\linewidth,
width=0.3\linewidth,
legend pos=north east,
enlarge x limits=0.02,
enlarge y limits=0.05,
grid=both,
grid style={dashed,gray!30},
smooth,
tension=0.0,
xlabel={xFLOPs}, % reduction [\%]},
ylabel={Accuracy [\%]},
xmin=1,
xmax=20,
ymin=10,
ymax=90,
xtick={1, 2, 4, 10, 20},
ytick={10, 30, 50, 70, 90},
xticklabels={\num{1}, \num{2}, \num{4}, \num{10}, \num{20}},
yticklabel style = {font=\fontsize{5}{5}\selectfont},
xticklabel style = {font=\fontsize{5}{5}\selectfont},
ylabel style = {font=\fontsize{6}{6}\selectfont, yshift=-4ex},
xlabel style = {font=\fontsize{6}{6}\selectfont, yshift=2ex},
%x tick label as interval
scale only axis,
legend style={at={(0.0,0.03)}, font=\fontsize{9}{9}\selectfont, 
anchor=south west, 
legend columns=3, fill=white, draw=none, nodes={scale=0.5, transform 
shape}, column sep=1pt},
legend cell align={left}
]

%%%%%%%%%%%%%%%%%%%%%%%%%%%%%%%%
%           Nat Acc            %
%%%%%%%%%%%%%%%%%%%%%%%%%%%%%%%%

%%%%%%%%%%%% HARP %%%%%%%%%%%%%%
\addplot [mark=square*, color=secondarycolor, mark options={scale=0.5}, 
line 
width=0.6pt]
coordinates {
	(1, 79.68)
%	(2, 79.92)
	(2, 78.73)
	(4, 76.28)
	(10,	73.24)
	(20,	68.71)
}; 

%%%%%%%%%%%% R-ADMM %%%%%%%%%%%%%%
\addplot [mark=triangle*, color=primarycolor, mark options={scale=0.5}, 
line 
width=0.6pt]
coordinates {
	(1, 79.68)
%	(2, 79.45)
	(2, 78.14)
	(4,	75.72)
	(10,	67.01)
	(20,	64.47)
};

%%%%%%%%%%%% Hydra %%%%%%%%%%%%%%
\addplot [mark=*, color=tertiarycolor, mark options={scale=0.5}, 
line 
width=0.6pt]
coordinates {
	(1, 79.68)
	(2, 66.19)
	(4,	53.89)
	(10,	34.41)
	(20,	26.57)
};

\addplot [mark=square*, densely dashed, color=secondarycolor, mark 
options={scale=0.5}, line 
width=0.6pt]
coordinates {
	(1, 47.60)
	%	(2, 47.74)
	(2, 47.69)
	(4, 46.48)
	(10,	45.68)
	(20,	43.20)
};

\addplot [mark=triangle*, densely dashed, color=primarycolor, mark 
options={scale=0.5}, line 
width=0.6pt]
coordinates {
	(1, 47.60)
	%	(2, 45.98)
	(2, 44.82)
	(4,	44.33)
	(10,	41.25)
	(20,	40.28)
};

\addplot [mark=*, densely dashed, color=tertiarycolor, mark 
options={scale=0.5}, line 
width=0.6pt]
coordinates {
	(1, 47.60)
	(2, 41.60)
	(4,	38.75)
	(10,	28.89)
	(20,	22.55)
};

%%%%%%%%%%%% HARP %%%%%%%%%%%%%%
\addplot [mark=square*, color=secondarycolor, mark options={scale=0.5}, 
line 
width=0.6pt]
coordinates {
	(1, 79.68)
	%	(2, 79.92)
	(2, 78.73)
	(4, 76.28)
	(10,	73.24)
	(20,	68.71)
}; 

\addplot [mark=square*, densely dashed, color=secondarycolor, mark 
options={scale=0.5}, line 
width=0.6pt]
coordinates {
	(1, 47.60)
	%	(2, 47.74)
	(2, 47.69)
	(4, 46.48)
	(10,	45.68)
	(20,	43.20)
};

\addlegendentry{\ourmethod}
\addlegendentry{\radmms}
\addlegendentry{\hydra}
%\addlegendentry{\emph{\pgd-10} (\ourmethod)}
%\addlegendentry{\emph{\pgd-10} (\radmm)}

\end{axis}

\end{tikzpicture}
		
	};
	\node(text1) [below=2mm of plot1, anchor=south] {
		\scriptsize \qquad\vgg
	};
	\node(plot2) [] at (10, 0.1) {% was:[right=55mm of CP] {
		\begin{tikzpicture}
\begin{axis}[
height=0.2\linewidth,
width=0.3\linewidth,
legend pos=north east,
enlarge x limits=0.02,
enlarge y limits=0.05,
grid=both,
grid style={dashed,gray!30},
smooth,
tension=0.0,
xlabel={xFLOPs}, % reduction [\%]},
ylabel={Accuracy [\%]},
xmin=1,
xmax=20,
ymin=10,
ymax=90,
xtick={1, 2, 4, 10, 20},
ytick={10, 30, 50, 70, 90},
xticklabels={\num{1}, \num{2}, \num{4}, \num{10}, \num{20}},
yticklabel style = {font=\fontsize{5}{5}\selectfont},
xticklabel style = {font=\fontsize{5}{5}\selectfont},
ylabel style = {font=\fontsize{6}{6}\selectfont, yshift=-4ex},
xlabel style = {font=\fontsize{6}{6}\selectfont, yshift=2ex},
%x tick label as interval
scale only axis,
legend style={at={(0.0,0.05)}, font=\fontsize{9}{9}\selectfont, 
	anchor=south west, 
	legend columns=3, fill=white, draw=none, nodes={scale=0.5, transform 
		shape}, column sep=1pt},
legend cell align={left}
]

%%%%%%%%%%%% HARP %%%%%%%%%%%%%%
\addplot [mark=square*, densely dashed, color=primarycolor, mark 
options={scale=0.5}, line 
width=0.6pt]
coordinates {
	(1, 50.05)
	%	(2, 48.86)
	(2, 49.16)
	(4, 48.76)
	(10,	43.64)
	(20,	43.08)
}; 

%%%%%%%%%%%% R-ADMM %%%%%%%%%%%%%%
\addplot [mark=triangle*, densely dashed, color=secondarycolor, mark 
options={scale=0.5}, line 
width=0.6pt]
coordinates {
	(1, 50.05)
	%	(2, 49.72)
	(2, 49.18)
	(4, 48.13)
	(10,	47.28)
	(20,	46.91)
};

%%%%%%%%%%%% Hydra %%%%%%%%%%%%%%
\addplot [mark=*, densely dashed, color=tertiarycolor, mark 
options={scale=0.5}, line 
width=0.6pt]
coordinates {
	(1, 50.05)
%	(2, 46.67)
	(2, 43.97)
	(4,	40.65)
	(10,	29.57)
	(20, 22.46)
};

%%%%%%%%%%%% HARP %%%%%%%%%%%%%%
\addplot [mark=square*, densely dashed, color=primarycolor, mark 
options={scale=0.5}, line 
width=0.6pt]
coordinates {
	(1, 50.05)
	%	(2, 48.86)
	(2, 49.16)
	(4, 48.76)
	(10,	43.64)
	(20,	43.08)
};

%------------------------------%

%%%%%%%%%%%% R-ADMM %%%%%%%%%%%%%%
\addplot [mark=triangle*, color=primarycolor, mark options={scale=0.5}, 
line 
width=0.6pt]
coordinates {
	(1, 82.89)
%	(2, 82.83)
	(2, 82.05)
	(4, 79.68)
	(10,	74.78)
	(20,	70.41)
};

%%%%%%%%%%%% Hydra %%%%%%%%%%%%%%
\addplot [mark=*, color=tertiarycolor, mark options={scale=0.5}, 
line 
width=0.6pt]
coordinates {
	(1, 82.89)
%	(2, 77.29)
	(2, 70.21)
	(4,	54.85)
	(10,	37.13)
	(20, 27.15)
};

%%%%%%%%%%%% HARP %%%%%%%%%%%%%%
\addplot [mark=square*, color=secondarycolor, mark options={scale=0.5}, 
line 
width=0.6pt]
coordinates {
	(1, 82.89)
	%	(2, 83.10)
	(2, 82.28)
	(4, 80.85)
	(10,	78.22)
	(20,	77.73)
};

%\addlegendentry{\emph{\pgd-10} (\ourmethod)}
%\addlegendentry{\emph{\pgd-10} (\radmm)}
%\addlegendentry{\emph{\pgd-10} (\hydra)}

\end{axis}

\end{tikzpicture}
	};
	\node(text2) [below=2mm of plot2, anchor=south] {
		\scriptsize \qquad\resnet[18]
	};
	
	%\node(sep) [minimum height=25mm] at ($(CP)!0.5!(WP)$) {};
	%\draw [dashed] (sep.north) -- (sep.south) {};
	\node(dummy) [below=0mm of CP] {}; %was: -3mm
	\end{tikzpicture}
	\caption{\revision{
      Structural pruning by controlling FLOPs on a \vgg model and a
      \resnet[18] model for \cifar[10]. Solid lines show the natural
      accuracy of \ourmethod (orange), \hydra (green) and \radmm (blue).
      Dashed lines represent the methods' adversarial robustness of
      \pgd-10.
	}}%\vspace{-2mm}
	\label{fig:overview-flops}
\end{figure*}
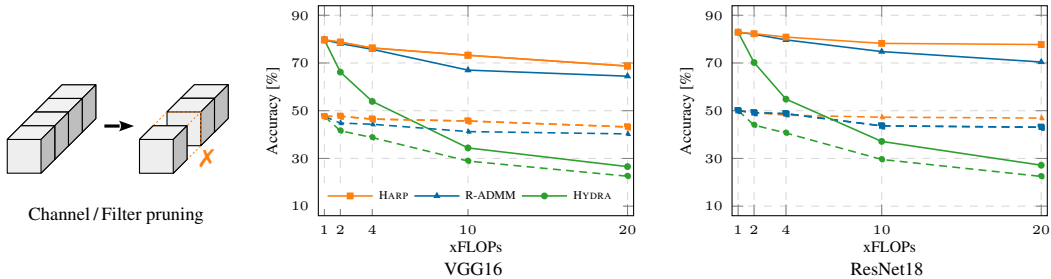

\paragraph{Comparing channel pruning with related work}
\label{app:compare-ch-radmm}
%.
\citet{Ye2019Adversarial} have shown the capability of \radmm in 
structurally pruning robust networks, controlling the network \flops 
straightforwardly by a global structural compression rate. 
\revision{
Similarly, \hydra~\citep{Sehwag2020HYDRA} extends to structural pruning
and improves over magnitude-based criterion~\citep{Han2015Learning}. 
}
For channel pruning with \ourmethod, we start off from the uniform strategy 
with $\compression_{init}=1.0$ and use $\gamma=0.02$ to ensure the 
arrival at the target compression rate $\compression_t$. As presented 
experiment results in \cref{fig:overview-flops}, \ourmethod shows 
promising performance in preserving a robust model while pruning 
channels. \radmm possesses a similar performance on pruning up to 
\num{4.0} x\flops, while experiencing higher model performance 
degradation at x\flops of \num{10} and \num{20}.
\revision{
Channel pruning with \hydra, in turn, significantly harms model 
performance at \num{2.0}~x\flops compression.
}

%\begin{revisionblock}
%In a next step, we compare \ourmethod with other, mode advanced methods 
%such as DNR and \anpvs that support structural pruning. 
%Additionally, we compare with other more advanced method DNR and \anpvs 
%that supports structural pruning on robust networks. We use the same 
%training settings and rely on the same \pgd robustness evaluation used 
%in~\citep{Kundu2021DNR} to compare to DNR and in 
%\citep{Madaan2020Adversarial} to compare to \anpvs, respectively. 
%In~\cref{tab:compare-channel}, we prune a 
%similar compression as achieved in DNR and \anpvs. Both DNR and \anpvs 
%offers a good performance on preserving the natural accuracy. However, 
%\ourmethod achieves the highest performance with regard to robustness 
%preservation. 
%%
%\begin{table}[!h]
%	\centering
%	\tablesize
%%	\captionsetup{width=.7\linewidth}
%	\revision{
%	\caption{Comparing \ourmethod's channel pruning with \anpvs and DNR. 
%	Values in brackets indicate the difference to \ourmethod.}
%	}
%	\vspace*{1mm}
%	\input{tables/compare-channel-prune.tex}
%	\label{tab:compare-channel}
%\end{table}
%\end{revisionblock}

\paragraph{Comparison to \anpvs}%\label{subsec:app-anpvs}
\citet{Madaan2020Adversarial} propose to suppress the latent feature
level vulnerability in neural networks by gradually pruning a model
during training based on a regularized loss function.
% The perturbation budget and the step size of the PGD attack as
% proposed by \cite{Madry2018Towards} is set to \mbox{\sfrac{8}{255}
% $\approx0.03$} and \mbox{\sfrac{2}{255} $\approx0.007$} for the
% \cifar[10] dataset, whose pixels/features are in the range of $[0,
% 255]$.
% .
In the open-source implementation of
\anpvs~\footnote{\url{https://github.com/divyam3897/ANP_VS}},
pre-processing \cifar[10] data is done using image standardization (\eg
\code{tf.image.per\_image\_standardization()} function). Standardization
transforms the input to have mean \num{0.0} and variance \num{1.0}, that
is, input features may fall outside of an $[0.0, 1.0]$ interval.
However, \anpvs clips adversarial perturbations to $[0.0, 1.0]$,
resulting in overall weaker adversarial examples used for the evaluation
as it would be the case for \mbox{0-1 rescaling} the inputs.
Note, the latter is used for evaluating \ourmethod in all experiments.
% .
% However, for constructing the PGD attack, \anpvs clips the final
% adversarial example to $[0.0, 1.0]$ and, thus, looses the $[-1.0,
% 0.0)$ range of the input. Consequently, 0-1 rescaling should have been
% used from the start.
We investigate the influence of input ranges by comparing training with
\anpvs using both pre-processing
\begin{wraptable}{r}{.45\linewidth}
	\centering
	%	\captionsetup{width=.8\linewidth}
	\setlength{\tabcolsep}{3pt}
	\tablesize
	\vspace{-7pt}
	\caption{Comparing \anpvs with different 
		pre-processing methods.}
	\vspace{-5pt}
	\sisetup{
	mode=text,
%	separate-uncertainty,
%	table-alignment-mode = format,
	table-number-alignment = center,	
}

%\begin{tabularx}{\linewidth}{
%    l
%    S[table-format=2.2]
%    S[table-format=2.2]
%    S[table-format=2.2]
%    S[table-format=2.2]
%}
%	\toprule
%	{\bfseries Pre-processing} &
%	{\bfseries Compress} &
%	{\bfseries Benign} &
%	{\bfseries PGD-10} &
%	{\bfseries PGD-40}\\
%	&
%	{\bfseries Rate} &
%	{\bfseries [\perc{}]} &
%	{\bfseries [\perc{}]} &
%	{\bfseries [\perc{}]}\\
%	\midrule
%	standardization %& \xmark 
%	& 0.19 & 88.34 & 58.07 & 57.33\\
%	0-1 re-scaling   %& \cmark 
%	& 0.12 & 76.30 & 33.78 & 31.10\\
%	\midrule
%\end{tabularx}

%\begin{tabularx}{\linewidth}{
%		l
%		c
%		S[table-format=2.3]
%		S[table-format=2.2]
%		S[table-format=2.2]
%		S[table-format=2.2]
%	}
%	\toprule
%	{\bfseries Pre-processing} 
%	&{\bfseries Method}
%	&{\bfseries \xflops} 
%	&{\bfseries Benign} 
%	&{\bfseries PGD-10} 
%	&{\bfseries PGD-40}
%	\\
%	&
%	&
%	&{\bfseries [\perc{}]} 
%	&{\bfseries [\perc{}]} 
%	&{\bfseries [\perc{}]}
%	\\
%	\midrule
%	\multirow{2}{*}{standardization} & %\xmark & 
%	\anpvs 		& 3.56		& 88.34		& 58.07		& 57.33\\
%	&
%	\ourmethod	& \bf 4.00	& \bf 88.93	& \bf 75.21	& \bf 73.42\\
%	\midrule
%	\multirow{2}{*}{0-1 re-scaling}  & %\cmark
%	\anpvs 		& 3.47		& 76.30		& 33.78		& 31.10\\
%	&
%	\ourmethod	& \bf 3.82	& \bf 76.89	& \bf 47.99	& \bf 46.78\\
%	\midrule
%\end{tabularx}

\begin{tabularx}{\linewidth}{
    l
    S[table-format=1.2]
    S[table-format=2.2]
    S[table-format=2.2]
}
	\toprule
	{\bfseries Pre-processing} &
	{\bfseries \xflops} &
	{\bfseries Benign} & %[\%]} &
%	{\bfseries PGD-10} &
	{\bfseries PGD-40}%[\%]}
	\\ \midrule
	Original~\citep{Madaan2020Adversarial}
	& 2.41 
	& 88.18%\pmsize{$\pm$0.5}
	& 56.21%\pmsize{$\pm$0.1} 
	\\ \midrule
	%& \xmark
	Standardization
	& 3.56	
	& 88.34
%	& 58.07 
	& 57.33
	\\
	0-1 re-scaling   %& \cmark 
	& 3.82 
	& 76.30
%	& 33.78 
	& 31.10
	\\
	\bottomrule
\end{tabularx}
	\label{tab:anp-vs-preprocess}
\end{wraptable}
%.
steps (standardization and simple 0-1 re-scaling) under the otherwise
exact same PGD settings.
The original publication reports an original accuracy of
$\SI{88.18}{\percent}$ %\pm \SI{0.5}{\percent}$ and an PGD-40
adversarial accuracy of~$\SI{56.21}{\percent}$ %\pm \SI{0.4}{\percent}$
for \vgg on \cifar[10], which match our reproduced results with
standardization (\cf~\cref{tab:anp-vs-preprocess}). We even yield higher
compression in our experiments.
With the correct input pre-processing step (\mbox{0-1 rescaling}), however,
\anpvs yields a significantly worse result, showing the criticality of
wrongly used pre-processing. With thus refrain from comparing \ourmethod
to \anpvs in pruning channels.

\begin{algorithm}[!h]
	\caption{\ourtitle}
%	\small
	\label{alg:harp-alg}
	\newcommand{\mystage}[2][]{\hspace*{\fill}\\\\\textbf{Step #1:} #2\\}
\newcommand{\myinput}[1]{\textbf{Input:} #1}
\newcommand{\myoutput}[1]{\textbf{Output:} #1}

\myinput{
	pre-trained model and its parameters $\model$,
	number layers $L$,
	%the used losses: $\lossce$, $\losshw$,
	target compression rate $\compression_{t}$,
	minimal compression rate $\compression_{min}$, 
	training data-set $\mathcal{D}$
}

\myoutput{Compressed network  $\oparams \odot \mask^{*}$}

%\mystage[1]{Robustly pre-train the network with $\lossce$ and 
%adversarial training}
%$\model = \underset{\model}{\argmin} \ 
%\replicatelabelxnoref{eq:at-inner}$

\mystage[1]{Initialization of trainable pruning parameters}
For each layer~\layer, initialize learnable rates 
$\replicatelabelxnoref{eq:rate-init}$\\
and importance scores:
\begin{center}
$\replicatelabelxnoref{eq:scores-init-w}$\hspace*{5mm}or
\hspace*{10mm}$\replicatelabelxnoref{eq:scores-init-c}$
\end{center}
%and learnable rates $\replicatelabelxnoref{eq:rate-init}$
%
%\revision{
%\mystage[1]{Initialization of learnable pruning rates by 
%\cref{eq:rate-init} 
%and importance scores by \cref{eq:scores-init-c}}
%}
%
\mystage[2]{Strategy search via robust training}
$\mask^{*} = \underset{\mask}{\argmin} \ 
 \replicatelabelxnoref{eq:total-prune-inner}$
  
  with $\replicatelabelxnoref{eq:masks}$ and
  $\replicatelabelxnoref{eq:p-mask}$

\mystage[3]{Fine-tuning of the pruned network $\prunedparams = \params \odot \mask^{*}$}
$\oparams = \underset{\prunedparams}{\argmin}~
  \underset{(\x,\y) \sim \mathcal{D}}{\mathbb{E}}
  \left[
    \underset{\delta}{\max}\left\{
      \lossce (\prunedparams, \x+\delta, \y)
    \right\}
  \right]$

\end{algorithm}

\begin{revisionblock}
\subsection{Training consumption}
\label{app:runtime}

In this section, we elaborate on the training (time) consumption of the
pruning approaches considered in our evaluation. As every method
operates on the complete training dataset, we estimate the consumption
in training epochs rather than wall-clock time as the latter varies with
the used hardware and its load. We report the results
in~\cref{tab:train-consume}.
% .
The methods either follow a three-stage pipeline~\citep{Han2015Learning}
or train the model from scratch with pruning considerations in their
optimization. Consequently, the distribution of training effort per
stage varies.
% .
Under this metric, \mad is most efficient. However, when targeting a
sparsity of \perc{99}, \ourmethod arrives close its to best performance
right after pruning (\num{20}~epochs) as can be seen
in~\cref{fig:learn-process}.
% .
At \perc{99.9} sparsity, our method uses \num{100}~epochs for
fine-tuning to recover natural accuracy but does not improve robustness
anymore. Thus, \ourmethod's pruning stage turns out to be most decisive
but helps for aggressive~pruning.
\end{revisionblock}

\vspace*{-1mm}
\begin{table}[!h]
  \tablesize
  \centering
  \caption{Training consumption in different robust pruning methods}
  \vspace*{-0.5mm}
  \newcolumntype{Y}{>{\centering\arraybackslash}X}
\newcommand{\mymidrule}{\cmidrule(lr){1-2}\cmidrule{3-5}\cmidrule(l){6-6}}

\sisetup{mode=text}
\setlength{\tabcolsep}{8pt}
\begin{tabular}{
	l
	c
	S[table-format=3]
	S[table-format=3]
	S[table-format=3]
	S[table-format=4]
}
\toprule
\bf Method
& \multirow{2}{*}{\makecell{\bf Multi-Stage Pipeline\\\bf\citep{Han2015Learning}}}
& \multicolumn{4}{c}{\bf Number of Epochs}
\\ \cmidrule{3-6}
&
& {\bf Pre-training}
& {\bf Pruning}
& {\bf Fine-tuning}
& {\bf Total}
\\
\mymidrule
\radmms~\citep{Ye2019Adversarial}    & \cmark & 100 & 100 & 100 & 300 \\
\hydra~\citep{Sehwag2020HYDRA}	     & \cmark & 100 &  20 & 100 & 220 \\
%\anpvs~\citep{Madaan2020Adversarial} & \cmark & 200 & 200 &{---}& 400 \\
\bcsp~\cite{Ozdenizci2021Bayesian}   & \xmark &{---}&{---}&{---}& 200 \\
DNR~\citep{Kundu2021DNR}             & \xmark &{---}&{---}&{---}& 200 \\
\mad~\citep{Lee2022MAD}	             & \cmark &  60 &  20 &  60 & 140 \\
\mymidrule
\bf\ourmethod (Ours)                 & \cmark & 100  &  20 & 100 & 220 \\
\bottomrule
\end{tabular}

  \label{tab:train-consume}
\end{table}

\subsection{$\text{CO}_2$ Emission}
We have conducted all our experiments on Nvidia RTX-3090 GPU cards and
have consumed about \SI{10198} % was 8950 for rebuttal was 8580 for ICLR 
%first % was 2688
{GPU~hours} in total. This amounts to an
estimated total $\text{CO}_2$ emissions of \SI{2177.27}{kgCO2eq}
% 1910.82 % 1831.83 % was 573.89
when using Google Cloud Platform in region \code{europe-west3}. 
However, our university consumes \perc{100} renewable-energy, such that 
our specific $\text{CO}_2$ emissions for the project amounts to 
\SI{1.74}{kgCO2eq}. 
% was 1.56 % was 1.50 % was 0.47
%have consumed about \SI{2688}{GPU~hours} in total based on \perc{100} 
%renewable-energy in our university. This amounts to an
%estimated total $\text{CO}_2$ emissions of \SI{573.89}{kgCO2eq} when
%using Google Cloud Platform in region \code{europe-west3}. 
%However, our
%university consumes \perc{100} renewable-energy, such that our specific
%$\text{CO}_2$ emissions for the project amounts to 
%\SI{666.66}{kgCO2eq}.
%% .
Estimates are conducted using the
\href{https://mlco2.github.io/impact#compute}{``Machine Learning Impact
	Calculator''}~\citep{lacoste2019quantifying}.

\end{document}